\documentclass[runningheads]{llncs}
\pdfoutput=1
\usepackage{eccv}
\usepackage{eccvabbrv}
\usepackage{multirow, comment}

\usepackage{graphicx}
\usepackage{booktabs}
\usepackage[numbers]{natbib}   

\usepackage[accsupp]{axessibility}
	
\usepackage{color, colortbl}
\usepackage{pifont}
\newcommand{\cmark}{\ding{51}}
\newcommand{\xmark}{\ding{55}}
\definecolor{Gray}{gray}{0.9}
\definecolor{LightCyan}{rgb}{0.88,1,1}

\newcommand{\PAR}[1]{\noindent{\bf #1~}}
\usepackage[pagebackref,breaklinks,colorlinks,citecolor=eccvblue]{hyperref}
\usepackage{hyperref}

\usepackage{orcidlink}

\newcommand{\B}[1]{{{\textbf{#1}}}}
\newcommand{\U}[1]{{{\underline{#1}}}}

\usepackage{comment}
\usepackage[acronym]{glossaries}
\usepackage{multirow}

\makeglossaries{}
\newacronym{sota}{SotA}{State-of-the-Art}
\newacronym{nn}{NN}{Nearest-Neighbor}
\newacronym{sfm}{SfM}{Structure-from-Motion}
\newacronym{slam}{SLAM}{Simultaneous Localization And Mapping}
\newacronym{tab}{Tab.}{Table}
\newacronym{sdf}{SDF}{Signed-Distance-Function}
\newacronym{udf}{UDF}{Unsigned-Distance-Function}
\newacronym{gs}{GS}{Gaussian Splats}
\newacronym{iou}{IoU}{Intersection over Union}
\newacronym{sam}{SAM}{SegmentAnything}
\newacronym{nerf}{NeRF}{Neural Radiance Field}

\begin{document}

\title{Remove360: Benchmarking Residuals\\After Object Removal in 3D Gaussian Splatting}
\titlerunning{Remove360}

\author{
  Simona Kocour\textsuperscript{1,2} \quad \quad Assia Benbihi\textsuperscript{1} \quad \quad Torsten Sattler\textsuperscript{1}\vspace{-10pt}}
\authorrunning{S.~Kocour et al.}
\institute{\textsuperscript{1}Czech Technical University in Prague, Czech Institute of Informatics, Robotics and Cybernetics \\
\textsuperscript{2}Czech Technical University in Prague, Faculty of Electrical Engineering}

\maketitle

\vspace{-2pt}
\begin{center}
    \includegraphics[width=\textwidth]{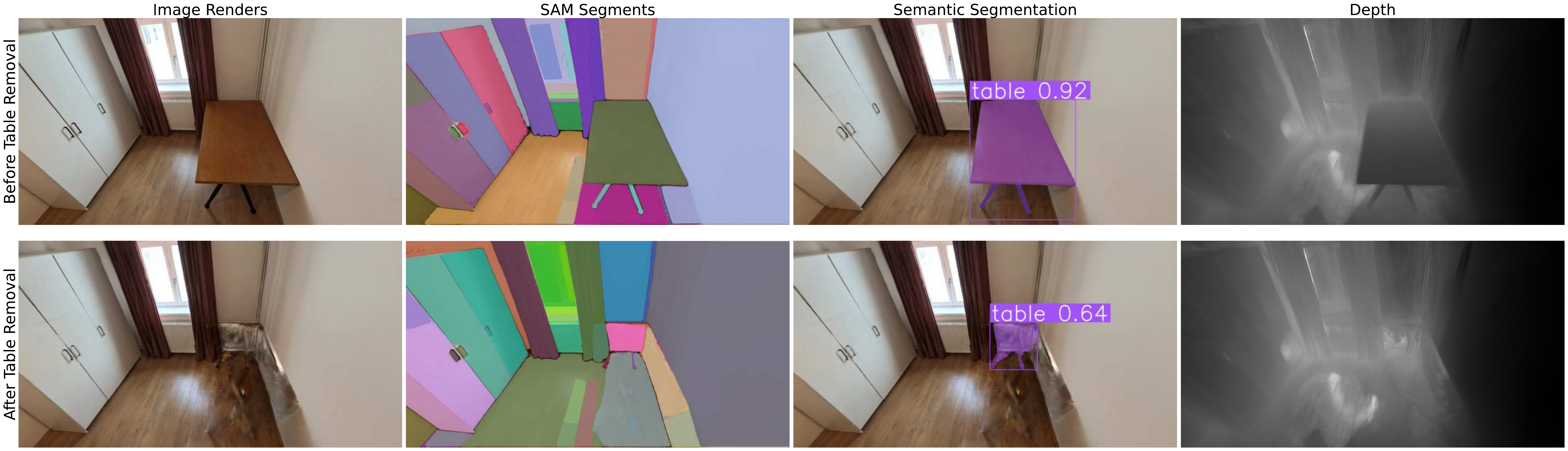}
\vspace{-20pt}
\captionof{figure}{\textbf{Recoverable object evidence after object removal in 3D Gaussian Splatting.}
Although the target object is no longer apparent in the RGB rendering, our pipeline can still detect object-like evidence.
Such evidence may arise from residual geometry, boundary artifacts, shadows, reconstruction artifacts, or inpainting artifacts. 
We evaluate removal quality using semantic, structural, and geometric probes. 
Top: scene before removal.
Bottom: scene after removal. Left to right: RGB rendering, SAM~\cite{ravi2024sam2} masks, GroundedSAM~\cite{kirillov2023segany,liu2023grounding,ren2024grounded} overlay, and rendered depth.
}
\end{center}
\vspace{-15pt}
\label{fig:teaser}

\begin{abstract}
Object removal in neural 3D scenes is usually evaluated by visual plausibility or image-level reconstruction quality. 
However, a visually plausible edit may still reveal what was removed when the edited scene is rendered and queried by downstream vision models. 
We introduce Remove360, a benchmark and evaluation protocol for measuring recoverable object evidence after object removal in 3D Gaussian Splatting. 
Remove360 contains pre- and physical post-removal RGB captures with object-level masks. 
Unlike prior object-centric or controlled removal datasets, Remove360 contains cluttered indoor and outdoor full-scene captures, enabling evaluation against the scene state after actual object removal. 
We evaluate removal quality along three oriented axes: category-level suppression, geometric change, and prompt-free structural consistency. 
We summarize these axes with a Removal Quality Score (RQS), defined as the geometric mean of the three axis scores, and use average rank as a robustness check. 
We also evaluate a 2D-first reconstruction baseline that removes target objects before reconstructing the edited 3D scene. 
The results show that current methods make different trade-offs across axes: methods can suppress semantic detections while producing poor geometry, or alter geometry while leaving object-like evidence detectable. 
Remove360 therefore complements visual-fidelity benchmarks by evaluating whether the final shared 3D representation removes recoverable object evidence, not only whether the edited image looks plausible.

Dataset and evaluation code:
\url{https://github.com/simonakocour/anything_left}.
\end{abstract}

\setlength{\abovecaptionskip}{2pt}

\section{Introduction}
\label{sec:intro}
Trainable scene representations such as Neural Radiance Fields (NeRFs)~\citep{mildenhall2020nerf,barron2022mip,reiser2021kilonerf,muller2022instant,chen2024lara,kulhanek2023tetra,martin2021nerf}  or 3D Gaussian Splatting (3DGS)~\citep{kerbl3Dgaussians,lin2024vastgaussian,yu2024gaussian,zhang2024gaussian,kulhanek2024wildgaussians,chen2024mvsplat,wang2024freesplat}, enable photorealistic 3D reconstructions from images.
Recent semantic and language-aligned extensions~\citep{kerr2023lerf,shi2024language,gaussian_grouping,wuopengaussian,zhou2024feature,hu2024semantic,jain2024gaussiancut} further allow users to query~\citep{peng2023openscene,takmaz2025search3d}, segment~\citep{peng2023openscene,takmaz2025search3d}, and edit~\citep{gaussian_grouping,zhou2024feature,chen2024dge,gu2025egolifter,choi2024click} scene content using natural-language prompts or object masks.
As consumer-grade capture tools make realistic 3D scans easier to create and share~\citep{scaniverse,nerfstudio,Yu2022SDFStudio,capturereality,polycam,ye2024gsplatopensourcelibrarygaussian,postshot,lumaAi}, users may wish to remove private or sensitive objects, such as
documents, photographs, medical items, children's objects, or personal
belongings, before sharing a reconstruction\footnote{For example, IKEA advises users to physically remove private items before scanning a room with their app, since the captured data is uploaded to their servers.}. 
This raises a fundamental question: \emph{after object removal, does the final edited 3D scene still contain recoverable evidence of the removed object?}

Existing evaluations of 3D object removal and inpainting mainly emphasize qualitative appearance, foreground/background separation, or image-level reconstruction quality when post-removal references are available~\citep{cen2023saga,choi2024click,mirzaei2023spin}. 
These criteria are important, but they do not fully validate the edited representation. 
A scene may look plausible from selected viewpoints yet still contain object-like evidence in other views, in rendered depth, in structural masks, at object boundaries, or in detector responses. 
Conversely, a method may suppress category detections while producing geometry or structure that does not match the post-removal scene. 
Evaluating 3D object removal therefore requires more than visual inspection or a single image-level score.

We use the term \emph{recoverable object evidence} to refer to any signal in the edited scene that allows a downstream vision model to localize or identify evidence associated with the removed object. 
Importantly, we do not assume that this evidence has a single causal source. 
It may arise from remaining Gaussians, incomplete geometric removal, shadows, reflections, contact artifacts, contextual structure, reconstruction errors, or content introduced by an
inpainting model. 
In a sharing scenario, all of these are relevant: the user shares the final edited scene, and any recoverable evidence in that scene can undermine the intended removal.
Therefore we evaluate the final edited 3D representation that a user would share.

Figure~\ref{fig:teaser} illustrates this setting. The top row shows the scene before removal, and the bottom row shows the edited scene after removal. 
Although the target object may no longer be apparent in the RGB rendering, downstream probes can still respond to object-like evidence in the edited scene, visible through segmentation, structural masks, or rendered depth. 
This example is not intended to identify a single causal source of evidence; rather, it motivates evaluating the final edited representation across semantic, structural, and geometric axes.

To support this evaluation, we introduce \emph{Remove360}, a real-world benchmark for object-removal quality in 3DGS. 
Remove360 contains pre- and physical post-removal captures of 11 indoor and outdoor scenes, together with object-level masks. 
Unlike controlled or object-centric removal datasets, such as 360-USID~\cite{wu2025aurafusion}, Remove360 captures full-scene settings with natural occlusions, contact regions, shadows, and interactions among multiple objects. 

Our protocol measures removal quality along three complementary axes: category-level suppression (Eqs. ~\eqref{eq:gsam_iou}--~\eqref{eq:acc_seg}), prompt-free structural consistency (Eq.~\eqref{eq:sim_sam}), and geometric modification (Eqs.~\eqref{eq:depth_part1}--~\eqref{eq:acc_depth}).
The probing models operate on the full rendered image, while ground-truth masks are used only for scoring within the removed-object region. 
This keeps the quantitative score localized to the intended edit region. 
To summarize these axes, we introduce the \emph{Removal Quality Score} (RQS), the geometric mean
of the oriented semantic, geometric, and structural scores. 
RQS provides a compact aggregate score, while the individual axes expose the trade-offs behind each method's behavior.

We evaluate representative 3DGS object-removal pipelines together with a 2D-first reconstruction baseline, where target objects are removed from the training images before reconstructing a new edited 3D scene. 
Our experiments show that current removal methods make different trade-offs across axes.
Some edits suppress category-level detections but fail geometrically or structurally, while
others modify geometry without fully removing object-like evidence. 
These results motivate evaluating 3D object removal as a multi-axis problem rather than by visual appearance, category suppression, or depth change alone.

In summary, our contributions are:
\begin{enumerate}
    \item We introduce the problem of quantifying object evidence after its
    removal in trainable 3D scene representations, framing object removal as a
    privacy-relevant evaluation task beyond visual appearance.

    \item We present \emph{Remove360}, a pre-/post-removal benchmark for
    evaluating object-removal quality in full 3DGS scenes, with physical
    post-removal reference images and object-level masks.

    \item We propose a multi-axis evaluation protocol that measures
    category-level suppression, geometric modification, and prompt-free structural consistency, and summarize these axes with the Removal Quality Score (RQS).

    \item We evaluate representative 3DGS object-removal pipelines and a
    2D-first reconstruction baseline, showing that visual plausibility or
    category suppression alone is insufficient for assessing 3D object removal.
\end{enumerate}

\section{Related Work}
\label{sec:related_work}
\PAR{Trainable 3D Scene Representations.}
Reconstructing scenes from images has long relied on point clouds~\cite{schonberger2016structure,yang2023sam3d,huang2024segment3d,peng2023openscene,liu2024uni3d,yin2024sai3d}, meshes~\cite{schoenberger2016mvs,furukawa2009accurate,lazebnik2001computing,kundu2020virtual}, and recently on trainable representations that enable high-quality novel-view synthesis.
Neural Radiance Fields (NeRFs)~\citep{mildenhall2020nerf,barron2022mip,reiser2021kilonerf,muller2022instant,chen2024lara,kulhanek2023tetra,martin2021nerf} represent scenes implicitly as radiance fields, while 3D Gaussian Splatting (3DGS)~\citep{kerbl3Dgaussians,lin2024vastgaussian,yu2024gaussian,zhang2024gaussian,kulhanek2024wildgaussians,chen2024mvsplat,wang2024freesplat} provides an explicit representation with fast rendering and straightforward component-level manipulation.
Our work focuses on 3DGS because its explicit structure makes object-level edits practical and widely adopted.

\PAR{Linking 3D Reconstructions and Semantics.}
Both NeRFs and 3DGS can be augmented with semantic or language-aligned features to support open-vocabulary querying and interaction~\citep{kerr2023lerf,tschernezki2022neural,wang2022clip,mirzaei2022laterf,shi2024language,liao2024clip,qin2024langsplat,gaussian_grouping,wuopengaussian,zhou2024feature,hu2024semantic,jain2024gaussiancut}.
These representations enable locating objects with text prompts~\citep{peng2023openscene,huang2024segment3d,takmaz2025search3d,liang2024supergseg,koch2024relationfield,kim2025drsplat,zhu2025objectgs} and associating regions with masks or instance labels for editing~\citep{gaussian_grouping,zhou2024feature,chen2024dge,gu2025egolifter,choi2024click,wang2025ag2aussian,mazzucchelli2026beags}.
Foundation models such as SAM/SAM2~\citep{kirillov2023segany,ravi2024sam2}, CLIP~\citep{radford2021learning}, DINO/DINOv2~\citep{caron2021emerging,oquab2024dinov2}, and open-vocabulary segmentation/detection models~\citep{lilanguage,liu2023grounding,ren2024grounded} further accelerate this trend by providing strong 2D supervision and query mechanisms.
Our evaluation leverages such off-the-shelf models as probes to measure what recoverable object evidence remains after removal.

\PAR{Object Removal and Editing in Trainable 3D Representations.}
A growing body of work explores object removal and editing in NeRFs and 3DGS, often using text features, distilled semantics, or multi-view mask lifting to identify Gaussians or radiance-field regions to modify~\citep{cen2023saga,mirzaei2023spin,gaussian_grouping,zhou2024feature,hu2024semantic,jain2024gaussiancut,wu2025aurafusion,wang2024gscream}. 
Recent approaches also incorporate cross-view consistency and inpainting to improve visual plausibility after removal~\citep{liu2024infusion,wu2025aurafusion,huang20253d,zhao2026goris,shi2025imfine,cheng2025painpainter,pan2025diga3d}.
However, evaluation in this line of work typically emphasizes visual quality, foreground/background separation, or qualitative examples, and often reports pixel-level metrics only when post-removal ground truth exists.
In contrast, we evaluate object removal through the lens of residual detectability: even when an object is visually removed, can a model still detect that it was present, or infer its type?
We propose metrics that quantify residual detectability rather than only appearance fidelity.

\PAR{Benchmarks and Evaluation for Object Removal.}
Existing datasets and evaluation protocols for 3D object removal are typically object-centric and limited in scene complexity, which restricts the ability to assess removal performance in realistic settings.
360-USID~\cite{wu2025aurafusion}, for example, focuses on staged single-object removals under controlled capture conditions.
While such settings are valuable for studying clean object isolation, they under-represent clutter, occlusion, and context-dependent artifacts that commonly occur in real user scans.

Remove360 complements these benchmarks with complex indoor and outdoor scenes containing interacting objects and natural occlusions.
Our RQS summarizes category-level suppression, geometric change, and structural agreement, while per-axis scores expose trade-offs that visual metrics alone can hide.

\PAR{Privacy in Visual Data and 3D Capture.}
As visual data becomes increasingly central to modern AI systems~\cite{rombach2022high,saharia2022photorealistic,brooks2024video,achiam2023gpt}, concerns about user privacy have gained renewed attention in research, industry, and policy~\cite{raina2023egoblur,speciale2019privacy,pittaluga2019revealing,chelani2023privacy,moon2024raycloud,nasr2023effectively,illman2019california,voigt2017eu}.
This concern is amplified by the growing deployment of sensor-equipped devices such as AR/VR glasses and mobile scanning tools~\citep{spectacles,engel2023project,xreal}, which continuously capture and reconstruct user environments.

One approach to privacy preservation is to anonymize images before reconstruction~\cite{liu2024infusion,weder2023removing}.
In this work, we consider a complementary and increasingly practical scenario: a user edits a reconstructed 3D scene, \eg by removing personal objects, before sharing it.
The implicit assumption is that once an object is removed, it is no longer recoverable from the shared representation.
We question this assumption.
Our goal is not to reconstruct pixel-level private content, such as document text or the appearance of a specific photograph. 
Instead, we evaluate whether the final edited representation still supports target-associated category
detections or contains structural and geometric evidence in the edited region.
Such evidence is privacy-relevant because it may reveal the existence or category of a removed object, but it is not equivalent to recovering its identity, appearance, or textual content.

\section{Quantifying Recoverable Object Evidence}
\label{sec:metrics}

This section defines a protocol for evaluating whether an edited 3D scene still contains recoverable object evidence after removal. 
Given a scene before and after editing, we render both versions from multiple viewpoints and probe the resulting images with off-the-shelf vision models~\cite{kirillov2023segany,liu2023grounding,ren2024grounded,ravi2024sam2}.
Models operate on the full rendered image without access to ground-truth masks.
Ground-truth masks are used only for scoring within the removed-object region.
Therefore, the evaluation captures any remaining evidence in the final edited representation, including geometric, structural, contextual, or inpainting-related cues.

For each evaluation view $v\in\mathcal V$, let $I_{\mathrm{pre}}^v$, $I_{\mathrm{post}}^v$, and $I_{\mathrm{ref}}^v$ denote renderings of the original, edited, and post-removal reference reconstructions at the same camera pose. 
Let $D_{\mathrm{pre}}^v$ and $D_{\mathrm{post}}^v$ denote the corresponding depth maps, and let $M^v$ denote the target-object mask.

\vspace{-10pt}
\subsection{Semantic Suppression}
\label{sec:method_iou_drop}
\begin{figure}[t]
    \centering
    \includegraphics[width=1\linewidth]{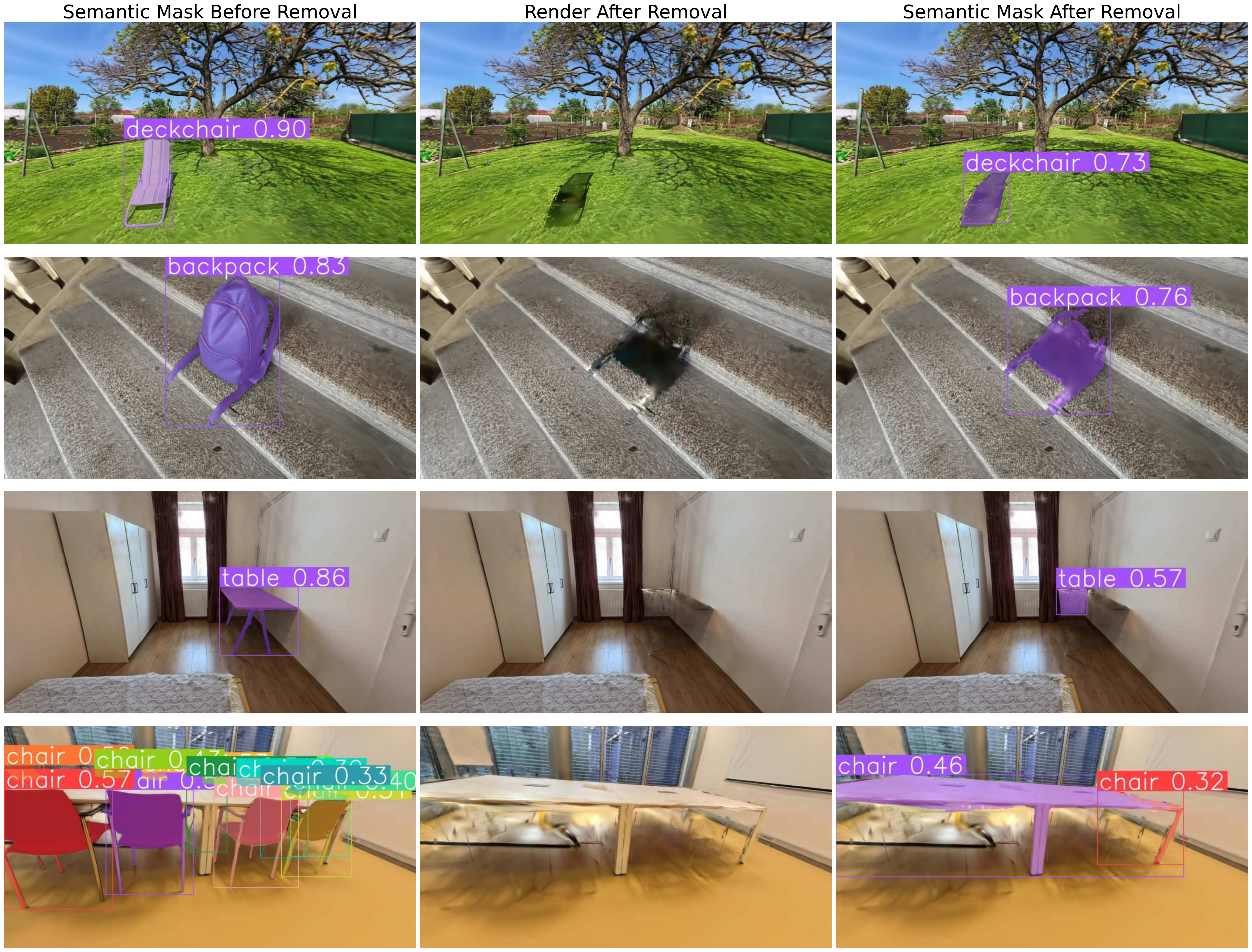}
    \caption{\textbf{Semantic segmentation before and after removal on Remove360.}
    GroundedSAM2~\cite{kirillov2023segany,liu2023grounding,ren2024grounded} detections shown before and after editing. 
    Although the edited renderings appear visually plausible, the semantic probe~\cite{kirillov2023segany,liu2023grounding,ren2024grounded} still identifies target-associated evidence in the final representation.
    The predicted masks are used to compute $\mathrm{IoU}_{drop}$ (Eq.~\eqref{eq:gsam_iou}) and $\mathrm{acc}_{seg}$ (Eq.~\eqref{eq:acc_seg}).
    Rows show different object removals.
    }
    \label{fig:new_groundedSam}
    \vspace{-15pt}
\end{figure}
We first evaluate whether the removed object category remains detectable after editing. 
As illustrated in Fig.~\ref{fig:new_groundedSam}, we use an open-vocabulary segmentation model as a category-level probe~\cite{lilanguage,liu2023grounding,ren2024grounded}.
The model receives only the rendered RGB image, before and after removal, and the object category prompt, without access to the ground-truth mask.
The predicted segmentation is compared with the target-object mask only during evaluation.

For each view $v$, let $\hat{M}_{\mathrm{pre}}^v$ and $\hat{M}_{\mathrm{post}}^v$ denote the predicted masks before and after removal.
We first compute the Intersection over Union (IoU)~\cite{chen2017deeplab,badrinarayanan2017segnet} and report the change in detectability:
\begin{equation}
\label{eq:gsam_iou}
\begin{array}{c}
\mathrm{IoU}^v_{drop}
=
\mathrm{IoU}(\hat{M}_{\mathrm{pre}}^v,M^v)
-
\mathrm{IoU}(\hat{M}_{\mathrm{post}}^v,M^v),
\\ 
\mathrm{IoU}_{drop}
=
\frac{1}{|\mathcal{V}|}
\sum_{v\in\mathcal{V}}
\mathrm{IoU}^v_{drop}.
\end{array}
\end{equation}
The metric ranges from $-1$ to $1$, where larger values indicate stronger reduction of category-level evidence. 
However, $\mathrm{IoU}_{drop}$ can be ambiguous: a small value may indicate that the object remains detectable after removal, or that it was poorly detected even before removal. 
Still, $\text{IoU}_{\text{drop}}$ is useful on its own as a warning signal, especially in interactive systems where human oversight is possible, \eg, active labeling.
In the experiments, we also report a more intuitive metric, the performance of the segmentation after removal, and analyse its correlation with $\text{IoU}_{\text{drop}}$.
We complement $\mathrm{IoU}_{drop}$ with a direct post-removal metric, $\mathrm{acc}_{seg}$, which measures the fraction of views where the removed object is no longer localized by the semantic probe:

\begin{equation}
\label{eq:acc_seg}
\mathrm{acc}_{\mathrm{seg},\xi_{\mathrm{IoU}}}
=
\frac{1}{|\mathcal{V}|}
\sum_{v\in\mathcal{V}}
\mathbf{1}
\left[
\mathrm{IoU}
(\hat{M}_{\mathrm{post}}^v,M^v)
<
\xi_{\mathrm{IoU}}
\right],
\end{equation}
where $\xi_{\mathrm{IoU}}$ is the selected IoU threshold for considering an object as detected. 
We set $\xi_{\mathrm{IoU}}=0.3$ in all experiments and provide a sensitivity analysis for different thresholds in the supplementary material.
Higher values indicate stronger category-level suppression.

The two semantic metrics capture complementary aspects:
IoU$_{\text{drop}}$ measures suppression relative to the original detectability,
while acc$_{\text{seg}}$ directly measures whether the object remains recognizable after removal.

\vspace{-10pt}
\subsection{Structural Change}

\begin{figure}[t]
    \centering
    \includegraphics[width=\linewidth]{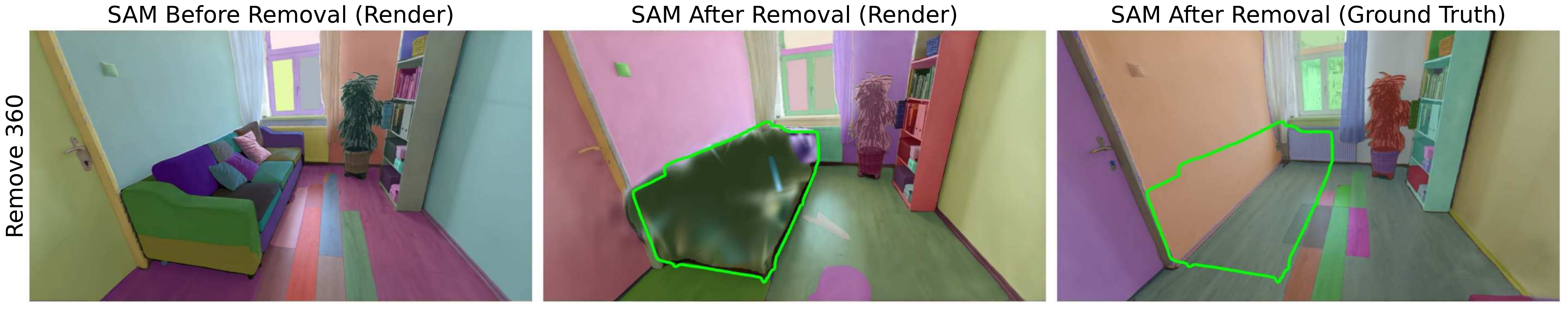}
    \caption{\textbf{Prompt-free structural evaluation.} SAM~\cite{ravi2024sam2} masks are extracted independently without category prompts, object outline in green. On Remove360, the edited rendering is compared with the pose-matched reference rendering, and higher $\mathrm{sim}_{\mathrm{SAM}}$ indicates better structural agreement. When no physical reference is available, the pre- and post-removal renderings are compared and we report $1-\mathrm{sim}_{\mathrm{SAM}}$, so that higher values indicate stronger structural change.
}
    \label{fig:new_sam}
    \vspace{-15pt}
\end{figure}

While semantic segmentation evaluates category-level detectability, structural cues may persist even when category recognition is suppressed. 
We therefore introduce an instance-level metric based on Segment Anything (SAM)~\cite{ravi2024sam2}, which produces object-agnostic masks without requiring text prompts.

SAM is applied without prompts to generate a set of masks in the rendered images (Fig.~\ref{fig:new_sam}). 
We compare the mask sets before and after removal to measure structural changes caused by editing.
If object structure remains after removal, masks in the edited region will remain similar to those before removal. 
Successful removal should lead to structural changes in the removed region.

Depending on reference availability, we compare masks from the edited rendering either with the pose-matched reference rendering or with the pre-removal rendering. 
Higher similarity indicates better reference agreement in the former setting; lower similarity indicates stronger structural change in the latter.

More formally, for view $v$, let
\begin{equation}
\mathcal A^v=\{a_i^v\}_{i=1}^{N_v},
\qquad
\mathcal B^v=\{b_j^v\}_{j=1}^{M_v}
\end{equation}
denote the two sets of retained SAM masks.
Let $\Pi^v$ denote the set of all valid one-to-one matchings between
indices in $\mathcal A^v$ and $\mathcal B^v$.
We obtain the optimal assignment
\begin{equation}
\label{eq:sim_sam_matching}
\pi^{v*}
=
\arg\max_{\pi\in\Pi^v}
\sum_{(i,j)\in\pi}
\mathrm{IoU}(a_i^v,b_j^v)
\end{equation}
using the Hungarian algorithm.
The per-view similarity is
\begin{equation}
\label{eq:sim_sam}
\mathrm{sim}_{\mathrm{SAM}}^v
=
\begin{cases}
1,
& N_v=M_v=0,
\\[2pt]
0,
& \min(N_v,M_v)=0 \text{ and } \max(N_v,M_v)>0,
\\[4pt]
\displaystyle
\frac{
\sum_{(i,j)\in\pi^{v*}}
\mathrm{IoU}(a_i^v,b_j^v)
}{
\max(N_v,M_v)
},
& \text{otherwise}.
\end{cases}
\end{equation}
The dataset-level score is
\begin{equation}
\label{eq:sim_sam_overall}
\mathrm{sim}_{\mathrm{SAM}}
=
\frac{1}{|\mathcal V|}
\sum_{v\in\mathcal V}
\mathrm{sim}_{\mathrm{SAM}}^v.
\end{equation}
The metric ranges from $0$ to $1$. 
The normalization by $\max(N_v,M_v)$ accounts not only for the spatial overlap of matched masks, but also for differences in the number of detected regions. 
Thus, methods that fragment an object into many regions or leave additional object-like structures are penalized. 
We retain masks whose overlap with the target region exceeds 0.1 and compute the reported similarity using the full retained masks. 
The supplementary erosion/dilation analysis, Supp.~\ref{supp:boundary_sensitivity}, evaluates sensitivity to perturbations of the target region boundary. 

Unlike semantic metrics, $\mathrm{sim}_{SAM}$ does not rely on object categories
or text prompts. 
It measures object-level structural consistency and captures changes caused by residual geometry, incomplete removal, or reconstruction and inpainting artifacts. 

\vspace{-10pt}
\subsection{Geometric Change} \label{subsec:depth_definition}

\begin{figure*}[t]
    \centering
    \includegraphics[width=1\linewidth]{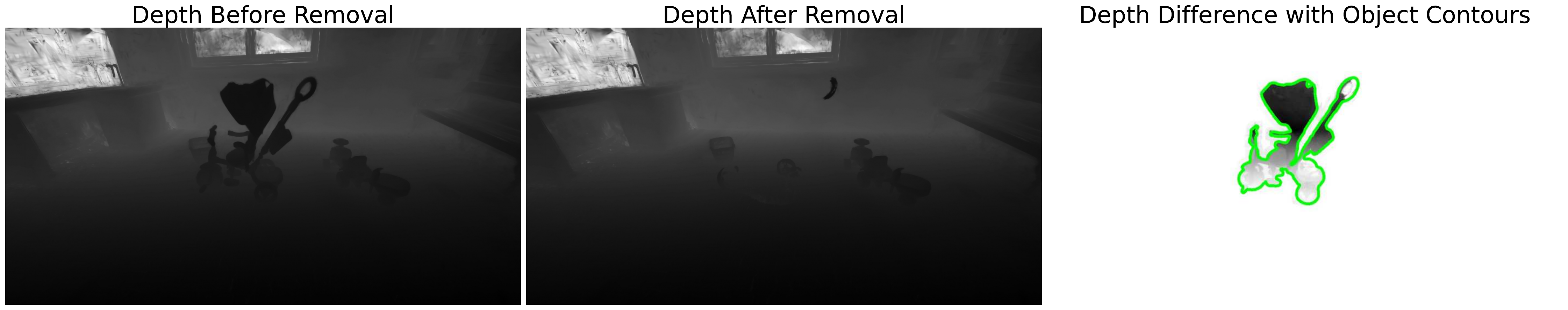}
    \caption{\textbf{Geometric changes before and after editing.}
    Left to right: rendered depth before editing, rendered depth after editing,
    thresholded depth difference, and the ground-truth outline of the removed
    object shown in green. The depth difference is used to compute
    $\mathrm{acc}_{\Delta depth}$ in Eq.~\eqref{eq:acc_depth} and highlights
    regions where the geometry was modified. The example illustrates that some
    areas inside the target-object region can remain insufficiently modified
    after editing.
    }
    \label{fig:new_depth}
    \vspace{-15pt}
\end{figure*}

Semantic segmentation evaluates whether object categories remain detectable after editing, while the structural metric evaluates changes in object-level regions.
However, successful object removal also requires modifying the underlying geometry of the reconstructed scene. 
We therefore introduce a geometry-based metric to assess whether removal alters the underlying 3D structure.
This provides an independent measurement of geometric modification and reduces
the influence of possible errors in semantic segmentation.

Inspired by scene change detection methods~\citep{adam2022objects}, we measure object removal through changes in the rendered depth maps before and after editing. 
A significant change in the depth map indicates that the underlying scene geometry has changed in the edited region. 
This metric does not directly measure semantic detectability, but captures incomplete geometric modification and provides a complementary signal independent of learned segmentation models.

For each evaluation view $v$, let $D_{\mathrm{pre}}^v$ and $D_{\mathrm{post}}^v$ denote the rendered depth maps before and after removal.
The absolute depth difference at pixel $p$ is defined as:
\begin{equation}
\label{eq:depth_part1}
\Delta D^v(p)
=
\left|
D_{\mathrm{pre}}^v(p)
-
D_{\mathrm{post}}^v(p)
\right|.
\end{equation}

A direct comparison of depth values is sensitive to small reconstruction errors, rendering noise, and differences in depth estimation. 
Therefore, instead of using a fixed threshold, we automatically determine a view-dependent threshold $\xi_{\mathrm{depth}}^v$ from the depth-change distribution. 
The threshold is derived automatically using Generalized Histogram Thresholding~\citep{BarronECCV2020} on the histogram of depth differences over the whole image (Fig.~\ref{fig:new_depth}).
We define the depth-change rate as the ratio of pixels inside the removed-object region whose depth changes exceed the selected threshold:
\begin{equation}
\label{eq:acc_depth}
acc_{\Delta depth}^{v}
=
\frac{1}{|M^v|}
\sum_{p\in M^v}
\mathbf{1}
\left[
\Delta D^v(p)>\xi_{\mathrm{depth}}^{v}
\right], \quad 
acc_{\Delta depth}
= \frac{1}{|\mathcal V|}
\sum_{v\in\mathcal V}
acc_{\Delta depth}^{v}.
\end{equation}
The metric is averaged over all evaluation views and measures the fraction of pixels within the target-object region whose depth changes exceed the selected threshold. 
Higher values indicate stronger geometric modification after editing.
The target-object pixels are defined by the available ground-truth object mask. 
This metric is intentionally complementary to semantic and structural measurements: large geometric changes do not necessarily imply complete object removal, since the edited representation may
still contain object-related structures or reconstruction artifacts.

\subsection{Removal Quality Score}
\label{sec:rqs}
The previous metrics capture complementary aspects of object removal: semantic suppression, prompt-free structural changes, and geometric modification. 
However, individual metrics may favor different methods. 
For example, a method may achieve large geometric changes while leaving semantic evidence, or strongly suppress semantic predictions while introducing structural artifacts. 
Therefore, we combine the three evaluation axes into a single aggregate score.

For each method, we define three removal quality axes in $[0,1]$, oriented such that higher values always indicate better removal quality:
\begin{equation}
\begin{array}{c}
s_{\mathrm{sem}}
=
\mathrm{acc}_{\mathrm{seg},\xi_{\mathrm{IoU}}}, 
\\
s_{\mathrm{geo}}
=
\mathrm{acc}_{\Delta\mathrm{depth}},
\end{array}
\end{equation}
and
\begin{equation}
s_{\mathrm{str}}
=
\begin{cases}
\mathrm{sim}_{\mathrm{SAM}}
\left(I_{\mathrm{post}},I_{\mathrm{ref}}\right),
&
\text{if a post-removal reference is available},
\\[3pt]
1-\mathrm{sim}_{\mathrm{SAM}}
\left(I_{\mathrm{pre}},I_{\mathrm{post}}\right),
&
\text{otherwise}.
\end{cases}
\label{eq:structural_axis}
\end{equation}

For datasets without physical post-removal references, the structural score is therefore oriented such that larger values correspond to stronger structural change in the removed-object region. 
For datasets with references, it measures agreement with the desired final scene.

We define the Removal Quality Score (RQS) as the geometric mean of the three
axes:
\begin{equation}
\label{eq:rqs}
\mathrm{RQS}
=
\left(
s_{\mathrm{sem}}
\cdot
s_{\mathrm{geo}}
\cdot
s_{\mathrm{str}}
\right)^{1/3}.
\end{equation}

The geometric mean is chosen because it penalizes unbalanced performance.
Unlike an arithmetic mean, the geometric mean prevents a method from obtaining a high score by optimizing only one axis while failing on another.
Achieving a high score requires a method to perform well across all three dimensions rather than optimizing a single criterion.
Aggregating the axes is only justified if they are not substitutes for one
another. We verify this in Supp.~\ref{supp:axis_independence}: across the 66
(scene, method) cells of Remove360 the geometric axis is essentially unrelated
to the other two ($|\rho| \leq 0.13$), and the moderate raw association between
the semantic and structural axes ($\rho = 0.573$) is largely shared scene
difficulty, falling to $\rho = 0.299$ once the scene effect is removed.
The three axes therefore order methods close to independently.
In addition to RQS, we report the average rank across the individual metrics as a scale-independent robustness analysis. 
This prevents the final ranking from being dominated by the numerical range of any single metric.
Let
$\mathcal{Q}=\{\mathrm{sem},\mathrm{geo},\mathrm{str}\}$,
and let $r_q(m)$ denote the rank of method $m$ according to its dataset-level score on axis $q$, where rank 1 is best.
Ties are assigned their average rank.
We define
\begin{equation}
\mathrm{AvgRank}(m)
=
\frac{1}{|\mathcal{Q}|}
\sum_{q\in\mathcal{Q}} r_q(m).
\label{eq:avg_rank}
\end{equation}
Lower values indicate more consistent performance across the three axes.
We use the mean rather than the median of the three ranks deliberately: with
only three axes the median is the middle rank and is by construction blind to
the worst one, which is the failure mode this protocol exists to detect.
Supp.~\ref{supp:rank_aggregation} reports both statistics and shows that median
rank would promote a method that ranks last on the geometric axis to first
place overall.
IoU$_{\mathrm{drop}}$ and the image-fidelity metrics are reported as
diagnostics but are not included in AvgRank.

\section{Remove360: A Real-World Benchmark for 3D Object Removal}

\begin{figure}[t]
    \centering
    \includegraphics[width=\linewidth]{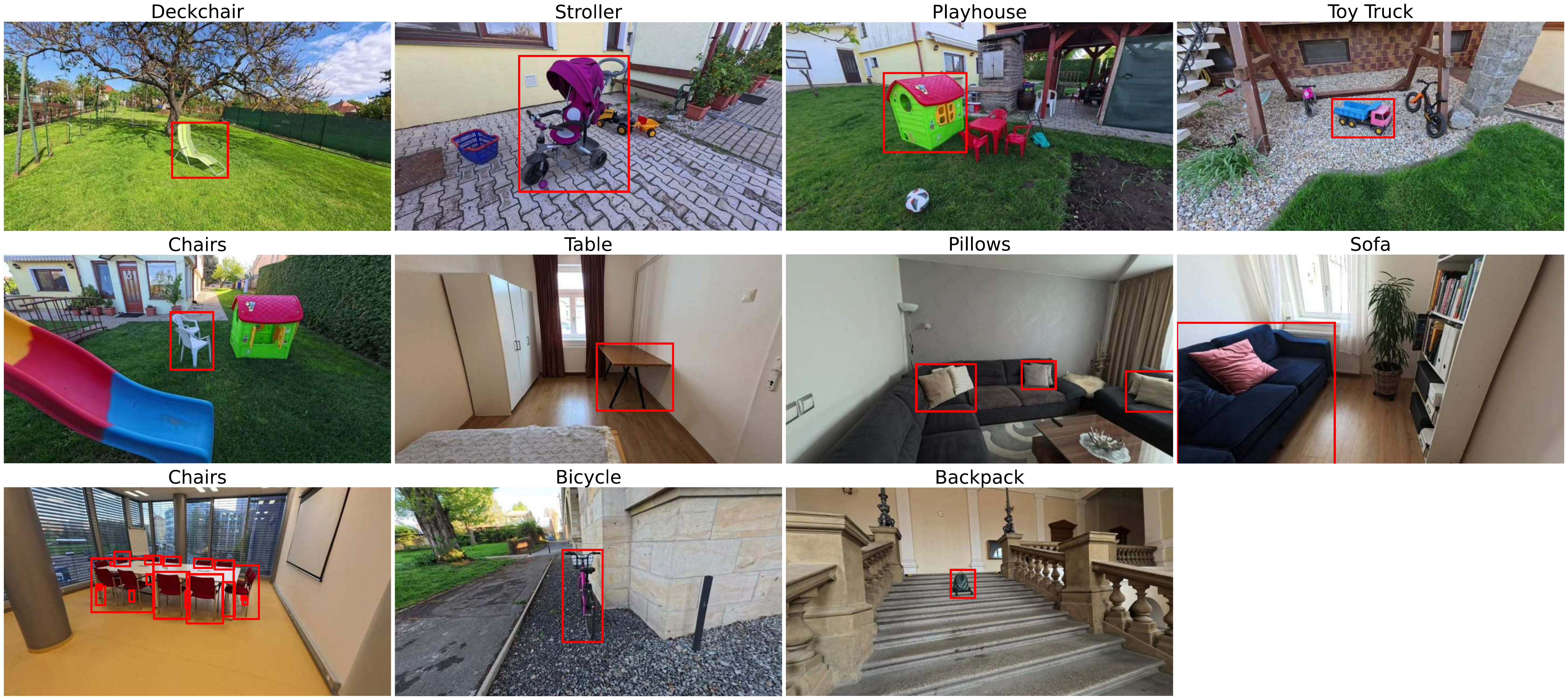}
    \caption{
    \textbf{Overview of the Remove360 dataset.}
    Samples from 11 scenes (5 indoor, 6 outdoor) with diverse object categories, layouts, and levels of clutter. Removed objects are highlighted with bounding boxes.
    }
    \label{fig:neurips_dataset}
    \vspace{-15pt}
\end{figure}
To enable rigorous evaluation of object removal in realistic settings, we introduce Remove360, a dataset of pre- and post-removal RGB captures with object-level annotations. 
Unlike 360-USID~\cite{wu2025aurafusion}, which focuses on staged single-object removals under controlled alignment, Remove360 targets real-world scenes with interacting objects and natural occlusions. 
This setup enables evaluation of removal methods in full-scene contexts beyond isolated object editing.

\PAR{Dataset Composition.}
Remove360 contains 11 scenes (5 indoor, 6 outdoor), illustrated in
Fig.~\ref{fig:neurips_dataset}. Each scene includes:

(1.) \textbf{Pre-removal views:} RGB images of the original scene with the target object present, together with object masks defining the evaluation region.

(2.) \textbf{Post-removal views:} RGB images of the same scene after physically removing the selected object(s), providing a real-world reference for evaluating edited reconstructions.

Each scene contains 150--300 views before and after removal. 
Removed objects range from compact items (\eg \textit{backpack, toy truck}) to larger or geometrically complex objects (\eg \textit{sofa, bicycle}), and some scenes contain multiple removed instances (\eg \textit{pillows, chairs}).
Object masks were initialized using prompted SAM~\cite{ravi2024sam2} and refined manually; details are in Supp.~\ref{supp:object_masks}. 
To evaluate robustness to minor annotation inaccuracies, we perform mask erosion and dilation analysis and verify that method rankings remain stable under boundary perturbations (Supp.~\ref{supp:boundary_sensitivity}).

\PAR{Dataset Collection Protocol.}
We recorded 4K, 60~fps videos using an Insta360 AcePro camera. 
Each scene was captured before and after physically removing the target object along a similar camera trajectory under comparable lighting conditions.
Frames were selected from the captured videos using variance-of-Laplacian sharpness scoring~\cite{marr1980theory}. 
Camera poses were estimated independently for the pre- and post-removal sequences using the hloc
structure-from-motion pipeline~\cite{sarlin2019coarse}, with SuperPoint
~\cite{detone2018superpoint} and LightGlue~\cite{lindenberger2023lightglue}
for feature matching (Supp.~\ref{supp:dataset_protocol}).

\vspace{-5pt}
\section{Experiments}
\label{sec:eval}

\PAR{Methods.}
We evaluate existing 3D Gaussian Splatting object-removal methods, including Gaussian Cut (GC)~\cite{jain2024gaussiancut}, Gaussian Grouping (GG)~\cite{gaussian_grouping}, SAGS~\cite{hu2024semantic}, Feature 3DGS~\cite{zhou2024feature}, AuraFusion360 (AF)~\cite{wu2025aurafusion}, and 3DGIC~\cite{huang20253d}.
For methods with optional inpainting (AuraFusion360+i and 3DGIC+i), we report the full inpainting variants as well.

We focus on Gaussian Splatting because its explicit representation supports direct manipulation of scene components. 
However, the proposed evaluation is representation-agnostic and applies to any editable 3D representation that provides RGB and depth renderings before and after removal. 
We therefore additionally evaluate the 2D-first ObjectClearer (OC)~\cite{zhao2026objectclear}. 
The method set differs by dataset: Feature 3DGS~\cite{zhou2024feature}, Gaussian Grouping~\cite{gaussian_grouping}, and SAGS~\cite{hu2024semantic} did not produce valid removal outputs on Remove360, while 3DGIC~\cite{zhou2024feature} and ObjectClearer~\cite{zhao2026objectclear} were not included in the earlier Mip-NeRF360 experiments. 
Detailed method configurations are provided in Supp.~\ref{supp:methods}.

\PAR{Datasets.}
Methods are evaluated on our Remove360 dataset and the Mip-NeRF360 dataset~\cite{barron2022mip}. 
Remove360 provides real-world pre- and post-removal captures with object-level annotations, enabling direct evaluation of edited reconstructions.
Since Mip-NeRF360~\cite{barron2022mip} does not provide post-removal references, we generate pseudo-ground-truth object masks using SAM2~\cite{ravi2024sam2} and human annotations (Supp.~\ref{supp:object_masks}).
Mip-NeRF360 is included because it is commonly used by existing 3D editing methods.

\PAR{Evaluation Protocol.}
Due to memory constraints, we use 100--150 images per selected Mip-NeRF360
scene (\textit{kitchen, counter, room, garden}), following the settings used in
prior 3D editing studies~\citep{jain2024gaussiancut,zhou2024feature}.
All methods use the same image subsets. Remove360 uses all available input views, which are also the evaluation views.
For evaluation, we use a Grounded-SAM2-based pipeline~\cite{kirillov2023segany,ravi2024sam2,liu2023grounding,ren2024grounded}
to compute category-level semantic metrics and SAM~\cite{ravi2024sam2} for prompt-free structural evaluation.
We report the individual metrics introduced in Sec.~\ref{sec:metrics} and the
proposed Removal Quality Score (RQS) in Eq.~\eqref{eq:rqs}.
To provide a scale-independent comparison, we additionally report the average rank across the three oriented RQS axes: category-level suppression, geometric modification, and structural agreement.
Higher RQS indicates stronger removal quality, while lower average rank
indicates better overall performance.

\subsection{Results}
\PAR{Overall Comparison.}
Table~\ref{tab:main_results} demonstrates that object removal quality cannot be captured by a single criterion. 
Detailed results are in Supp.~\ref{supp:results_remove360}.
Methods achieve different trade-offs between semantic suppression, geometric modification, structural consistency, and visual fidelity.
No method dominates all evaluation axes.
3DGIC+i~\cite{huang20253d} obtains the numerically highest RQS (0.693) and lowest average rank (2.33), closely followed by 3DGIC (0.690).
Both methods obtain strong geometric-modification scores.
These aggregate differences are, however, within the resolution of an 11-scene
benchmark: a paired scene-level bootstrap places Gaussian Cut and 3DGIC above
3DGIC+i on $27\%$ and $30\%$ of resampled scene sets, so the three leading
methods are not separated by the current scene count (Supp.~\ref{supp:reliability}).
The large per-axis separations discussed below are resolved reliably.
ObjectClearer~\cite{zhao2026objectclear} achieves the strongest category-level suppression ($\mathrm{acc}_{\mathrm{seg}}=0.976$) and the highest reported structural agreement, but its substantially lower geometric-modification score ($0.320$) limits its aggregate RQS.
Gaussian Cut~\cite{jain2024gaussiancut} achieves the strongest pixel-level fidelity, but falls behind on the evidence-oriented axes.
This highlights the importance of evaluating the final 3D representation: removing objects before reconstruction suppresses semantic responses, yet ObjectClearer has the lowest depth-change score.
Together, these results confirm the motivation of our benchmark: successful object removal requires eliminating recoverable object evidence while maintaining consistency with the desired edited 3D scene, rather than only improving visual appearance or suppressing individual detection models.

\begin{table}[t]
\centering
\scriptsize
\caption{
\textbf{Overall evaluation on Remove360.}
Left: residual-based metrics evaluate semantic suppression, structural agreement,
and geometric modification after object removal. RQS summarizes the balance
between these complementary dimensions, while AvgRank reports the average ranking
across axes.
Right: pixel-level reconstruction metrics measure similarity to pose-matched
post-removal reference renderings. These metrics capture visual fidelity but do not
directly measure whether object evidence remains recoverable.
\B{Best} and \U{second best} values highlighted.
}
\vspace{-10pt}
\label{tab:main_results}
\resizebox{\textwidth}{!}{
\begin{tabular}{lcccccc|ccc}
\toprule
Method &
IoU$_{\mathrm{drop}}\uparrow$ &
acc$_{\mathrm{seg}}\uparrow$ &
acc$_{\Delta depth}\uparrow$ &
sim$_{\mathrm{SAM}}\uparrow$ &
RQS$\uparrow$ &
AvgRank$\downarrow$ &
PSNR$\uparrow$ &
SSIM$\uparrow$ &
LPIPS$\downarrow$
\\
\midrule
Gaussian Cut~\cite{jain2024gaussiancut}
&   0.832   & 0.875     & 0.749     & \U{0.472} & 0.676     & 4.00  & \B{18.28} & \B{0.572} & \B{0.310}\\
AuraFusion360~\cite{wu2025aurafusion}
& \U{0.838} & 0.891     & 0.685     & 0.409     & 0.630     & 4.50      & 17.94     & \U{0.570} & \U{0.372}\\
AuraFusion360 + i~\cite{wu2025aurafusion}
&   0.800   & 0.891     & 0.775     & 0.407     & 0.655     & 4.17      & 17.09     & 0.553     & 0.473\\
3DGIC~\cite{huang20253d}
&   0.772   & 0.882     & \U{0.801} & 0.464     & \U{0.690} & 3.33      & 17.73     & 0.565     & 0.440\\
3DGIC + i~\cite{huang20253d}
&   0.799   & \U{0.900} & \B{0.825} & 0.448     & \B{0.693} & \B{2.33}  & \U{18.02} & 0.566     & 0.403\\
ObjectClearer~\cite{zhao2026objectclear}
& \B{0.910} & \B{0.976} & 0.320     & \B{0.650} & 0.588     & \U{2.67}  & 16.26     & 0.504     & 0.432 \\
\bottomrule
\end{tabular}
}
\vspace{-15pt}
\end{table}

For Mip-NeRF360~\cite{barron2022mip} we provide a complementary reference-free evaluation setting.
Table~\ref{tab:mipnerf_results} shows that Gaussian Cut~\cite{jain2024gaussiancut} achieves the highest RQS and lowest AvgRank, indicating the strongest balance across semantic suppression, structural change, and geometric modification.
Gaussian Grouping~\cite{gaussian_grouping} also performs competitively, particularly in geometric and structural axes. 
Unlike Remove360, Mip-NeRF360~\cite{barron2022mip} cannot directly measure agreement with
the true post-removal scene, motivating the need for dedicated benchmarks such
as Remove360.
Detailed results are in Supp.~\ref{supp:results_mipnerf360}.

The two datasets place Gaussian Cut~\cite{jain2024gaussiancut} very differently, first on
Mip-NeRF360 but mid-field on Remove360. Supp.~\ref{supp:cross_dataset} explains why:
the structural axis is not the same quantity without a post-removal reference.

\begin{table}[t]
\centering
\scriptsize
\caption{
\textbf{Overall evaluation on Mip-NeRF360.}
Since Mip-NeRF360 does not provide physical post-removal references,
structural change is measured as $1-\mathrm{sim}_{SAM}$, where lower
pre/post-removal similarity indicates stronger structural change.
RQS is computed as the geometric mean of semantic suppression, geometric
change, and structural change. Lower average rank indicates more consistent ranking
across metrics.
}
\label{tab:mipnerf_results}
\vspace{-5pt}
\begin{tabular}{lcccccc}
\toprule
Method &
IoU$_{\mathrm{drop}}\uparrow$ &
acc$_{seg}\uparrow$ &
acc$_{\Delta depth}\uparrow$ &
$(1-\mathrm{sim}_{SAM})\uparrow$ &
RQS$\uparrow$ &
AvgRank$\downarrow$
\\
\midrule

Feature 3DGS~\cite{zhou2024feature}
&0.396      & 0.660     & 0.604     & 0.653     & 0.639     & 4.00\\
Gaussian Grouping~\cite{gaussian_grouping}
&0.585      & 0.849     & \U{0.841} & \B{0.853} & \U{0.848} & \U{2.00} \\
SAGS~\cite{hu2024semantic}
&0.346      & 0.557     & 0.503     & 0.541     & 0.533     & 5.00\\
Gaussian Cut~\cite{jain2024gaussiancut}
&\B{0.680}  & \U{0.914} & \B{0.896} & \U{0.845} & \B{0.885} & \B{1.67}\\
AuraFusion360~\cite{wu2025aurafusion}
&\U{0.615}  & \B{0.938} & 0.617     & 0.795     & 0.772     & 2.33\\
\bottomrule
\end{tabular}
\vspace{-10pt}
\end{table}

\PAR{Scene-Level Analysis.}
Table~\ref{tab:remove360_rqs_scene} reveals substantial variation across scenes.
Stroller, White Chairs, Playhouse, and Bicycle obtain relatively high RQS values for several methods, whereas Sofa and Pillows remain challenging.
The best-performing method also varies by scene: 3DGIC~\cite{huang20253d} performs strongly
on White Chairs and Office Chairs, and with inpainting on Playhouse, Table, and
Pillows. 
Gaussian Cut~\cite{jain2024gaussiancut} performs the best on Stroller and Bicycle.
These differences indicate that no removal strategy consistently dominates across object size, scene context, and interaction with surrounding geometry.

\begin{table}[t]
\centering
\scriptsize
\caption{
\textbf{Per-scene Removal Quality Score (RQS) on Remove360.}
Higher values indicate better removal quality.
The results show that removal difficulty varies substantially across object
categories and scene complexity. Full per-axis results are provided in the
supplementary material.
}
\vspace{-10pt}
\label{tab:remove360_rqs_scene}
\resizebox{\textwidth}{!}{
\begin{tabular}{lcccccc}
\toprule
Scene & GaussianCut~\cite{jain2024gaussiancut} & AuraFusion360~\cite{wu2025aurafusion} & +i~\cite{wu2025aurafusion} & 3DGIC~\cite{huang20253d} & +i~\cite{huang20253d} & ObjectClearer~\cite{zhao2026objectclear} \\
\midrule
Deckchair      &0.697     &0.689    &\B{0.779}  &0.739      &\U{0.745}  &0.468\\
White Chairs   &0.855     &0.744    &\U{0.866}  &\B{0.898}  &0.793      &0.612\\
Stroller       &\B{0.911} &0.807    &\U{0.883}  &0.867      &0.871      &0.718\\
Playhouse      &\U{0.767} &0.751    &0.561      &0.766      &\B{0.866}  &0.665\\
Toy Truck      &0.543     &0.498    &0.528      &\U{0.590}  &\B{0.605}  &0.490\\
Table          &0.643     &0.634    &0.634      &\U{0.646}  &\B{0.686}  &0.570\\
Pillows        &0.420     &0.432    &0.438      &0.427      &\U{0.515}  &\B{0.524}\\
Sofa           &\U{0.371} &0.369    &0.209      &0.233      &0.238      &\B{0.574}\\
Office Chairs  &0.626     &0.584    &0.630      &\B{0.741}  &\U{0.658}  &0.530\\
Bicycle        &\B{0.849} &0.727    &0.718      &\U{0.830}  &0.827      &0.655\\
Backpack       &0.625     &0.559    &\B{0.719}  &0.610      &\U{0.640}  &0.551\\
\bottomrule
\end{tabular}
}
\vspace{-15pt}
\end{table}
\PAR{Effect of Inpainting.}
Inpainting results are presented in Tables~\ref{tab:main_results}--~\ref{tab:remove360_rqs_scene}. 
It increases the aggregate RQS for both evaluated pipelines: AuraFusion360~\cite{wu2025aurafusion} increases from $0.630$ to $0.655$, while 3DGIC~\cite{huang20253d} increases from
$0.690$ to $0.693$.
For both methods, the main gain occurs in geometric modification.
However, structural agreement slightly decreases ($0.409\!\rightarrow\!0.407$ for AuraFusion360~\cite{wu2025aurafusion} and
$0.464\!\rightarrow\!0.448$ for 3DGIC~\cite{huang20253d}), showing that stronger geometric completion does not uniformly improve every evaluation axis.
These results are consistent with inpainting producing a mixture of useful background completion and additional object-like or inconsistent structures.

\PAR{Qualitative Results.}
Figs.~\ref{fig:new_groundedSam},~\ref{fig:new_sam}, and~\ref{fig:new_depth} visualize the category-level, structural, and geometric evaluation axes.
Fig.~\ref{fig:new_groundedSam} shows post-removal renderings in which the target object is no longer readily apparent, while the semantic probe~\cite{kirillov2023segany,liu2023grounding,ren2024grounded} still produces target-associated detections in the removed-object region.
Such responses may arise from residual geometry, boundaries, shadows, contextual structure, reconstruction errors, or inpainting artifacts.
Our evaluation measures their detectability without attributing them to a single cause.
Fig.~\ref{fig:new_sam} illustrates how successful removal changes the prompt-free SAM~\cite{ravi2024sam2} segmentation by revealing structures previously occluded by the \textit{sofa}.
Fig.~\ref{fig:new_depth} visualizes the corresponding depth changes within the target region.
The figures in the main paper deliberately show cases where evidence survives, since that is the phenomenon the benchmark measures, but they are not representative of every scene: removal succeeds on a substantial part of the benchmark, with the best per-scene RQS reaching $0.911$ on Stroller and $0.898$ on White Chairs, against $0.574$ on the hardest scene, Sofa (Tab.~\ref{tab:remove360_rqs_scene}).
To avoid a selection-biased impression, Supp.~\ref{supp:results_remove360_qualitative} provides qualitative grids for all 11 scenes, spanning the full difficulty range rather than the failure cases alone.
Additional qualitative and quantitative results are provided in Supp.~\ref{supp:results_remove360} and Supp.~\ref{supp:results_mipnerf360}.

\PAR{Pixel-Level Metrics.}
For Remove360, we additionally report PSNR, SSIM, and LPIPS against pose-matched reference renderings in Tab.~\ref{tab:main_results}.
These metrics measure image fidelity but do not directly evaluate recoverable object evidence.
Their rankings differ from those of the semantic, structural, and geometric axes: high image similarity may coexist with target-associated detections or weak structural and geometric scores.
Pixel-level metrics are therefore complementary, but insufficient on their own for evaluating object removal.

\PAR{Scope and Limitations.}
\emph{Probe dependence.}
Our metrics use off-the-shelf segmentation models as adversarial probes: they measure what current vision systems can recover from the shared representation, and are not intended as ground truth.
Scores are therefore relative to the probe family used, and we report them as such.
Establishing how far the rankings transfer across probe families is a natural next step.
For a privacy-motivated user the practical protocol is to run several probes and act on the most alarming.\footnote{This simulates an attacker who takes a manipulated reconstruction and applies multiple pre-trained models to probe it for residuals.}

\emph{Benchmark scale.}
With eleven scenes the benchmark resolves the large per-axis differences between methods, while the smaller aggregate gaps among the leading methods fall within its resolution (Supp.~\ref{supp:reliability}).
This is why we emphasise per-axis diagnosis over a single ranking.

\emph{Privacy scope.}
We measure whether the existence and category of a removed object remain detectable by automated probes.
Re-identifying a specific instance, recovering appearance or textual content, and human adversaries lie outside the present scope.

\section{Conclusion}
We introduced Remove360, a benchmark and evaluation protocol for assessing recoverable object evidence after removal in 3D Gaussian Splatting.
The protocol evaluates category-level suppression, prompt-free structural consistency, and geometric modification, while retaining the individual diagnostic metrics alongside the aggregate RQS.
Experiments reveal substantial trade-offs: methods that strongly suppress category-level detections may modify geometry only weakly, while methods with large geometric changes may still produce structural or target-associated evidence.
The physical post-removal captures in Remove360 enable these trade-offs to be evaluated in realistic full-scene settings.
Our results motivate future object-removal methods that jointly improve visual fidelity and reduce recoverable evidence in the final shared representation.

\section*{Acknowledgements}
This work was supported by 
the Czech Science Foundation (GACR) EXPRO (grant no. 23-07973X),
the Ministry of Education, Youth and Sports of the Czech Republic through the e-INFRA CZ (ID:90254),
the Grant Agency of the Czech Technical University in Prague, grant No. SGS25/060/OHK3/1T/13.

\bibliographystyle{splncs04}
\bibliography{main}

\clearpage 

\clearpage
\renewcommand{\thesection}{\Alph{section}}
\title{Remove360: Benchmarking Residuals After Object Removal in 3D Gaussian Splatting \\(Supplementary Material)}
\titlerunning{Supplementary Material}

\author{
  Simona Kocour\textsuperscript{1,2} \quad \quad Assia Benbihi\textsuperscript{1} \quad \quad Torsten Sattler\textsuperscript{1}\vspace{-10pt}}
\authorrunning{S.~Kocour et al.}
\institute{\textsuperscript{1}Czech Technical University in Prague, Czech Institute of Informatics, Robotics and Cybernetics \\
\textsuperscript{2}Czech Technical University in Prague, Faculty of Electrical Engineering}

\maketitle
\label{supp:material}

\newcommand{\suppTableStyle}{%
  \scriptsize
  \setlength{\tabcolsep}{3.2pt}%
  \renewcommand{\arraystretch}{0.92}%
}
\setlength{\textfloatsep}{7pt plus 1pt minus 2pt}
\setlength{\floatsep}{6pt plus 1pt minus 2pt}
\setlength{\intextsep}{6pt plus 1pt minus 2pt}

This supplementary provides implementation details and additional analyses supporting the main paper. Sec.~\ref{supp:implementation} documents the evaluated methods, Remove360 processing, annotations, and metric implementation. Sec.~\ref{supp:results_remove360} reports compact per-scene Remove360 comparisons, robustness analyses, and qualitative results. Sec.~\ref{supp:results_mipnerf360} gives the corresponding reference-free evaluation on Mip-NeRF360~\cite{barron2022mip}.

\section{Implementation Details}\label{supp:implementation}
\subsection{Evaluated Methods}\label{supp:methods}
We use the official implementations and released checkpoints of all methods and retain their default optimization settings unless stated otherwise. No hyperparameter is tuned per scene. All mask-based methods receive the same target-object masks within a dataset.

\paragraph{Feature 3DGS (FGS)~\cite{zhou2024feature}.}
Feature 3DGS associates distilled language-aligned features with individual Gaussians. We use one positive target prompt and scene-specific negative prompts, normalize similarities with a softmax, and remove Gaussians whose positive-prompt similarity exceeds 0.4. For Mip-NeRF360, the negative prompts are: Garden \{grass, sidewalk, tree, house\}; Room \{sofa, rug, television, floor\}; Kitchen \{rug, table, chair\}; and Counter \{oranges, wooden rolling pin, coconut oil\}.

\paragraph{SAGS~\cite{hu2024semantic}.}
SAGS is training-free and feature-free. It projects each Gaussian center into the input views and defines removal likelihood as the fraction of visible projections falling inside the target masks. Gaussians with likelihood above 0.7 are removed.

\paragraph{Gaussian Grouping~\cite{gaussian_grouping}.}
Gaussian Grouping assigns an instance label to every Gaussian using SAM features~\cite{kirillov2023segany}. We use a SAM IoU-prediction threshold of 0.8. The default training settings are \texttt{densify\_until\_iter=10000}, \texttt{num\_classes}=256, \texttt{reg3d\_interval}=5, \texttt{reg3d\_k}=5, \texttt{reg3d\_lambda\_val}=2, \texttt{reg3d\_max\_points}=200000, and \texttt{reg3d\_sample\_size}=1000. At test time, Gaussians assigned to the target label are removed with the default threshold 0.3.

\paragraph{Gaussian Cut~\cite{jain2024gaussiancut}.}
Gaussian Cut represents a trained 3DGS scene~\cite{kerbl3Dgaussians} as a graph. The unary term is initialized by lifting the target masks to 3D, and pairwise terms encode spatial and color similarity. Each Gaussian is connected to its 10 nearest neighbors. We use five source and five sink terminal clusters, a leaf size of 40, and a foreground threshold of 0.9.

\paragraph{AuraFusion360~\cite{wu2025aurafusion}.}
AuraFusion360 jointly fuses 2D masks with the Gaussian representation. Training runs for 20,000 iterations, and supervision masks are dilated with a kernel size of 10. A Gaussian is removed when its predicted removal confidence exceeds 0.6; the unseen-object threshold is 0.0. We evaluate both the removal-only output (AF) and the optional inpainting output (AF+i).

\paragraph{3DGIC~\cite{huang20253d}.}
3DGIC first removes Gaussians associated with the multi-view target masks and optionally performs depth-guided cross-view inpainting. We evaluate the removal stage alone (3DGIC) and the full pipeline (3DGIC+i), using the official implementation and default visibility settings.

\paragraph{ObjectClearer~\cite{zhao2026objectclear}.}
ObjectClearer is the 2D-first baseline. We apply the released ObjectClear model to the target-masked training images and reconstruct a new post-removal 3D scene from the processed views. We use the released checkpoint and the standard configuration in the official repository, without scene-specific tuning or changes to the default inference parameters. The reconstructed scene is evaluated with the same view list and semantic, structural, geometric, and image-fidelity protocol as the other Remove360 methods.

\paragraph{Method coverage.}
Nine pipelines were attempted in total. Table~\ref{tab:supp_coverage} in
Sec.~\ref{supp:coverage} documents each one, the datasets on which it produced
valid outputs, and the reason for every exclusion. Because the two datasets
carry different method sets, rankings should be interpreted within each dataset
rather than across datasets.

\subsection{Remove360 Processing}\label{supp:dataset_protocol}
\paragraph{Capture and camera poses.}
Each pre- and post-removal sequence is recorded at 4K/60 fps for approximately 4--6 minutes. We retain the sharpest frame in each one-second interval using variance of the Laplacian implemented in OpenCV~\cite{opencv_library}. Camera poses and sparse geometry are reconstructed with hLoc~\cite{sarlin2019coarse}. NetVLAD~\cite{Arandjelovic16} provides global retrieval, SuperPoint~\cite{detone2018superpoint} and LightGlue~\cite{lindenberger2023lightglue} provide local matching, and COLMAP~\cite{schoenberger2016sfm} performs structure from motion with the RADIAL camera model. Sequential pairs use an overlap of 10 frames with quadratic overlap enabled, and loop closure retrieves the top 20 NetVLAD~\cite{Arandjelovic16} neighbors every fifth frame. For each evaluation pose $v$, the post-removal reference reconstruction is rendered using the same camera intrinsics and extrinsics as the evaluated result, producing the reference rendering $I_{\mathrm{ref}}^v$.

\begin{table}[t]
\centering
\caption{\textbf{Remove360 scene inventory.} The benchmark contains six outdoor and five indoor scenes, each with 150--300 pre-removal views and a comparable physical post-removal capture.}
\label{tab:supp_scene_inventory}
\suppTableStyle
\begin{tabular}{lcc}\toprule
Scene / target & Setting & Removed instances\\\midrule
Deckchair & Outdoor & Single\\
White Chairs & Outdoor & Multiple\\
Stroller & Outdoor & Single\\
Playhouse & Outdoor & Single\\
Toy Truck & Outdoor & Single\\
Bicycle & Outdoor & Single\\
\addlinespace[2pt]
Table & Indoor & Single\\
Pillows & Indoor & Multiple\\
Sofa & Indoor & Single\\
Office Chairs & Indoor & Multiple\\
Backpack & Indoor & Single\\\bottomrule
\end{tabular}
\end{table}

\subsection{Object Masks}\label{supp:object_masks}
\paragraph{Remove360.}
Target masks are initialized with a Segment Anything model~\cite{kirillov2023segany,ravi2024sam2} and manually verified. Oversegmented regions are merged, and boundary pixels are corrected when necessary. Fewer than 10 images per scene required manual correction; final inspection found no missing target parts or unrelated included regions.

\paragraph{Mip-NeRF360.}
Mip-NeRF360~\cite{barron2022mip} does not provide target masks. We generate pseudo-ground-truth masks with a Segment Anything model~\cite{kirillov2023segany,ravi2024sam2}, select all segments belonging to the target object, merge overlapping segments when necessary, and exclude views with incomplete target coverage.

\subsection{Evaluation Details}\label{supp:metrics}
\paragraph{Semantic probe.}
We use GroundedSAM2~\cite{kirillov2023segany,liu2023grounding,ren2024grounded,ravi2024sam2} with its released inference configuration. The model receives the complete rendering and target category prompt. Ground-truth masks are used only for scoring. If no target mask is retained, the prediction is empty and its IoU is zero. The main evaluation uses $\xi_{\mathrm{IoU}}=0.3$.

\paragraph{Pre-removal detectability.}
A low post-removal response is difficult to interpret when the target was not localized in the original rendering. We therefore report $\mathrm{IoU}_{\mathrm{drop}}$ together with the separate pre- and post-removal target IoUs in Table~\ref{tab:supp_remove360_iou}. This exposes whether apparent suppression is caused by weak pre-removal detection rather than by the edit.

\paragraph{Prompt-free structural matching.}
Segment Anything masks~\cite{kirillov2023segany,ravi2024sam2} are matched one-to-one using the Hungarian algorithm~\cite{munkres1957algorithms}. If both retained mask sets are empty, similarity is 1; if exactly one set is empty, similarity is 0. Otherwise, unmatched masks are penalized through normalization by the larger set size, as defined in the main paper.

\paragraph{Depth threshold.}
The view-dependent depth-change threshold is estimated with Generalized Histogram Thresholding~\cite{BarronECCV2020} over the full rendered depth-difference image. The score measures modification relative to the pre-removal reconstruction, not correctness with respect to physical post-removal geometry.

\paragraph{Aggregation.}
For each scene, per-view values are averaged to obtain the reported scene-level axes, and scene-level RQS is their geometric mean. The \emph{Overall} rows use the view-weighted aggregation employed in the main paper: all valid evaluation views are pooled across scenes for each axis, after which RQS is computed from the three pooled axis scores. Consequently, the Overall entries are not arithmetic means of the rounded per-scene values. $\mathrm{IoU}_{\mathrm{drop}}$ and image-fidelity metrics remain diagnostic and do not enter RQS or AvgRank.

\section{Remove360 Results}\label{supp:results_remove360}
\subsection{Quantitative Results}\label{supp:remove360_quantitative}
This section reports every Remove360 scene on each evaluation axis. To make comparisons readable, scenes are rows and methods are columns; bold and underlined entries denote the best and second-best result within each scene. The Overall rows reproduce the main-paper values exactly.

\begin{table}[!htbp]
\centering
\caption{\textbf{Per-scene semantic evaluation on Remove360.} Panel (a) reports the decrease in target IoU, while panel (b) reports post-removal non-detection at $\xi_{\mathrm{IoU}}=0.3$. GC denotes Gaussian Cut~\cite{jain2024gaussiancut}, AF denotes AuraFusion360~\cite{wu2025aurafusion}, 3DGIC denotes 3DGIC~\cite{huang20253d}, and OC denotes ObjectClearer~\cite{zhao2026objectclear}. Bold and underlined values are best and second best within each scene and Overall row.}
\label{tab:supp_remove360_semantic}
\label{tab:supp_iou_drop}
\label{tab:supp_accseg}
\suppTableStyle
\begin{minipage}{0.99\textwidth}
\centering
\textbf{(a) $\mathrm{IoU}_{\mathrm{drop}}\uparrow$}\\[2pt]
\begin{tabular}{lcccccc}
\toprule
Scene & GC & AF & AF+i & 3DGIC & 3DGIC+i & OC \\
\midrule
\multicolumn{7}{l}{\itshape Outdoor scenes}\\[-2pt]
Deckchair & 0.850 & 0.840 & \U{0.870} & 0.830 & 0.850 & \B{0.980} \\
White Chairs & 0.850 & \U{0.870} & \U{0.870} & 0.830 & 0.850 & \B{0.950} \\
Stroller & \B{0.920} & \U{0.910} & \U{0.910} & 0.900 & 0.900 & \U{0.910} \\
Playhouse & \U{0.950} & \B{0.970} & 0.940 & 0.910 & \U{0.950} & 0.860 \\
Toy Truck & \U{0.950} & 0.930 & 0.930 & 0.710 & 0.720 & \B{0.960} \\
Bicycle & \B{0.950} & \B{0.950} & 0.900 & 0.880 & \U{0.930} & 0.840 \\
\addlinespace[2pt]
\multicolumn{7}{l}{\itshape Indoor scenes}\\[-2pt]
Table & \U{0.910} & \U{0.910} & 0.770 & 0.830 & 0.860 & \B{0.940} \\
Pillows & 0.620 & \U{0.760} & 0.750 & 0.510 & 0.610 & \B{0.940} \\
Sofa & 0.570 & \U{0.620} & 0.300 & 0.450 & 0.460 & \B{0.860} \\
Office Chairs & 0.690 & 0.640 & 0.690 & 0.770 & \U{0.790} & \B{0.880} \\
Backpack & \U{0.890} & 0.820 & 0.870 & 0.870 & 0.870 & \B{0.980} \\
\midrule
\textbf{Overall} & 0.832 & \U{0.838} & 0.800 & 0.772 & 0.799 & \B{0.910} \\
\bottomrule
\end{tabular}
\end{minipage}

\vspace{5pt}
\begin{minipage}{0.99\textwidth}
\centering
\textbf{(b) $\mathrm{acc}_{\mathrm{seg},0.3}\uparrow$}\\[2pt]
\begin{tabular}{lcccccc}
\toprule
Scene & GC & AF & AF+i & 3DGIC & 3DGIC+i & OC \\
\midrule
\multicolumn{7}{l}{\itshape Outdoor scenes}\\[-2pt]
Deckchair & 0.903 & 0.932 & \U{0.989} & 0.966 & 0.959 & \B{1.000} \\
White Chairs & \U{0.990} & \U{0.990} & \U{0.990} & 0.981 & 0.989 & \B{1.000} \\
Stroller & \B{1.000} & \B{1.000} & \B{1.000} & \B{1.000} & \B{1.000} & \U{0.980} \\
Playhouse & 0.980 & \U{0.995} & 0.989 & 0.956 & \B{1.000} & 0.930 \\
Toy Truck & 0.995 & 0.962 & \U{0.997} & \B{1.000} & \B{1.000} & 0.990 \\
Bicycle & 0.990 & \B{1.000} & 0.979 & 0.940 & \U{0.999} & 0.960 \\
\addlinespace[2pt]
\multicolumn{7}{l}{\itshape Indoor scenes}\\[-2pt]
Table & \U{0.973} & \B{1.000} & 0.838 & 0.952 & 0.949 & \B{1.000} \\
Pillows & 0.738 & 0.877 & \U{0.918} & 0.626 & 0.750 & \B{1.000} \\
Sofa & 0.483 & \U{0.625} & 0.190 & 0.352 & 0.351 & \B{0.890} \\
Office Chairs & 0.793 & 0.735 & 0.981 & 0.953 & \U{0.989} & \B{0.990} \\
Backpack & 0.904 & 0.727 & \U{0.962} & 0.818 & 0.799 & \B{1.000} \\
\midrule
\textbf{Overall} & 0.875 & 0.891 & 0.891 & 0.882 & \U{0.900} & \B{0.976} \\
\bottomrule
\end{tabular}
\end{minipage}
\end{table}

\begin{table}[!htbp]
\centering
\caption{\textbf{Per-scene geometric and structural evaluation on Remove360.} Panel (a) reports the fraction of target-region pixels with a depth change above the automatically selected threshold. Panel (b) reports category-prompt-free structural agreement with the pose-matched post-removal reference rendering. Method abbreviations and citations follow Table~\ref{tab:supp_remove360_semantic}. Bold and underlined values are best and second best.}
\label{tab:supp_remove360_geo_struct}
\label{tab:supp_depth}
\label{tab:supp_sam}
\suppTableStyle
\begin{minipage}{0.99\textwidth}
\centering
\textbf{(a) $\mathrm{acc}_{\Delta\mathrm{depth}}\uparrow$}\\[2pt]
\begin{tabular}{lcccccc}
\toprule
Scene & GC & AF & AF+i & 3DGIC & 3DGIC+i & OC \\
\midrule
\multicolumn{7}{l}{\itshape Outdoor scenes}\\[-2pt]
Deckchair & 0.670 & 0.650 & \B{0.810} & 0.760 & \U{0.770} & 0.130 \\
White Chairs & 0.760 & 0.670 & \U{0.820} & \B{0.840} & \B{0.840} & 0.290 \\
Stroller & \B{0.890} & 0.730 & 0.800 & \U{0.870} & \U{0.870} & 0.490 \\
Playhouse & \U{0.920} & 0.870 & 0.850 & \B{0.940} & \B{0.940} & 0.410 \\
Toy Truck & 0.730 & 0.640 & \B{0.820} & 0.760 & \U{0.790} & 0.350 \\
Bicycle & 0.910 & 0.800 & 0.840 & \U{0.950} & \B{0.960} & 0.390 \\
\addlinespace[2pt]
\multicolumn{7}{l}{\itshape Indoor scenes}\\[-2pt]
Table & 0.570 & 0.580 & \U{0.800} & 0.630 & \B{0.830} & 0.280 \\
Pillows & 0.530 & 0.510 & 0.610 & \U{0.620} & \B{0.650} & 0.240 \\
Sofa & 0.620 & 0.620 & 0.480 & \B{0.720} & \U{0.640} & 0.400 \\
Office Chairs & 0.910 & 0.820 & 0.850 & \U{0.950} & \B{0.960} & 0.320 \\
Backpack & 0.730 & 0.650 & \B{0.840} & 0.770 & \U{0.820} & 0.220 \\
\midrule
\textbf{Overall} & 0.749 & 0.685 & 0.775 & \U{0.801} & \B{0.825} & 0.320 \\
\bottomrule
\end{tabular}
\end{minipage}

\vspace{5pt}
\begin{minipage}{0.99\textwidth}
\centering
\textbf{(b) $\mathrm{sim}_{\mathrm{SAM}}\uparrow$}\\[2pt]
\begin{tabular}{lcccccc}
\toprule
Scene & GC & AF & AF+i & 3DGIC & 3DGIC+i & OC \\
\midrule
\multicolumn{7}{l}{\itshape Outdoor scenes}\\[-2pt]
Deckchair & 0.560 & 0.540 & \U{0.590} & 0.550 & 0.560 & \B{0.790} \\
White Chairs & \U{0.830} & 0.620 & 0.800 & \B{0.880} & 0.600 & 0.790 \\
Stroller & \U{0.850} & 0.720 & \B{0.860} & 0.750 & 0.760 & 0.770 \\
Playhouse & 0.500 & 0.490 & 0.210 & 0.500 & \U{0.690} & \B{0.770} \\
Toy Truck & 0.220 & 0.200 & 0.180 & 0.270 & \U{0.280} & \B{0.340} \\
Bicycle & \U{0.680} & 0.480 & 0.450 & 0.640 & 0.590 & \B{0.750} \\
\addlinespace[2pt]
\multicolumn{7}{l}{\itshape Indoor scenes}\\[-2pt]
Table & \U{0.480} & 0.440 & 0.380 & 0.450 & 0.410 & \B{0.660} \\
Pillows & 0.190 & 0.180 & 0.150 & 0.200 & \U{0.280} & \B{0.600} \\
Sofa & \U{0.170} & 0.130 & 0.100 & 0.050 & 0.060 & \B{0.530} \\
Office Chairs & 0.340 & 0.330 & 0.300 & \U{0.450} & 0.300 & \B{0.470} \\
Backpack & 0.370 & 0.370 & \U{0.460} & 0.360 & 0.400 & \B{0.760} \\
\midrule
\textbf{Overall} & \U{0.472} & 0.409 & 0.407 & 0.464 & 0.448 & \B{0.650} \\
\bottomrule
\end{tabular}
\end{minipage}
\end{table}

\begin{table}[!htbp]
\centering
\caption{\textbf{Per-scene Removal Quality Score (RQS) on Remove360.} RQS is the geometric mean of the semantic, geometric, and structural axes. Method abbreviations refer to Gaussian Cut~\cite{jain2024gaussiancut}, AuraFusion360~\cite{wu2025aurafusion}, 3DGIC~\cite{huang20253d}, and ObjectClearer~\cite{zhao2026objectclear}. Bold and underlined values are best and second best.}
\label{tab:supp_rqs}
\suppTableStyle
\begin{tabular}{lcccccc}
\toprule
Scene & GC & AF & AF+i & 3DGIC & 3DGIC+i & OC \\
\midrule
\multicolumn{7}{l}{\itshape Outdoor scenes}\\[-2pt]
Deckchair & 0.697 & 0.689 & \B{0.779} & 0.739 & \U{0.745} & 0.468 \\
White Chairs & 0.855 & 0.744 & \U{0.866} & \B{0.898} & 0.793 & 0.612 \\
Stroller & \B{0.911} & 0.807 & \U{0.883} & 0.867 & 0.871 & 0.718 \\
Playhouse & \U{0.767} & 0.751 & 0.561 & 0.766 & \B{0.866} & 0.665 \\
Toy Truck & 0.543 & 0.498 & 0.528 & \U{0.590} & \B{0.605} & 0.490 \\
Bicycle & \B{0.849} & 0.727 & 0.718 & \U{0.830} & 0.827 & 0.655 \\
\addlinespace[2pt]
\multicolumn{7}{l}{\itshape Indoor scenes}\\[-2pt]
Table & 0.643 & 0.634 & 0.634 & \U{0.646} & \B{0.686} & 0.570 \\
Pillows & 0.420 & 0.432 & 0.438 & 0.427 & \U{0.515} & \B{0.524} \\
Sofa & \U{0.371} & 0.369 & 0.209 & 0.233 & 0.238 & \B{0.574} \\
Office Chairs & 0.626 & 0.584 & 0.630 & \B{0.741} & \U{0.658} & 0.530 \\
Backpack & 0.625 & 0.559 & \B{0.719} & 0.610 & \U{0.640} & 0.551 \\
\midrule
\textbf{Overall} & 0.676 & 0.630 & 0.655 & \U{0.690} & \B{0.693} & 0.588 \\
\bottomrule
\end{tabular}
\end{table}

\begin{table}[!htbp]
\centering
\caption{\textbf{Per-scene image fidelity on Remove360.} Each entry reports PSNR/SSIM/LPIPS against the pose-matched rendering of the post-removal reference reconstruction. Higher PSNR and SSIM and lower LPIPS are better; bold and underlined components denote the best and second-best values for that scene and metric. GC, AF, and 3DGIC denote Gaussian Cut~\cite{jain2024gaussiancut}, AuraFusion360~\cite{wu2025aurafusion}, and 3DGIC~\cite{huang20253d}.}
\label{tab:supp_remove360_fidelity}
\label{tab:supp_psnr}
\label{tab:supp_ssim}
\label{tab:supp_lpips}
\suppTableStyle
\setlength{\tabcolsep}{2.3pt}
\resizebox{\linewidth}{!}{%
\begin{tabular}{lccccc}
\toprule
Scene & GC & AF & AF+i & 3DGIC & 3DGIC+i \\
\midrule
\multicolumn{6}{l}{\itshape Outdoor scenes}\\[-2pt]
Deckchair    & 15.62/.255/\B{.427} & 15.86/\B{.275}/\U{.488} & \B{16.06}/\U{.273}/.538 & \U{16.02}/.271/.540 & 15.85/.253/.507 \\
White Chairs & 17.99/\B{.397}/\B{.389} & 17.89/\B{.397}/.479 & \B{18.14}/\U{.391}/.518 & 17.95/\B{.397}/.505 & \U{18.13}/\B{.397}/\U{.450} \\
Stroller     & 19.19/.587/\B{.207} & 18.74/.590/\U{.250} & \U{19.22}/\B{.600}/.368 & 18.86/\B{.600}/.361 & \B{19.34}/\U{.593}/.342 \\
Playhouse    & 18.05/.387/\B{.396} & \U{18.07}/\B{.389}/.483 & 15.61/.352/.587 & 18.05/\U{.388}/.499 & \B{18.43}/\B{.389}/\U{.434} \\
Toy Truck    & 15.62/.376/\B{.359} & 15.66/\U{.388}/\U{.404} & 15.63/\U{.388}/.477 & \U{15.87}/\B{.396}/.461 & \B{16.11}/\U{.388}/.438 \\
Bicycle      & 17.00/.425/\B{.372} & 16.61/.430/\U{.441} & \B{17.91}/\U{.432}/.481 & 16.59/.430/.500 & \U{17.37}/\B{.437}/.451 \\
\addlinespace[2pt]
\multicolumn{6}{l}{\itshape Indoor scenes}\\[-2pt]
Table         & \U{21.76}/\U{.849}/\B{.170} & \B{21.92}/.848/\U{.204} & 21.60/\B{.851}/.316 & 20.98/.843/.308 & 20.85/.842/.270 \\
Pillows       & \B{21.45}/\B{.822}/\B{.252} & \U{20.41}/.793/\U{.298} & 11.76/.643/.636 & 19.47/.761/.432 & 19.88/\U{.800}/.381 \\
Sofa          & \B{17.45}/\B{.785}/\B{.275} & \U{16.81}/.770/\U{.313} & 15.95/\U{.776}/.395 & 15.40/.749/.393 & 16.03/.766/.360 \\
Office Chairs & \B{17.27}/\B{.764}/\B{.296} & 15.93/\U{.736}/\U{.395} & \U{16.09}/.731/.481 & 15.93/\U{.736}/.445 & 16.00/.724/.443 \\
Backpack      & 19.71/\U{.648}/\B{.272} & 19.49/\B{.655}/\U{.334} & \U{20.00}/.643/.402 & 19.86/.644/.400 & \B{20.23}/.642/.355 \\
\midrule
\textbf{Overall} & \B{18.28}/\B{.572}/\B{.310} & 17.94/\U{.570}/\U{.372} & 17.09/.553/.473 & 17.73/.565/.440 & \U{18.02}/.566/.403 \\
\bottomrule
\end{tabular}}
\end{table}

\begin{table}[!htbp]
\centering
\caption{\textbf{GroundedSAM2~\cite{kirillov2023segany,liu2023grounding,ren2024grounded,ravi2024sam2} target IoU before and after removal on Remove360.} Each cell reports $\mathrm{mIoU}_{\mathrm{pre}}\!\rightarrow\!\mathrm{mIoU}_{\mathrm{post}}$ with $\mathrm{IoU}_{\mathrm{drop}}$ in brackets. Bold and underlined cells have the largest and second-largest drop for that scene.}
\label{tab:supp_remove360_iou}
\suppTableStyle
\setlength{\tabcolsep}{2.7pt}
\resizebox{\textwidth}{!}{%
\begin{tabular}{lcccc}
\toprule
Scene & GC~\cite{jain2024gaussiancut} & AF~\cite{wu2025aurafusion} & 3DGIC~\cite{huang20253d} & OC~\cite{zhao2026objectclear} \\
\midrule
Deckchair & \U{0.90$\rightarrow$0.05 [0.85]} & 0.88$\rightarrow$0.04 [0.84] & 0.86$\rightarrow$0.03 [0.83] & \B{1.00$\rightarrow$0.02 [0.98]} \\
White Chairs & 0.89$\rightarrow$0.04 [0.85] & \U{0.89$\rightarrow$0.02 [0.87]} & 0.85$\rightarrow$0.02 [0.83] & \B{1.00$\rightarrow$0.05 [0.95]} \\
Stroller & \B{0.92$\rightarrow$0.00 [0.92]} & \U{0.91$\rightarrow$0.00 [0.91]} & 0.90$\rightarrow$0.00 [0.90] & \U{1.00$\rightarrow$0.09 [0.91]} \\
Playhouse & \U{0.97$\rightarrow$0.03 [0.95]} & \B{0.99$\rightarrow$0.02 [0.97]} & 0.95$\rightarrow$0.03 [0.91] & 1.00$\rightarrow$0.14 [0.86] \\
Toy Truck & \U{0.95$\rightarrow$0.00 [0.95]} & 0.95$\rightarrow$0.02 [0.93] & 0.72$\rightarrow$0.01 [0.71] & \B{1.00$\rightarrow$0.04 [0.96]} \\
Table & \U{0.93$\rightarrow$0.02 [0.91]} & \U{0.92$\rightarrow$0.01 [0.91]} & 0.88$\rightarrow$0.05 [0.83] & \B{1.00$\rightarrow$0.06 [0.94]} \\
Pillows & 0.90$\rightarrow$0.28 [0.62] & \U{0.89$\rightarrow$0.13 [0.76]} & 0.79$\rightarrow$0.28 [0.51] & \B{1.00$\rightarrow$0.06 [0.94]} \\
Sofa & 0.96$\rightarrow$0.39 [0.57] & \U{0.95$\rightarrow$0.33 [0.62]} & 0.95$\rightarrow$0.50 [0.45] & \B{1.00$\rightarrow$0.14 [0.86]} \\
Office Chairs & 0.85$\rightarrow$0.18 [0.67] & 0.83$\rightarrow$0.19 [0.64] & \U{0.80$\rightarrow$0.03 [0.77]} & \B{1.00$\rightarrow$0.12 [0.88]} \\
Bicycle & \B{0.97$\rightarrow$0.02 [0.95]} & \B{0.95$\rightarrow$0.00 [0.95]} & \U{0.94$\rightarrow$0.06 [0.88]} & 0.90$\rightarrow$0.07 [0.84] \\
Backpack & \U{0.94$\rightarrow$0.05 [0.89]} & 0.96$\rightarrow$0.14 [0.82] & 0.97$\rightarrow$0.10 [0.87] & \B{1.00$\rightarrow$0.02 [0.98]} \\
\bottomrule
\end{tabular}}
\end{table}

\begin{table}[!htbp]
\centering
\caption{\textbf{Sensitivity of category-level suppression to the IoU threshold on Remove360.} Values are scene-macro averages of $\mathrm{acc}_{\mathrm{seg},\xi}$ over the 11 scenes. The main paper uses $\xi_{\mathrm{IoU}}=0.3$. The ranking is stable over the tested thresholds, although high thresholds saturate and are therefore less discriminative.}
\label{tab:supp_remove360_thresholds}
\suppTableStyle
\setlength{\tabcolsep}{5pt}
\begin{tabular}{lcccc}
\toprule
Method & $\xi_{\mathrm{IoU}}=0.3$ & $0.5$ & $0.7$ & $0.9$ \\
\midrule
GC~\cite{jain2024gaussiancut} & 0.886 & 0.909 & 0.946 & 0.985 \\
AF~\cite{wu2025aurafusion} & 0.895 & 0.918 & 0.944 & 0.970 \\
3DGIC~\cite{huang20253d} & 0.868 & 0.912 & 0.943 & 0.985 \\
OC~\cite{zhao2026objectclear} & 0.976 & 1.000 & 1.000 & 1.000 \\
\bottomrule
\end{tabular}
\end{table}

\subsection{Mask-Boundary Sensitivity}\label{supp:boundary_sensitivity}
We erode and dilate each target mask by 5, 10, and 15 pixels using an elliptical OpenCV kernel~\cite{opencv_library}. Erosion focuses on the object core; dilation includes surrounding context, shadows, and unmodified pixels. Table~\ref{tab:supp_boundary_compact} reports scene-macro averages to avoid repeating 231 closely related per-scene entries. The results change gradually rather than abruptly, showing that exact boundary placement affects absolute scores but does not explain the overall findings.

\begin{table}[!htbp]
\centering
\caption{\textbf{Mask-boundary sensitivity on Remove360.} Scene-macro averages are reported after erosion (E) or dilation (D) by 5, 10, and 15 pixels using an elliptical OpenCV kernel~\cite{opencv_library}. Bold marks the unperturbed result, not the best value. Changes are gradual, indicating that the principal comparisons are not caused by small annotation-boundary perturbations.}
\label{tab:supp_boundary_compact}
\label{tab:supp_mask_iou}
\label{tab:supp_mask_depth}
\label{tab:supp_mask_sam}
\suppTableStyle
\setlength{\tabcolsep}{2.6pt}
\begin{tabular}{llccccccc}
\toprule
Metric & Method & E15 & E10 & E5 & Original & D5 & D10 & D15 \\
\midrule
\multirow{3}{*}{$\mathrm{IoU}_{\mathrm{drop}}\uparrow$}
 & GC~\cite{jain2024gaussiancut} & .854 & .852 & .848 & \textbf{.832} & .792 & .746 & .705 \\
 & AF~\cite{wu2025aurafusion} & .863 & .860 & .855 & \textbf{.838} & .797 & .752 & .709 \\
 & 3DGIC~\cite{huang20253d} & .816 & .805 & .785 & \textbf{.772} & .658 & .592 & .543 \\
\addlinespace[2pt]
\multirow{3}{*}{$\mathrm{acc}_{\Delta\mathrm{depth}}\uparrow$}
 & GC~\cite{jain2024gaussiancut} & .756 & .755 & .751 & \textbf{.749} & .742 & .735 & .730 \\
 & AF~\cite{wu2025aurafusion} & .689 & .688 & .686 & \textbf{.685} & .679 & .659 & .639 \\
 & 3DGIC~\cite{huang20253d} & .850 & .832 & .823 & \textbf{.801} & .727 & .712 & .702 \\
\addlinespace[2pt]
\multirow{3}{*}{$\mathrm{sim}_{\mathrm{SAM}}\uparrow$}
 & GC~\cite{jain2024gaussiancut} & .475 & .474 & .473 & \textbf{.472} & .472 & .471 & .474 \\
 & AF~\cite{wu2025aurafusion} & .415 & .414 & .412 & \textbf{.409} & .405 & .403 & .404 \\
 & 3DGIC~\cite{huang20253d} & .489 & .484 & .475 & \textbf{.464} & .457 & .447 & .446 \\
\bottomrule
\end{tabular}
\end{table}

\subsection{Are the Three Axes Redundant?}\label{supp:axis_independence}
Combining three axes is only meaningful if they measure different things. We
test this on the 66 (scene, method) cells of Remove360. Part of any correlation
is scene difficulty -- a hard scene such as Sofa is hard on every axis -- so
Table~\ref{tab:supp_axis_corr} reports the raw coefficient, the coefficient
after centring within each scene, and the within-scene rank agreement over the
six methods.

\begin{table}[!htbp]
\centering
\caption{\textbf{Association between the three evaluation axes on Remove360.}
Spearman $\rho$ over the 66 (scene, method) cells, the same coefficient after
centring each axis within a scene, and the mean within-scene Kendall $\tau$
over the six methods ($\pm$ standard deviation over the 11 scenes).
The raw coefficient deliberately contains the shared scene effect, so no
$p$-value is quoted for it; the scene-centred $p$-values are two-sided
permutation tests with 5000 permutations of the method labels \emph{within}
each scene, which preserves the clustering of the observations.
Values near zero indicate that the axes order methods independently.}
\label{tab:supp_axis_corr}
\suppTableStyle
\setlength{\tabcolsep}{4pt}
\begin{tabular}{lccc}
\toprule
Axis pair & $\rho$ (raw) & $\rho$ (scene-centred) & mean $\tau$ (within scene)\\
\midrule
$s_{\mathrm{sem}}$ vs.\ $s_{\mathrm{geo}}$ & $0.117$ & $-0.130$~($p{=}0.32$) & $-0.07 \pm 0.45$\\
$s_{\mathrm{sem}}$ vs.\ $s_{\mathrm{str}}$ & $0.573$ & $0.299$~($p{=}0.028$) & $0.15 \pm 0.35$\\
$s_{\mathrm{geo}}$ vs.\ $s_{\mathrm{str}}$ & $0.089$ & $-0.222$~($p{=}0.084$) & $-0.13 \pm 0.32$\\
\bottomrule
\end{tabular}
\end{table}

The geometric axis is unrelated to the other two. The semantic and structural
axes correlate moderately in raw form ($\rho = 0.573$), but most of that is
shared scene difficulty: centring within scenes drops it to $0.299$, and the
mean within-scene rank agreement is only $\tau = 0.15$. No pair comes close to
being interchangeable, which is what justifies reporting the axes separately and
combining them with a penalising geometric mean.

\paragraph{The two semantic metrics.}
$\mathrm{IoU}_{\mathrm{drop}}$ and $\mathrm{acc}_{\mathrm{seg}}$ are more closely
related to each other than any pair of axes ($\rho = 0.717$ over the 66 cells,
$0.722$ after centring). That is one reason $\mathrm{IoU}_{\mathrm{drop}}$ stays
a diagnostic and is excluded from RQS and AvgRank. They are still not
interchangeable: $\mathrm{IoU}_{\mathrm{drop}}$ cannot separate a genuinely
suppressed detection from an object that was never detected before the edit.
Table~\ref{tab:supp_remove360_iou} reports both IoUs so this stays visible.

\subsection{Average versus Median Rank}\label{supp:rank_aggregation}
Ranks are discrete, so a median is the usual robust alternative to the mean in
Eq.~\eqref{eq:avg_rank}. Table~\ref{tab:supp_rank_agg} reports both. They
disagree, and the disagreement is the point.

\begin{table}[!htbp]
\centering
\caption{\textbf{Rank aggregation on Remove360.} Ranks of the six methods on the
three dataset-level RQS axes (rank~1 is best, ties receive the average rank),
summarised by mean, median, and worst single-axis rank.
Median rank promotes ObjectClearer to first place despite its last-place
geometric score, because a median of three ranks discards the failing axis.}
\label{tab:supp_rank_agg}
\suppTableStyle
\setlength{\tabcolsep}{4pt}
\begin{tabular}{lcccccc}
\toprule
Method & $s_{\mathrm{sem}}$ & $s_{\mathrm{geo}}$ & $s_{\mathrm{str}}$ & AvgRank$\downarrow$ & MedRank$\downarrow$ & worst axis$\downarrow$\\
\midrule
GC~\cite{jain2024gaussiancut}      & 6.0 & 4.0 & 2.0 & 4.00 & 4.0 & 6.0\\
AF~\cite{wu2025aurafusion}         & 3.5 & 5.0 & 5.0 & 4.50 & 5.0 & 5.0\\
AF+i~\cite{wu2025aurafusion}       & 3.5 & 3.0 & 6.0 & 4.17 & 3.5 & 6.0\\
3DGIC~\cite{huang20253d}           & 5.0 & 2.0 & 3.0 & 3.33 & 3.0 & 5.0\\
3DGIC+i~\cite{huang20253d}         & 2.0 & 1.0 & 4.0 & \B{2.33} & 2.0 & \B{4.0}\\
OC~\cite{zhao2026objectclear}      & 1.0 & 6.0 & 1.0 & 2.67 & \B{1.0} & 6.0\\
\bottomrule
\end{tabular}
\end{table}

With three axes the median is the middle rank, so it cannot see the worst one.
ObjectClearer ranks first on two axes and last on the third, scoring $0.320$
against $0.825$: it barely changes the geometry. Median rank calls it the best
of the six; average rank puts it second, and RQS -- which multiplies the axes --
puts it last. For a protocol whose purpose is to catch failure on any single
evidence channel, a summary that discards the worst axis is the wrong statistic.
We therefore keep average rank and add the worst single-axis rank as the
conservative reading. The same argument motivates the geometric mean in
Eq.~\eqref{eq:rqs}.

\subsection{How Much of the Ranking Is Significant?}\label{supp:reliability}
Per-scene RQS varies widely (Table~\ref{tab:supp_rqs}), so we ask how much of
the ordering in Table~\ref{tab:main_results} eleven scenes can support.
Table~\ref{tab:supp_reliability} reports the scene-macro mean RQS with a 95\%
confidence interval and a paired scene-level bootstrap over 20\,000 resamples.

\begin{table}[!htbp]
\centering
\caption{\textbf{Reliability of the Remove360 ranking across scenes.}
Scene-macro mean RQS over the 11 scenes with standard deviation, range, and
95\% confidence interval, and the paired bootstrap probability that a method
outranks 3DGIC+i. Note that the scene-macro mean differs slightly from the
view-pooled Overall row of Table~\ref{tab:main_results}, which weights scenes by
their number of views.}
\label{tab:supp_reliability}
\suppTableStyle
\setlength{\tabcolsep}{4pt}
\begin{tabular}{lccccc}
\toprule
Method & mean RQS & sd & range & 95\% CI & $P(\text{beats 3DGIC+i})$\\
\midrule
GC~\cite{jain2024gaussiancut}      & 0.664 & 0.175 & 0.371--0.911 & [0.561, 0.768] & 0.266\\
AF~\cite{wu2025aurafusion}         & 0.618 & 0.142 & 0.369--0.807 & [0.534, 0.702] & 0.006\\
AF+i~\cite{wu2025aurafusion}       & 0.633 & 0.197 & 0.209--0.883 & [0.517, 0.749] & 0.064\\
3DGIC~\cite{huang20253d}           & 0.668 & 0.199 & 0.233--0.898 & [0.550, 0.786] & 0.304\\
3DGIC+i~\cite{huang20253d}         & \B{0.677} & 0.184 & 0.238--0.871 & [0.568, 0.786] & --\\
OC~\cite{zhao2026objectclear}      & 0.578 & 0.077 & 0.468--0.718 & [0.532, 0.624] & 0.026\\
\bottomrule
\end{tabular}
\end{table}

The intervals of the three leading methods overlap almost entirely, and the
bootstrap makes the consequence concrete: Gaussian Cut and 3DGIC beat 3DGIC+i on
$27\%$ and $30\%$ of resampled scene sets. At this scale the benchmark does not
order GC, 3DGIC and 3DGIC+i, and the differences between them should be read as
ties. What it does resolve are the large per-axis gaps -- ObjectClearer sits more
than $0.35$ below every 3DGS pipeline on the geometric axis. This is why we
present Remove360 as a diagnostic for per-axis failure rather than a
leaderboard, and why more scenes are the direct route to sharper aggregate
comparisons.

\subsection{Metric Design Choices}\label{supp:metric_alternatives}
Several parts of the protocol admit simpler forms. We record why the reported
one was chosen.

\paragraph{Why not post-removal IoU alone.}
Averaging $\mathrm{IoU}(\hat M^v_{\mathrm{post}}, M^v)$ is simpler, and it is the
quantity behind $\mathrm{acc}_{\mathrm{seg}}$. We threshold rather than average
because most views yield an empty or near-empty prediction, so the mean sits
near zero for every method and separates them poorly
(Table~\ref{tab:supp_remove360_iou}). Thresholding asks the operational
question -- in what fraction of shared views is the object still localisable? --
and spans a usable range.

\paragraph{The $\max(N_v, M_v)$ normalisation.}
This penalises a difference in the number of retained masks as well as their
overlap, and its effect is not symmetric across the two settings. With a
reference, $\mathrm{sim}_{\mathrm{SAM}}(I_{\mathrm{post}}, I_{\mathrm{ref}})$ is
maximised by reproducing the reference mask set, so fragmenting the region and
collapsing it are both penalised correctly. Reference-free we report
$1 - \mathrm{sim}_{\mathrm{SAM}}(I_{\mathrm{pre}}, I_{\mathrm{post}})$, where the
same term credits \emph{any} change in mask count, including fragmentation from
inpainting. The reference-free axis therefore measures change rather than
correctness, which is why we use it only on Mip-NeRF360.

\paragraph{Why not SSIM between segmentation maps.}
Prompt-free SAM returns an unordered set of binary masks, not an image.
Rendering it as a label map makes SSIM~\cite{wang2004image} depend on arbitrary
mask indices, so a relabelling that leaves the segmentation identical can change
the score. The Hungarian matching in Eq.~\eqref{eq:sim_sam_matching} removes that
dependence. SSIM would also reward texture agreement inside the masks, which the
pixel-level metrics already cover.

\paragraph{The adaptive depth threshold.}
Where a physical reference exists, comparing the edited depth against the
reference depth is simpler and better posed, since it asks whether the geometry
became \emph{correct} rather than merely \emph{different}. We keep the adaptive
threshold for two reasons. It is the only form that also works without
post-removal captures, as on Mip-NeRF360. And it compares two renderings of the
same reconstruction lineage at one pose, so it is unaffected by the registration
error between the independently posed pre- and post-removal sequences.

\paragraph{Notation.}
Table~\ref{tab:supp_notation} collects the symbols used in Sec.~\ref{sec:metrics}.

\begin{table}[!htbp]
\centering
\caption{\textbf{Notation used in the evaluation protocol} (Sec.~\ref{sec:metrics}).}
\label{tab:supp_notation}
\suppTableStyle
\setlength{\tabcolsep}{4pt}
\begin{tabular}{ll}
\toprule
Symbol & Meaning\\
\midrule
$\mathcal V$, $v$ & set of evaluation views; a single view\\
$I^v_{\mathrm{pre}}$, $I^v_{\mathrm{post}}$ & rendering before / after the evaluated edit\\
$I^v_{\mathrm{ref}}$ & rendering of the physical post-removal reconstruction at pose $v$\\
$D^v_{\mathrm{pre}}$, $D^v_{\mathrm{post}}$ & rendered depth before / after the edit\\
$M^v$ & ground-truth target-object mask (scoring region only)\\
$\hat M^v_{\mathrm{pre}}$, $\hat M^v_{\mathrm{post}}$ & semantic-probe prediction before / after the edit\\
$\mathcal A^v$, $\mathcal B^v$ & the two retained prompt-free mask sets, of sizes $N_v$, $M_v$\\
$\xi_{\mathrm{IoU}}$, $\xi^v_{\mathrm{depth}}$ & IoU detection threshold ($0.3$); per-view depth threshold\\
$s_{\mathrm{sem}}$, $s_{\mathrm{geo}}$, $s_{\mathrm{str}}$ & the three oriented axes in $[0,1]$, higher is better\\
\bottomrule
\end{tabular}
\end{table}

\subsection{Method Coverage}\label{supp:coverage}
Table~\ref{tab:supp_coverage} lists all nine pipelines we attempted, where each
produced valid output, and why any was excluded.

\begin{table}[!htbp]
\centering
\caption{\textbf{Method coverage.} \cmark: evaluated and reported.
\xmark: attempted but excluded, with the reason. --: not attempted for that
dataset. The two datasets carry different method sets, so rankings should be
read within a dataset.}
\label{tab:supp_coverage}
\suppTableStyle
\resizebox{\textwidth}{!}{
\begin{tabular}{llccl}
\toprule
Method & Type & Remove360 & Mip-NeRF360 & Outcome / reason for exclusion\\
\midrule
GC~\cite{jain2024gaussiancut}   & graph-cut       & \cmark & \cmark & --\\
AF~\cite{wu2025aurafusion}     & mask fusion     & \cmark & \cmark & --\\
AF+i~\cite{wu2025aurafusion}   & \quad+ inpainting  & \cmark & --     & not included in the earlier Mip-NeRF360 runs\\
3DGIC~\cite{huang20253d}                  & mask lifting    & \cmark & --     & not included in the earlier Mip-NeRF360 runs\\
3DGIC+i~\cite{huang20253d}                & \quad+ inpainting  & \cmark & --     & not included in the earlier Mip-NeRF360 runs\\
ObjectClearer~\cite{zhao2026objectclear}  & 2D-first        & \cmark & --     & not included in the earlier Mip-NeRF360 runs\\
\addlinespace[2pt]
GG~\cite{gaussian_grouping}& instance labels & \xmark & \cmark & no valid removal output on Remove360\\
Feature 3DGS~\cite{zhou2024feature}       & distilled feat. & \xmark & \cmark & no valid removal output on Remove360\\
SAGS~\cite{hu2024semantic}                & projection      & \xmark & \cmark & no valid removal output on Remove360\\
\bottomrule
\end{tabular}
}
\end{table}

The three excluded pipelines all identify the Gaussians to remove through a
per-object semantic or instance assignment. All three work on the object-centric
Mip-NeRF360 targets, and none produced a usable removal on the cluttered
Remove360 scenes. We report this as a property of the benchmark rather than a
defect of the methods: clutter is the condition Remove360 exists to test. Every
method ran with its released configuration and was not tuned per scene, so these
are out-of-the-box outcomes.

Two categories stay outside the evaluation. Diffusion-based
repair~\cite{liu2024infusion,shi2025imfine,cheng2025painpainter,pan2025diga3d} is
the most relevant gap, since a strong generative prior is exactly what could make
an edit look plausible while introducing new recoverable evidence; we cover only
the inpainting stages shipped with AuraFusion360 and 3DGIC. Methods that expose
no per-view depth cannot be scored on the geometric axis without modification.

\subsection{Target Scale and the IoU Threshold}\label{supp:object_scale}
A fixed $\xi_{\mathrm{IoU}} = 0.3$ is not equally strict for large and small
targets, since IoU is harder to reach on a small object.
Table~\ref{tab:supp_object_scale} sets target area against difficulty.

\begin{table}[!htbp]
\centering
\caption{\textbf{Target scale versus removal difficulty on Remove360.}
Target-mask area as a percentage of image area, averaged and median over the
evaluation views of each scene, together with the scene-mean RQS over the six
methods and its spread across methods. Target scale spans more than an order of
magnitude but is uncorrelated with difficulty (Spearman $\rho = -0.05$,
$p = 0.90$ between median area and scene-mean RQS).}
\label{tab:supp_object_scale}
\suppTableStyle
\setlength{\tabcolsep}{5pt}
\begin{tabular}{lcccc}
\toprule
Scene & mean area (\%) & median area (\%) & scene RQS & sd over methods\\
\midrule
\multicolumn{5}{l}{\itshape Outdoor scenes}\\[-2pt]
Deckchair      & 1.7  & 1.0  & 0.686 & 0.112\\
White Chairs   & 4.7  & 2.6  & 0.795 & 0.105\\
Stroller       & 9.2  & 6.2  & 0.843 & 0.070\\
Playhouse      & 14.8 & 11.7 & 0.729 & 0.104\\
Toy Truck      & 4.7  & 2.5  & 0.542 & 0.047\\
Bicycle        & 7.0  & 4.4  & 0.768 & 0.079\\
\addlinespace[2pt]
\multicolumn{5}{l}{\itshape Indoor scenes}\\[-2pt]
Table          & 6.1  & 5.5  & 0.636 & 0.038\\
Pillows        & 5.6  & 4.1  & 0.459 & 0.047\\
Sofa           & 14.9 & 13.2 & 0.332 & 0.138\\
Office Chairs  & 13.8 & 12.0 & 0.628 & 0.071\\
Backpack       & 2.0  & 1.5  & 0.617 & 0.061\\
\bottomrule
\end{tabular}
\end{table}

Median target area spans $1.0\%$ to $13.2\%$ of the image but does not predict
difficulty ($\rho = -0.05$). The individual axes relate to scale weakly and in
opposite directions ($-0.21$ semantic, $-0.22$ structural, $+0.23$ geometric):
a larger target is slightly easier to modify geometrically and slightly harder to
suppress semantically. Because the axes pull in opposite directions, a fixed
threshold introduces no systematic scale bias into RQS. Difficulty comes from
contact geometry and layout instead -- Sofa, the hardest scene, has the largest
target, and is hard because removing it exposes a large previously occluded
background. We keep $\xi_{\mathrm{IoU}} = 0.3$ as the most discriminative of the
tested values; Table~\ref{tab:supp_remove360_thresholds} shows every method
approaching saturation by $0.9$.

\subsection{Qualitative Results}\label{supp:results_remove360_qualitative}
The qualitative grids complement the per-scene tables by jointly showing RGB appearance, category-level responses, prompt-free structure, and depth changes. They do not assign a single causal source to a probe response.

\begin{figure*}[t]
\centering
\includegraphics[width=0.98\linewidth]{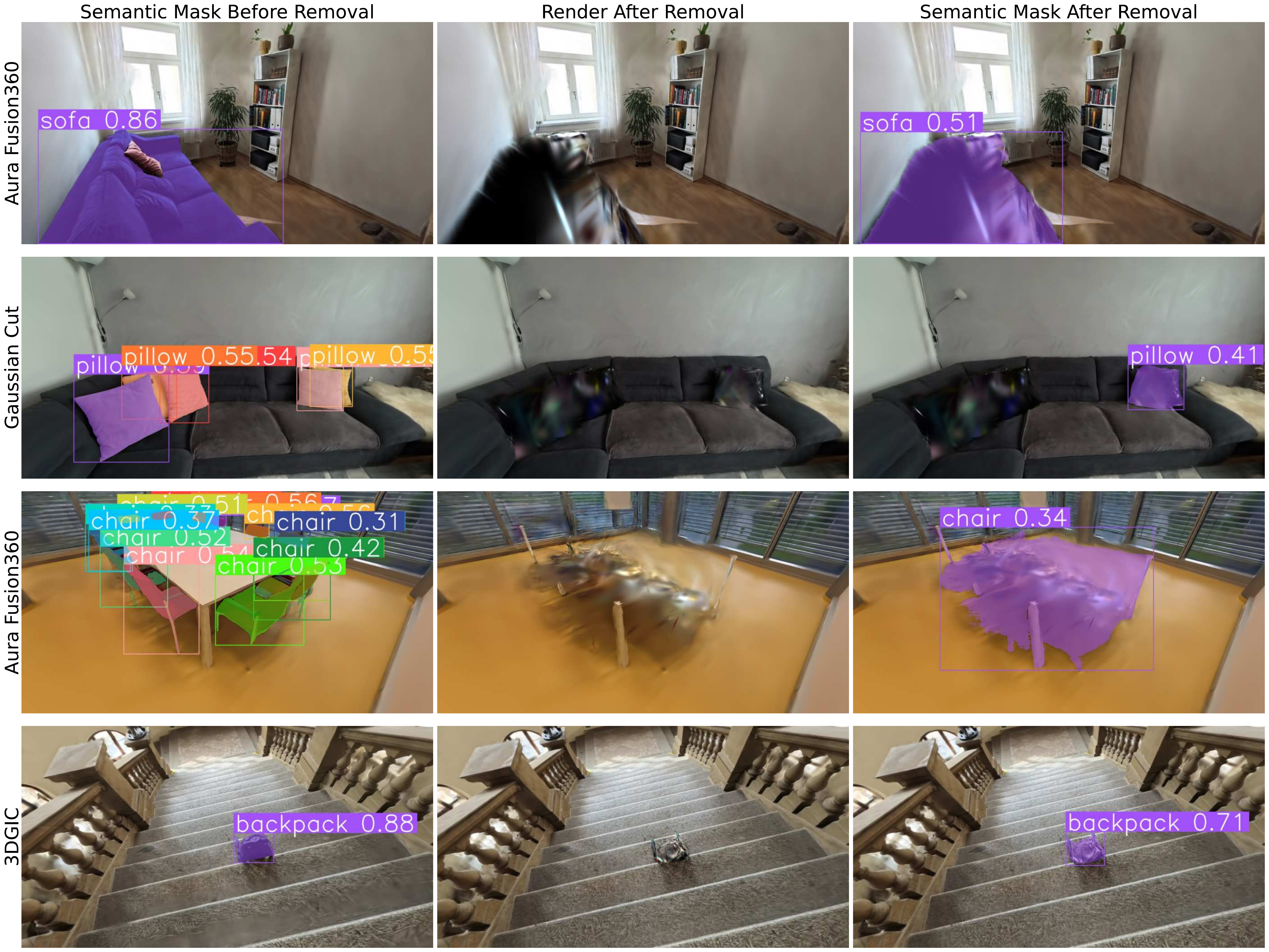}
\caption{\textbf{Category-level probe responses on Remove360.} GroundedSAM2~\cite{kirillov2023segany,liu2023grounding,ren2024grounded,ravi2024sam2} is applied to pre- and post-removal renderings using the target category prompt. The examples illustrate that a visually plausible post-removal rendering can still produce a target-associated response. The probe does not identify whether the response originates from geometry, boundaries, shadows, context, reconstruction error, or inpainting.}
\label{fig:supp_remove360_gsam}
\end{figure*}

\begin{figure*}[t]
\centering
\caption{\textbf{Remove360 qualitative comparison: Deckchair and White Chairs.} Rows compare the pose-matched post-removal reference rendering, Gaussian Cut~\cite{jain2024gaussiancut}, AuraFusion360~\cite{wu2025aurafusion}, and 3DGIC~\cite{huang20253d}. The grids jointly show RGB appearance, target-region depth change, and category-prompt-free Segment Anything structure~\cite{kirillov2023segany,ravi2024sam2}.}
\label{fig:supp_remove360_outdoor_1}
\begin{subfigure}[t]{0.99\linewidth}\centering
\includegraphics[width=\linewidth]{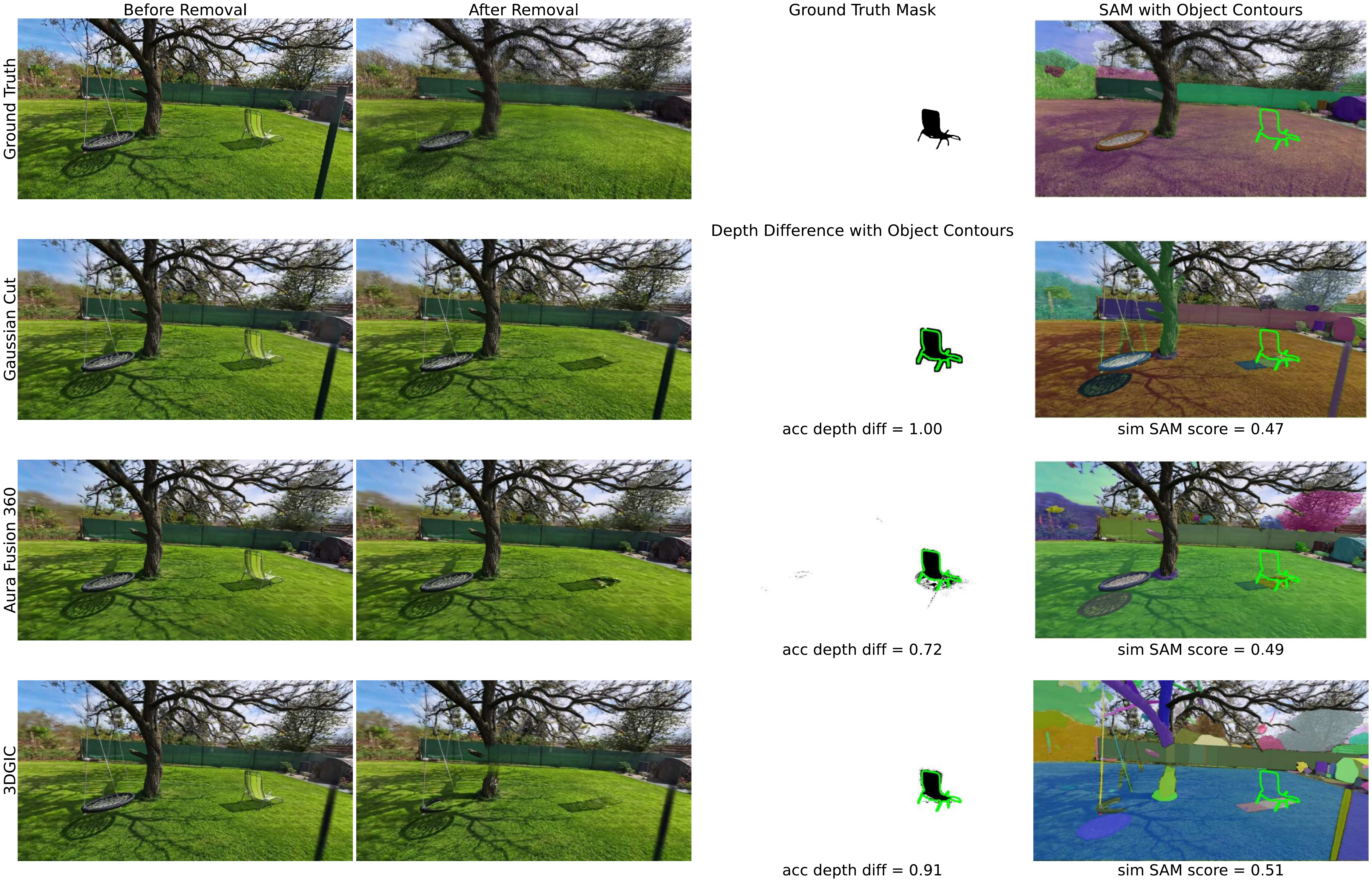}
\caption{Deckchair.}\end{subfigure}
\begin{subfigure}[t]{0.99\linewidth}\centering
\includegraphics[width=\linewidth]{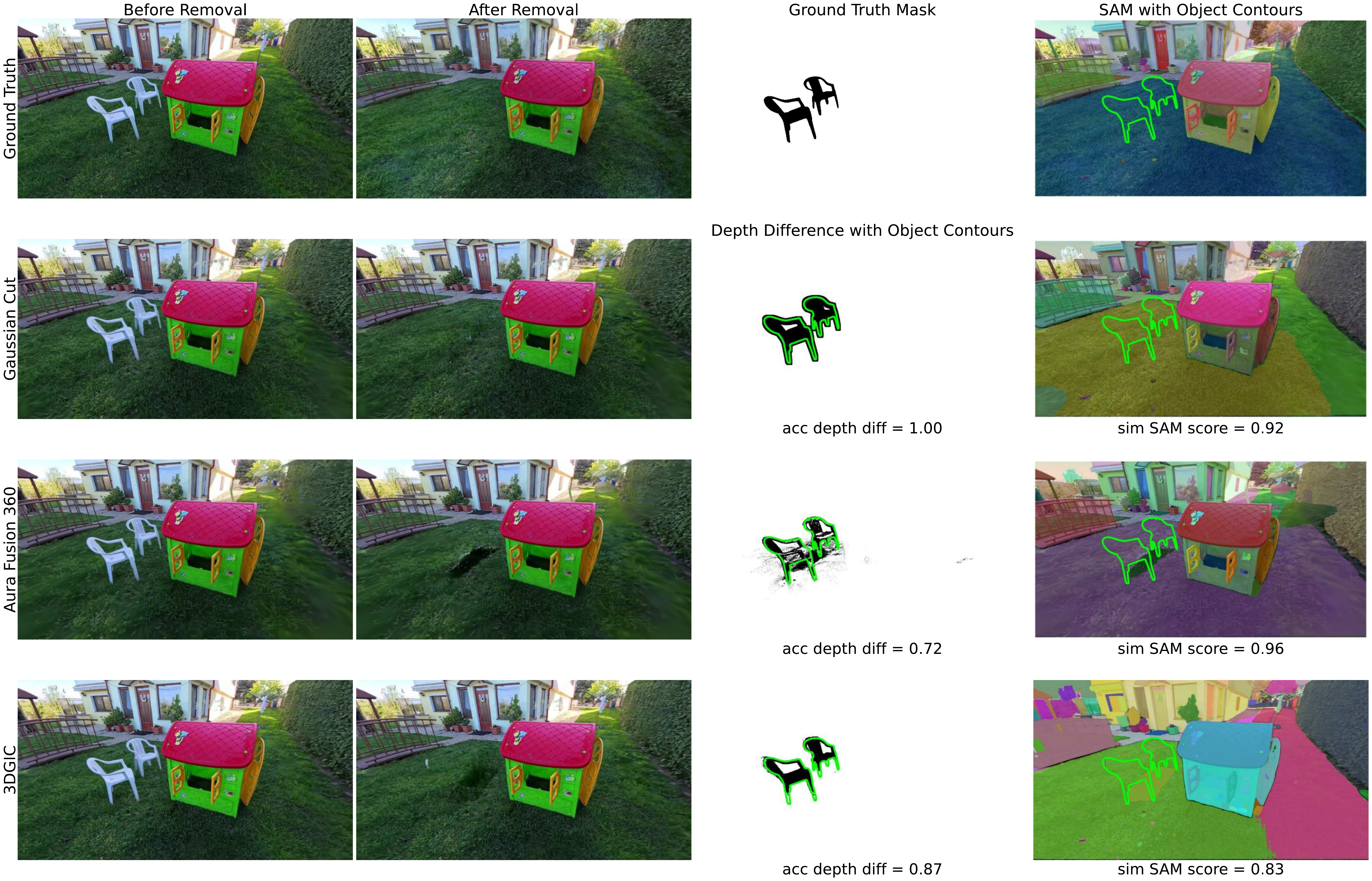}
\caption{White Chairs.}\end{subfigure}
\end{figure*}

\begin{figure*}[t]
\centering
\caption{\textbf{Remove360 qualitative comparison: Playhouse and Stroller.} The methods exhibit different trade-offs between visual completion, geometric modification, and structural agreement with the pose-matched post-removal reference rendering.}
\label{fig:supp_remove360_outdoor_2}
\begin{subfigure}[t]{0.99\linewidth}\centering
\includegraphics[width=\linewidth]{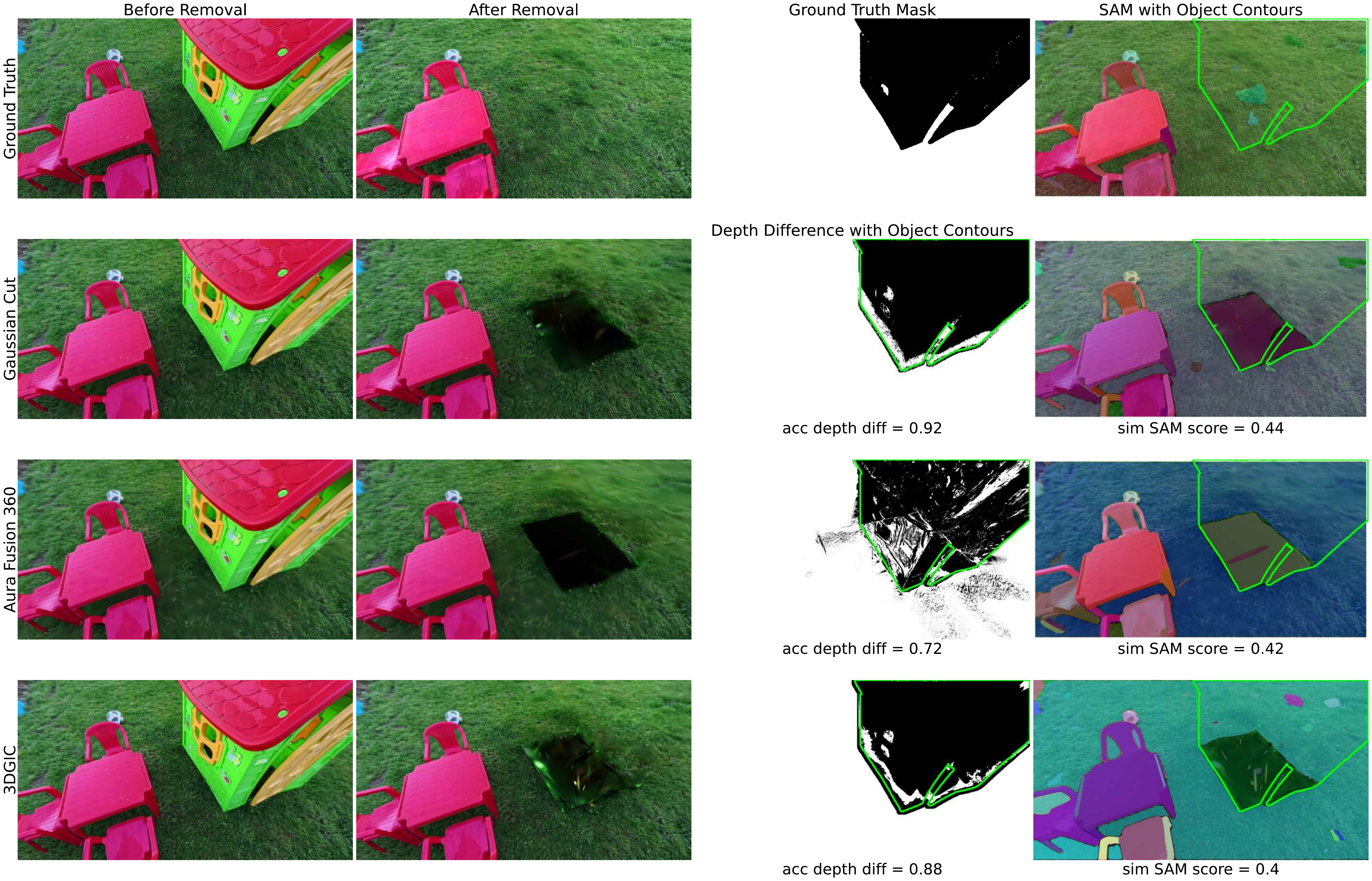}
\caption{Playhouse.}\end{subfigure}
\begin{subfigure}[t]{0.99\linewidth}\centering
\includegraphics[width=\linewidth]{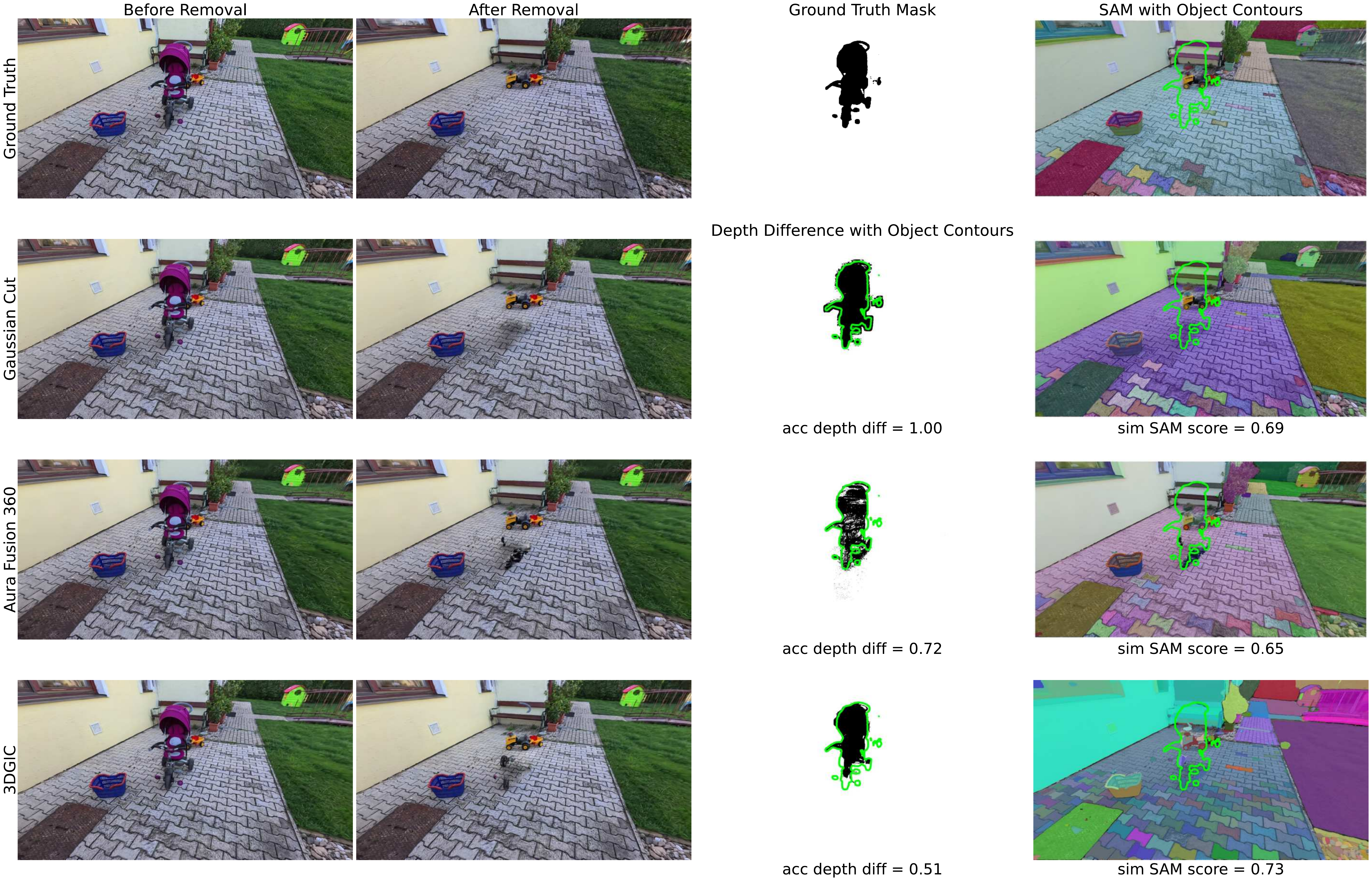}
\caption{Stroller.}\end{subfigure}
\end{figure*}

\begin{figure*}[t]
\centering
\caption{\textbf{Remove360 qualitative comparison: Toy Truck and Bicycle.} Residual structures and background-completion errors can remain visible in the semantic, structural, or depth probes even when the RGB result appears plausible.}
\label{fig:supp_remove360_outdoor_3}
\begin{subfigure}[t]{0.99\linewidth}\centering
\includegraphics[width=\linewidth]{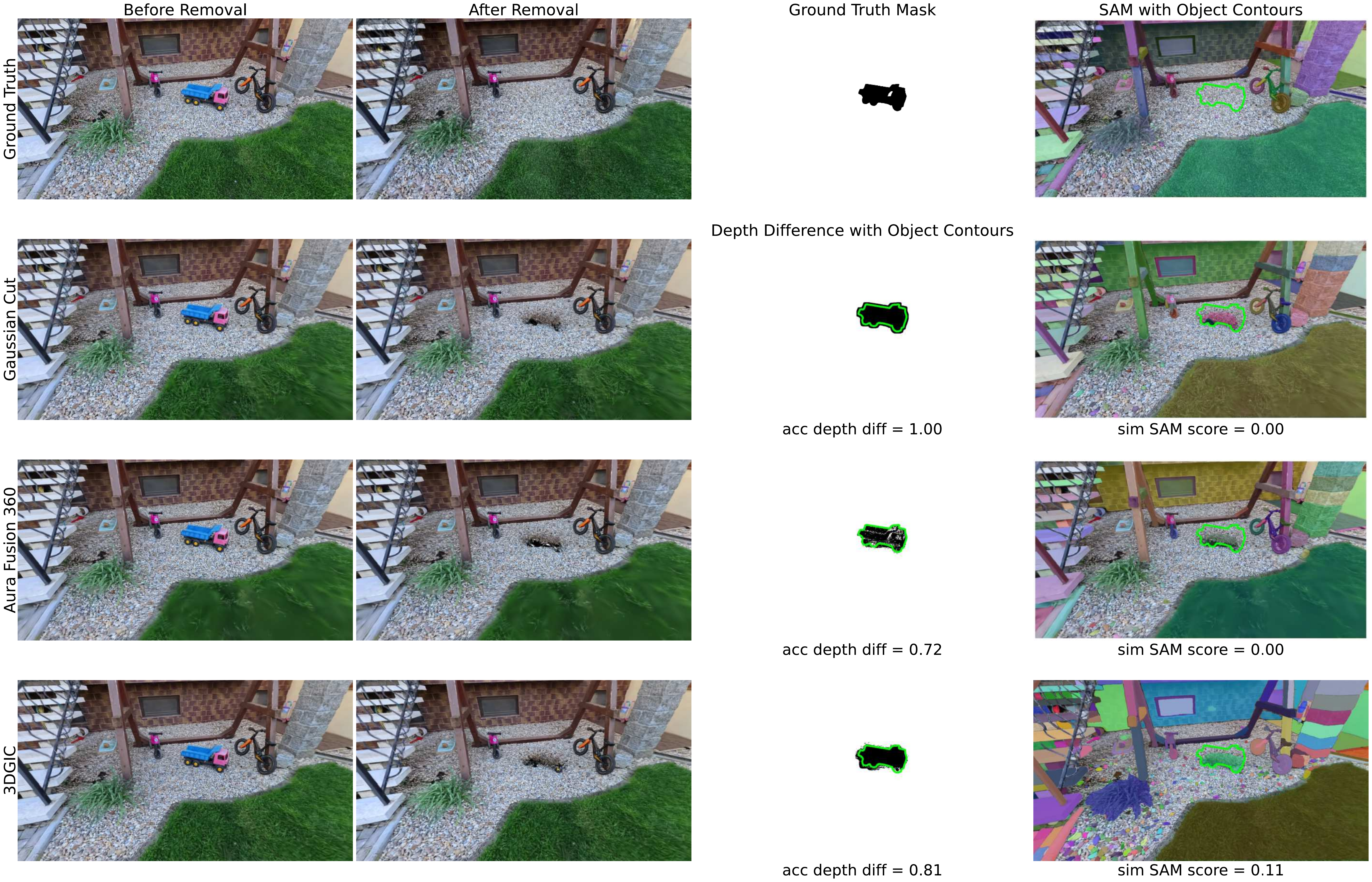}
\caption{Toy Truck.}\end{subfigure}
\begin{subfigure}[t]{0.99\linewidth}\centering
\includegraphics[width=\linewidth]{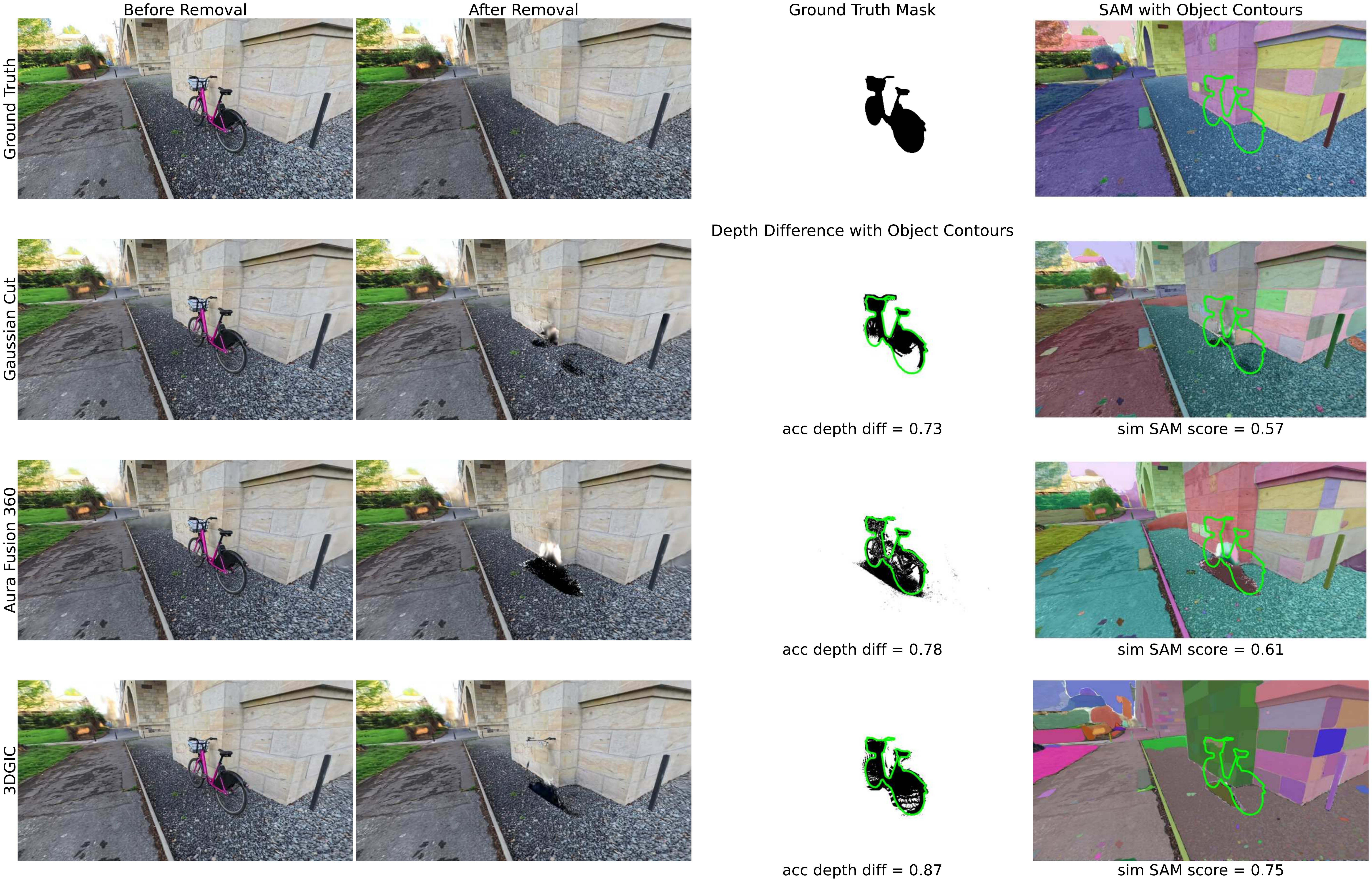}
\caption{Bicycle.}\end{subfigure}
\end{figure*}

\begin{figure*}[t]
\centering
\caption{\textbf{Remove360 qualitative comparison: Sofa and Backpack.} These scenes illustrate the difficulty of removing large or contextually embedded objects while recovering the newly exposed background visible in the pose-matched reference rendering.}
\label{fig:supp_remove360_indoor_1}
\begin{subfigure}[t]{0.99\linewidth}\centering
\includegraphics[width=\linewidth]{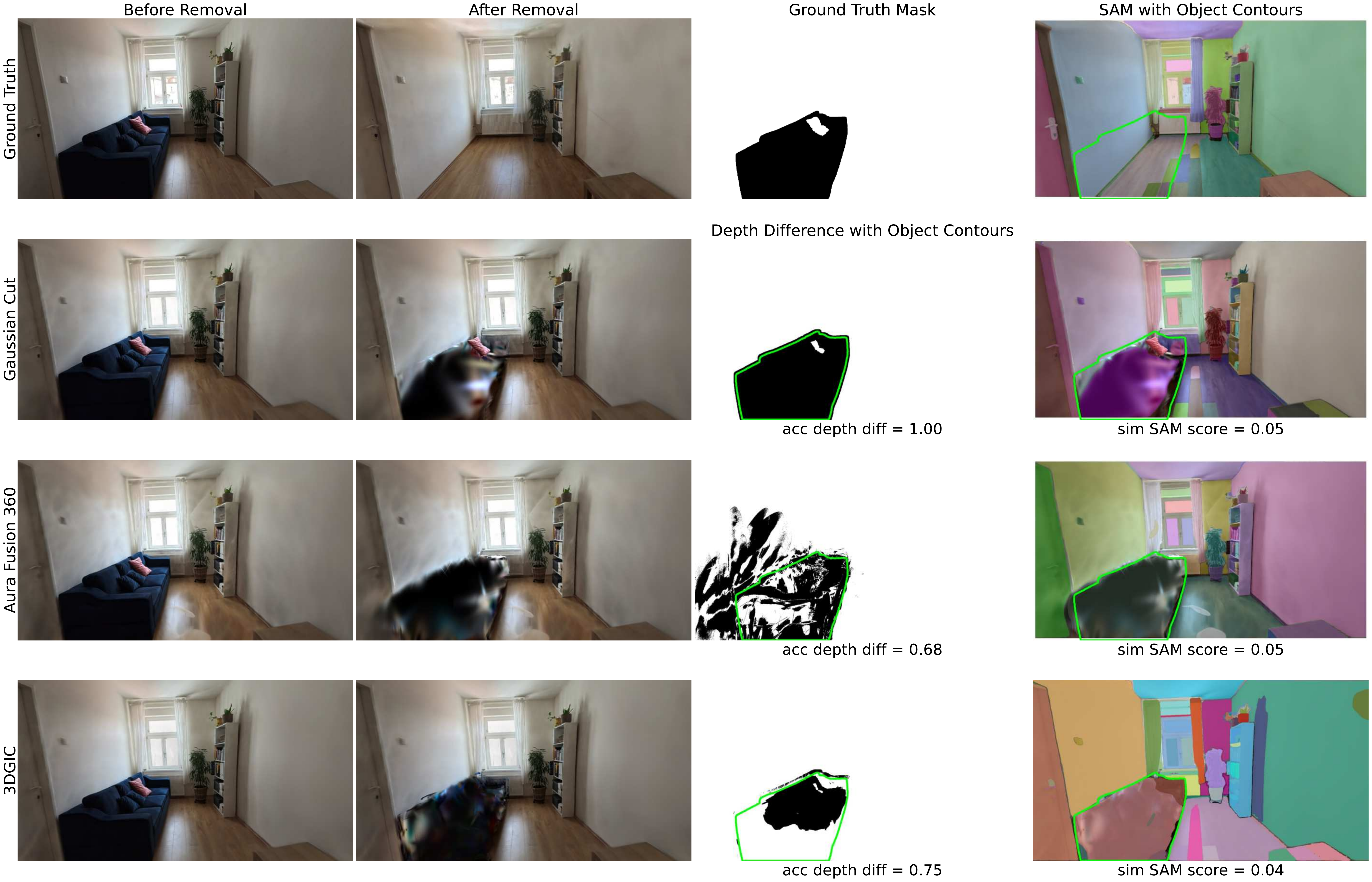}
\caption{Sofa.}\end{subfigure}
\begin{subfigure}[t]{0.99\linewidth}\centering
\includegraphics[width=\linewidth]{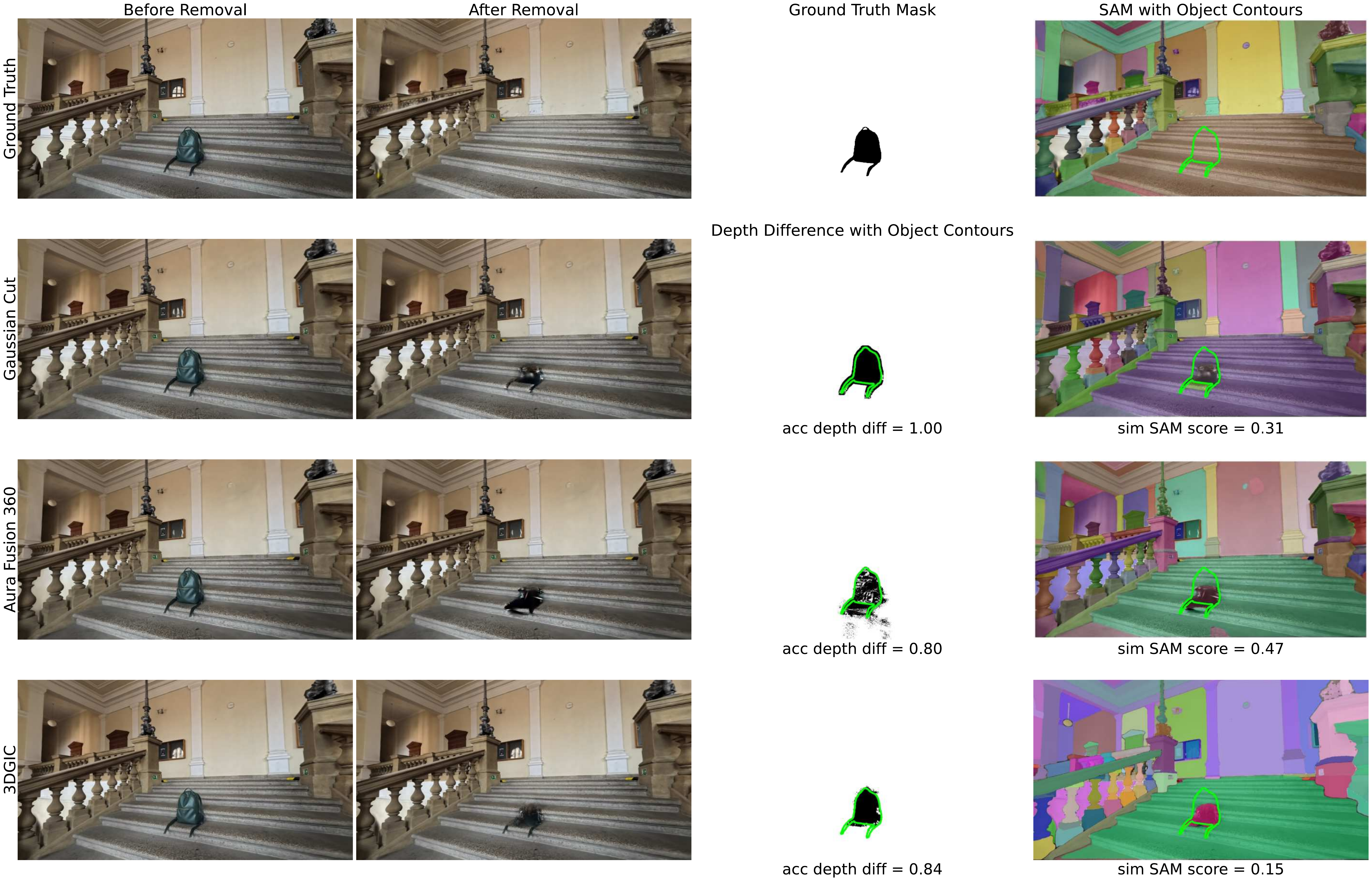}
\caption{Backpack.}\end{subfigure}
\end{figure*}

\begin{figure*}[t]
\centering
\caption{\textbf{Remove360 qualitative comparison: Office Chairs and Pillows.} Multiple-instance removal produces different failure modes across the category-level, structural, and geometric axes.}
\label{fig:supp_remove360_indoor_2}
\begin{subfigure}[t]{0.99\linewidth}\centering
\includegraphics[width=\linewidth]{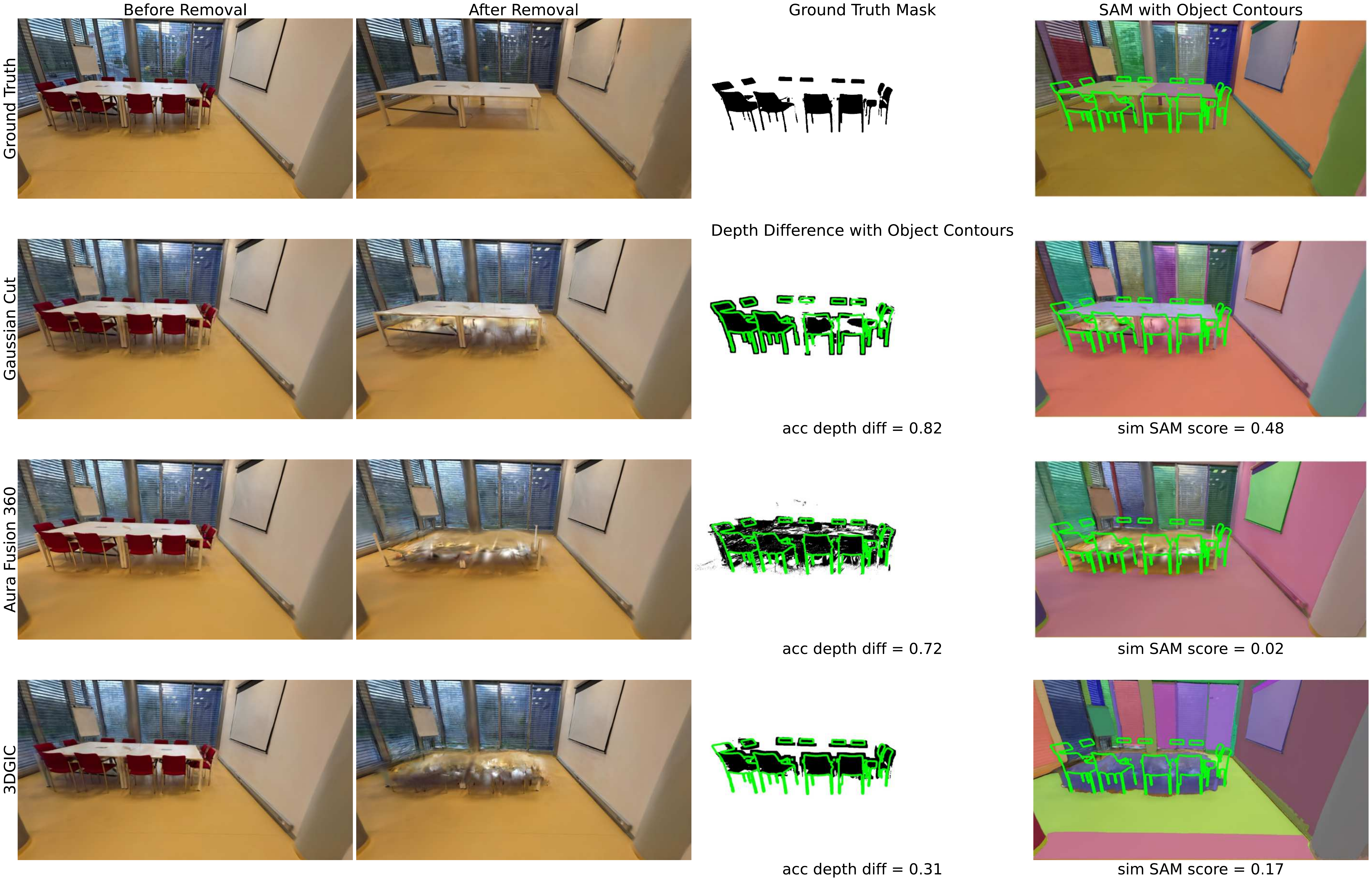}
\caption{Office Chairs.}\end{subfigure}
\begin{subfigure}[t]{0.99\linewidth}\centering
\includegraphics[width=\linewidth]{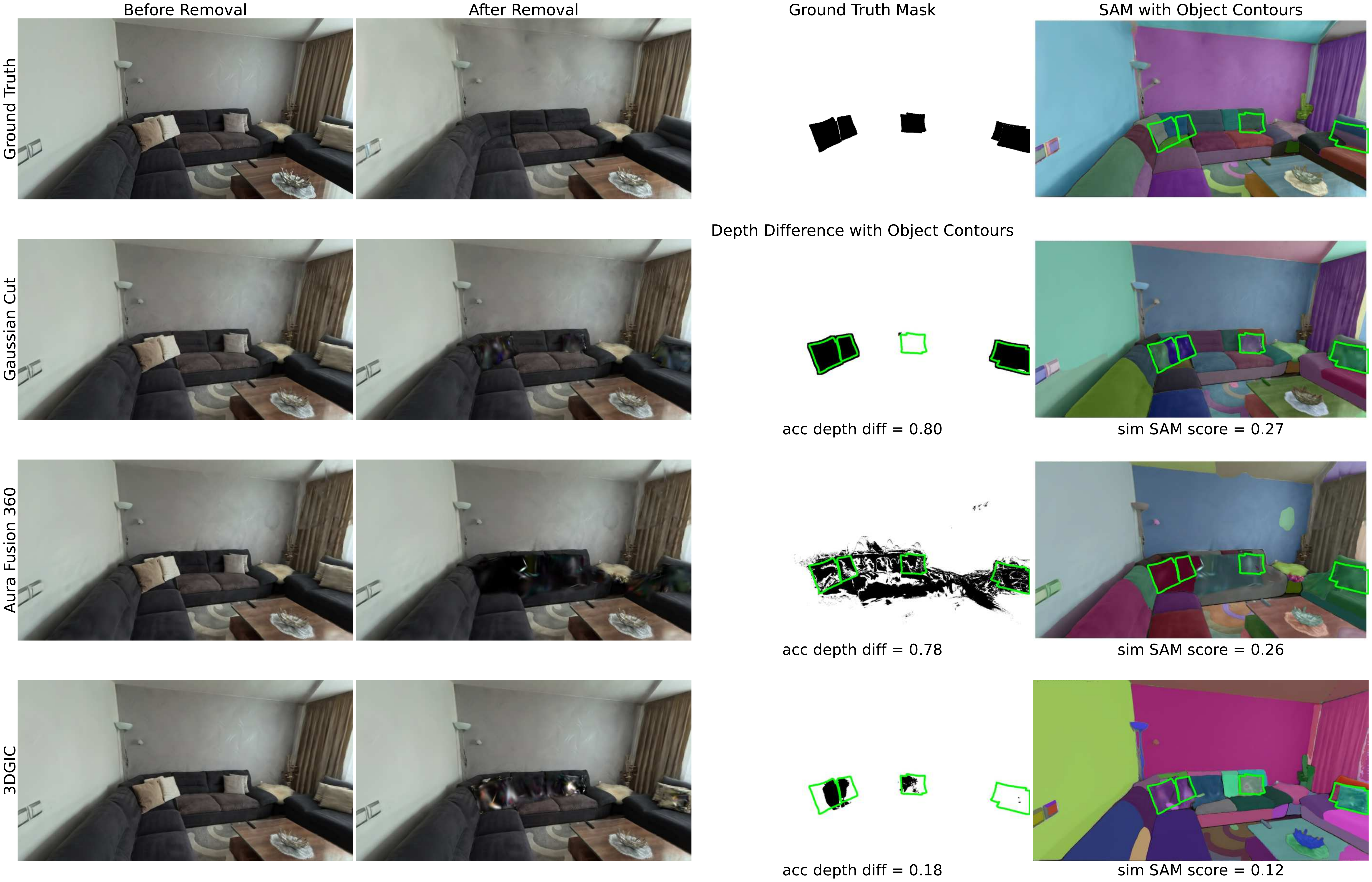}
\caption{Pillows.}\end{subfigure}
\end{figure*}

\begin{figure*}[t]
\centering
\caption{\textbf{Remove360 qualitative comparison: Table.} The table is strongly coupled to nearby walls and floor geometry, making the scene useful for inspecting whether geometric change agrees with the physical post-removal structure.}
\label{fig:supp_remove360_indoor_3}
\includegraphics[width=0.90\linewidth]{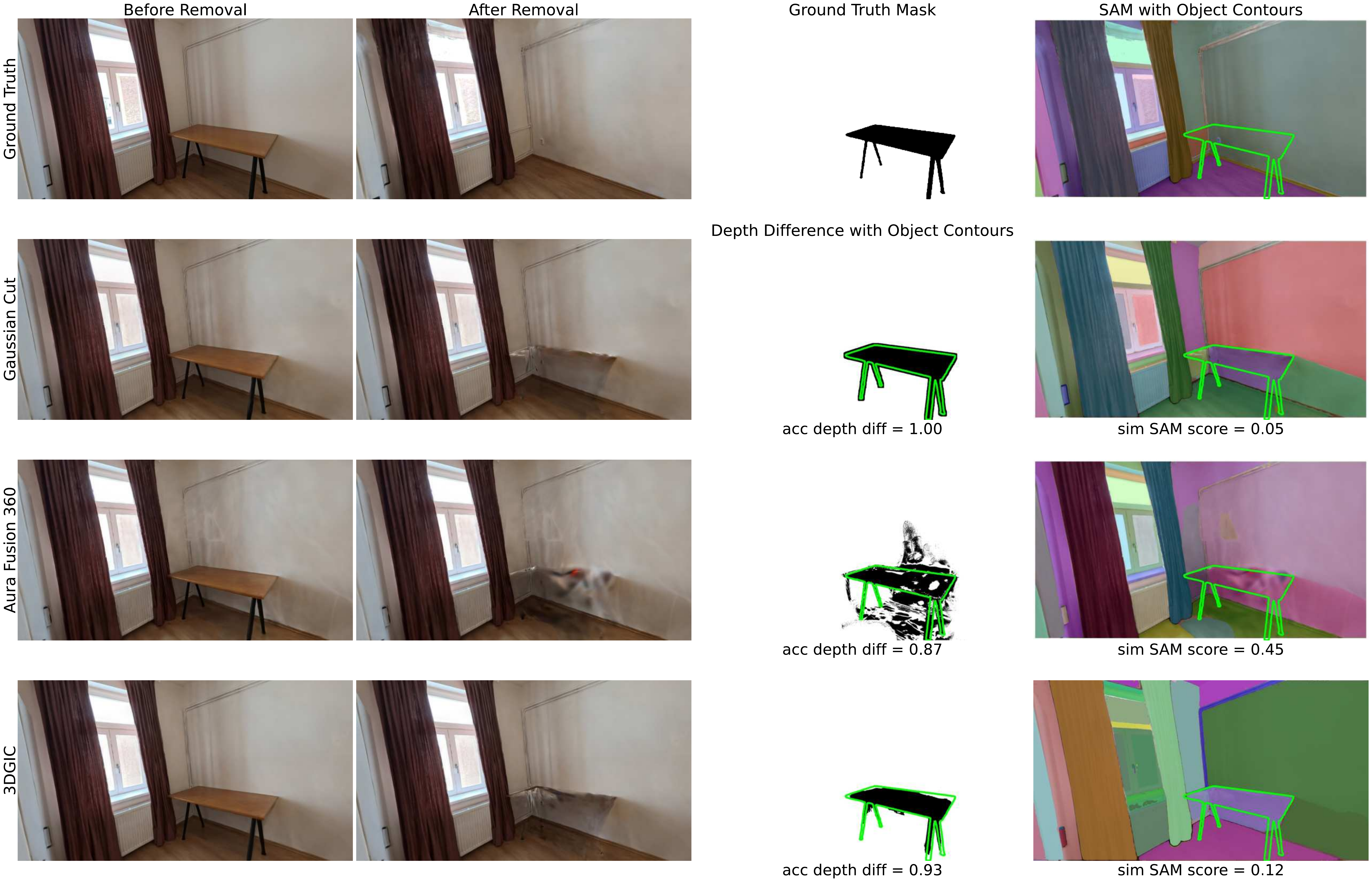}
\end{figure*}

\subsection{Effect of Inpainting}\label{supp:results_remove360_inpainting}
AuraFusion360 and 3DGIC provide optional inpainting stages. Inpainting increases aggregate RQS from 0.630 to 0.655 for AuraFusion360 and from 0.690 to 0.693 for 3DGIC, mainly through stronger geometric modification. Structural agreement decreases slightly, from 0.409 to 0.407 and from 0.464 to 0.448, respectively. The figures show why: inpainting can complete the background but can also introduce view-dependent or object-like content. The benchmark evaluates the final shared representation and does not attribute these responses to a single source.

\begin{figure*}[t]
\centering
\caption{\textbf{Removal-only versus inpainting-enabled outputs (1/3).} We compare AuraFusion360~\cite{wu2025aurafusion} and 3DGIC~\cite{huang20253d} before and after their optional inpainting stages. Inpainting can improve background completion, but can also introduce view-dependent or object-like content.}
\label{fig:supp_inpaint_1}
\begin{subfigure}[t]{0.99\linewidth}\centering\includegraphics[width=\linewidth]{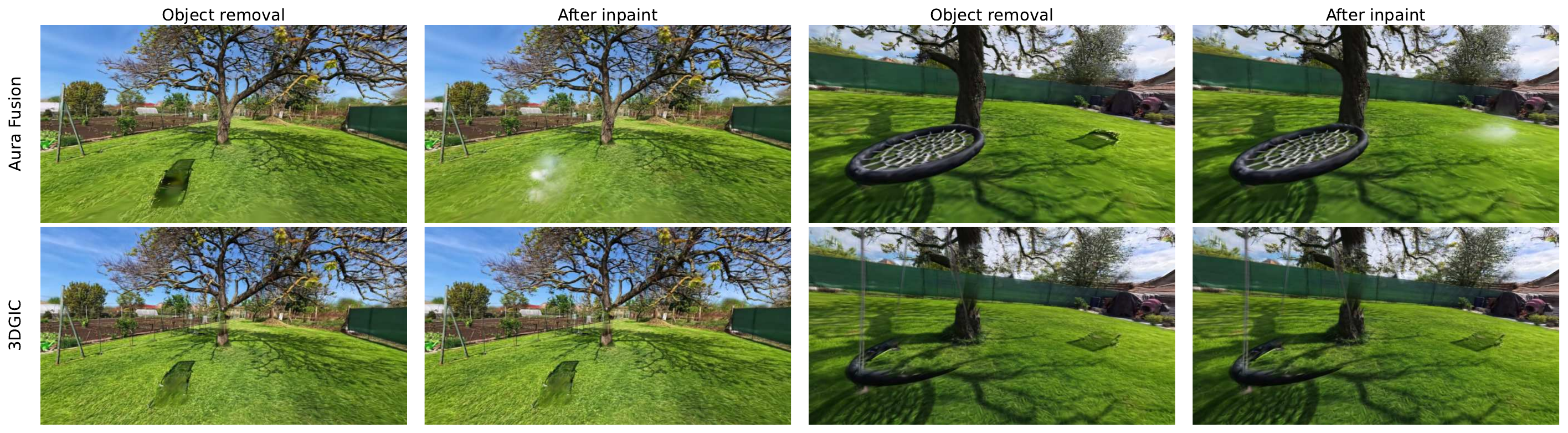}\caption{Deckchair.}\end{subfigure}
\begin{subfigure}[t]{0.99\linewidth}\centering\includegraphics[width=\linewidth]{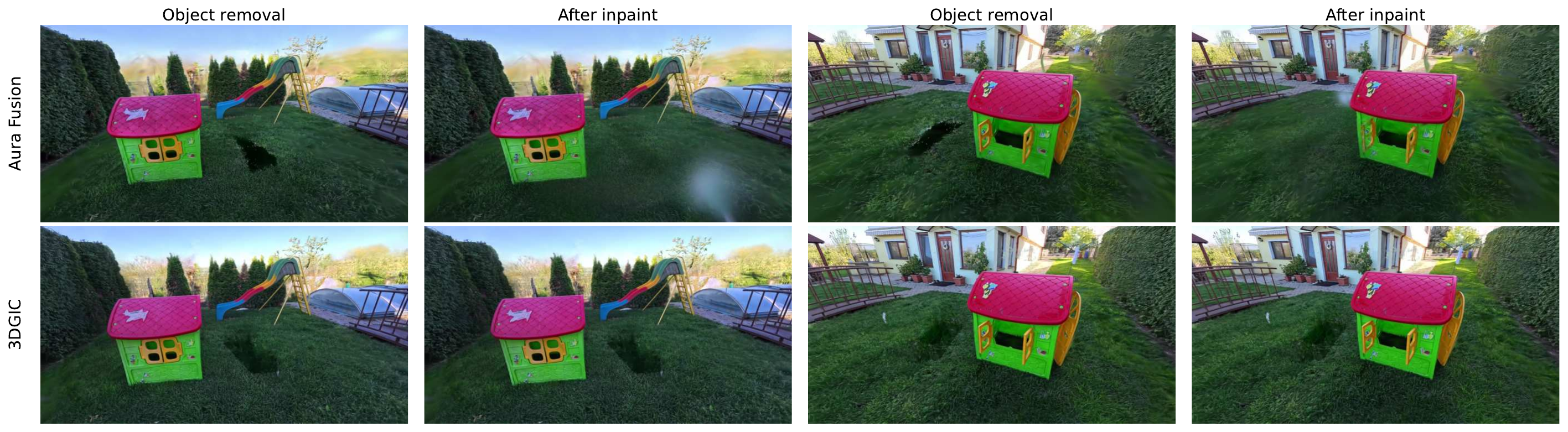}\caption{White Chairs.}\end{subfigure}
\begin{subfigure}[t]{0.99\linewidth}\centering\includegraphics[width=\linewidth]{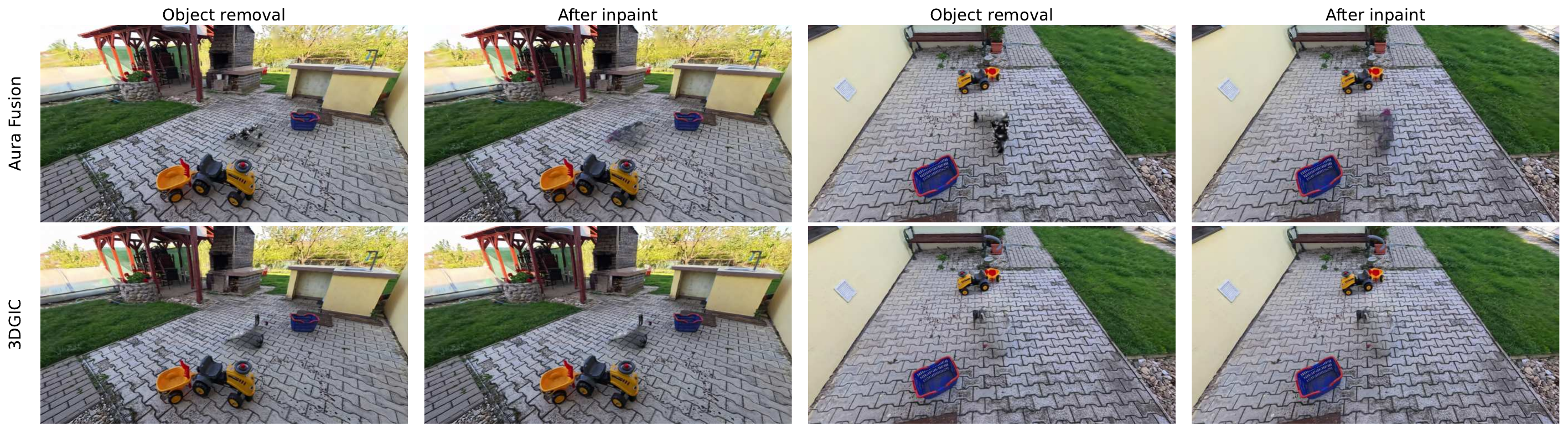}\caption{Stroller.}\end{subfigure}
\begin{subfigure}[t]{0.99\linewidth}\centering\includegraphics[width=\linewidth]{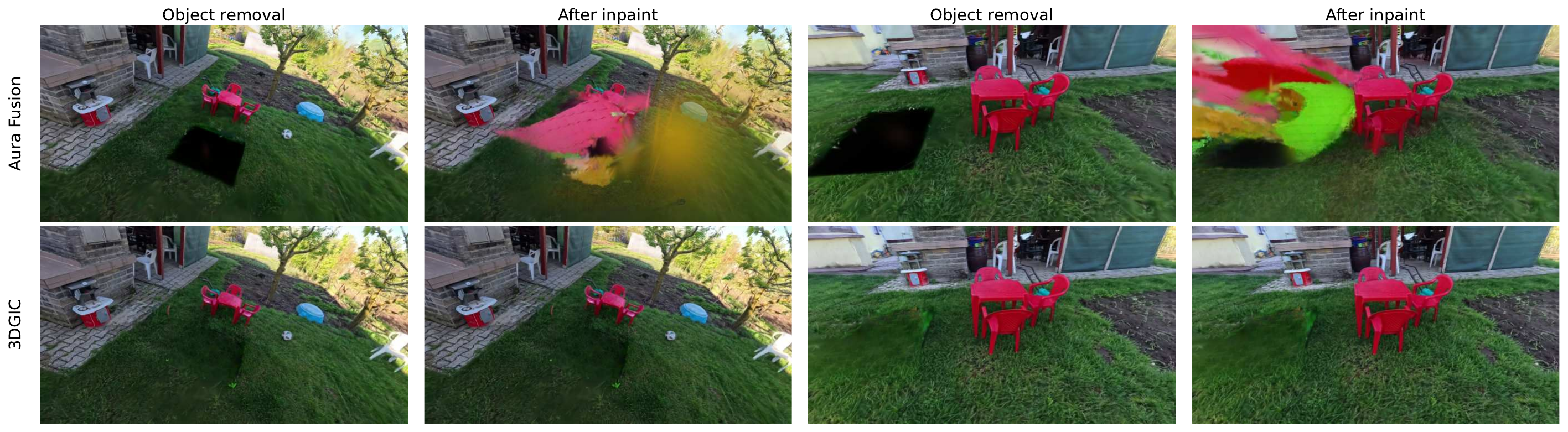}\caption{Playhouse.}\end{subfigure}
\end{figure*}

\begin{figure*}[t]
\centering
\caption{\textbf{Removal-only versus inpainting-enabled outputs (2/3).} The examples show that stronger geometric completion does not necessarily improve semantic suppression or prompt-free structural agreement.}
\label{fig:supp_inpaint_2}
\begin{subfigure}[t]{0.99\linewidth}\centering\includegraphics[width=\linewidth]{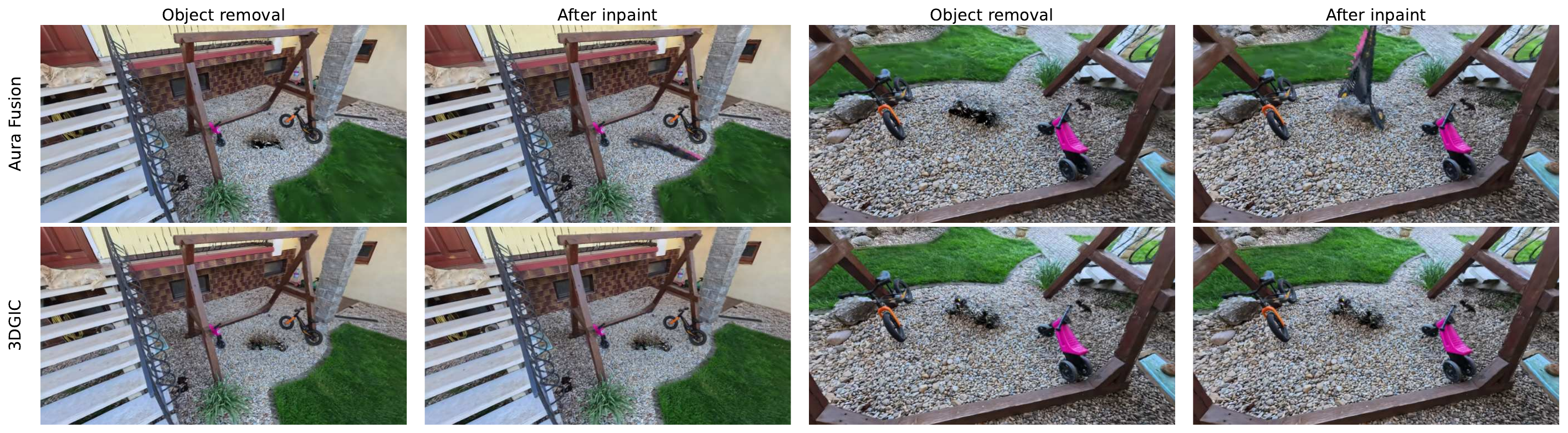}\caption{Toy Truck.}\end{subfigure}
\begin{subfigure}[t]{0.99\linewidth}\centering\includegraphics[width=\linewidth]{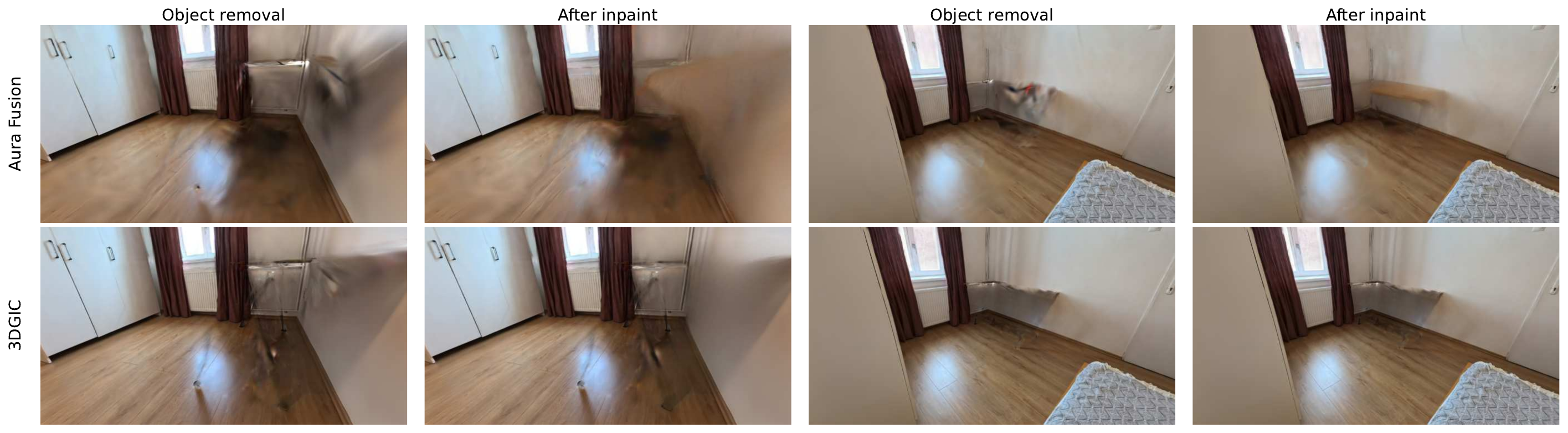}\caption{Table.}\end{subfigure}
\begin{subfigure}[t]{0.99\linewidth}\centering\includegraphics[width=\linewidth]{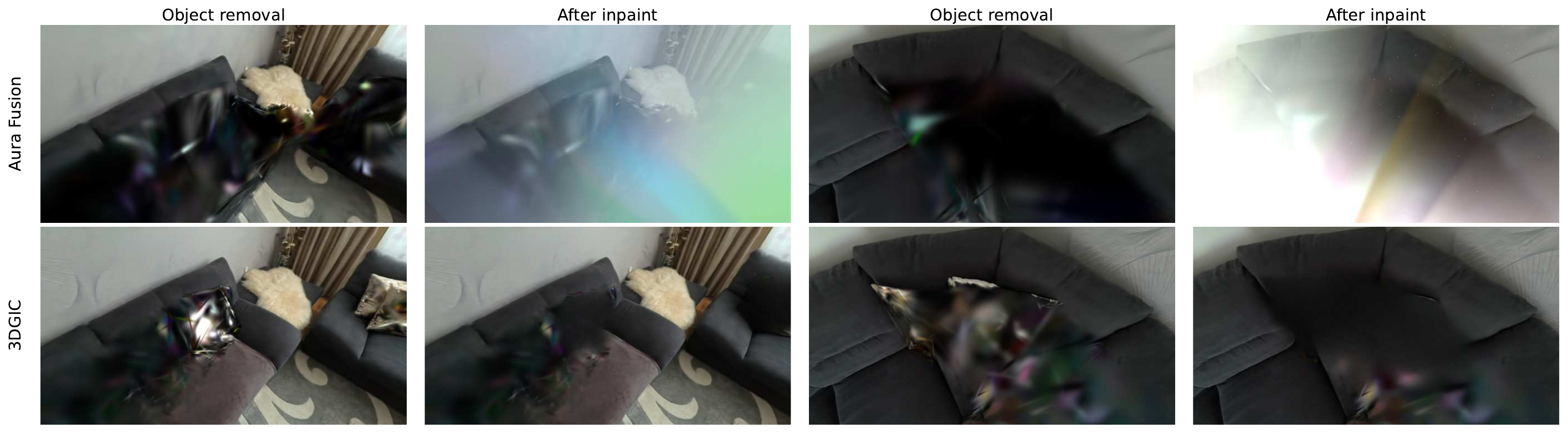}\caption{Pillows.}\end{subfigure}
\begin{subfigure}[t]{0.99\linewidth}\centering\includegraphics[width=\linewidth]{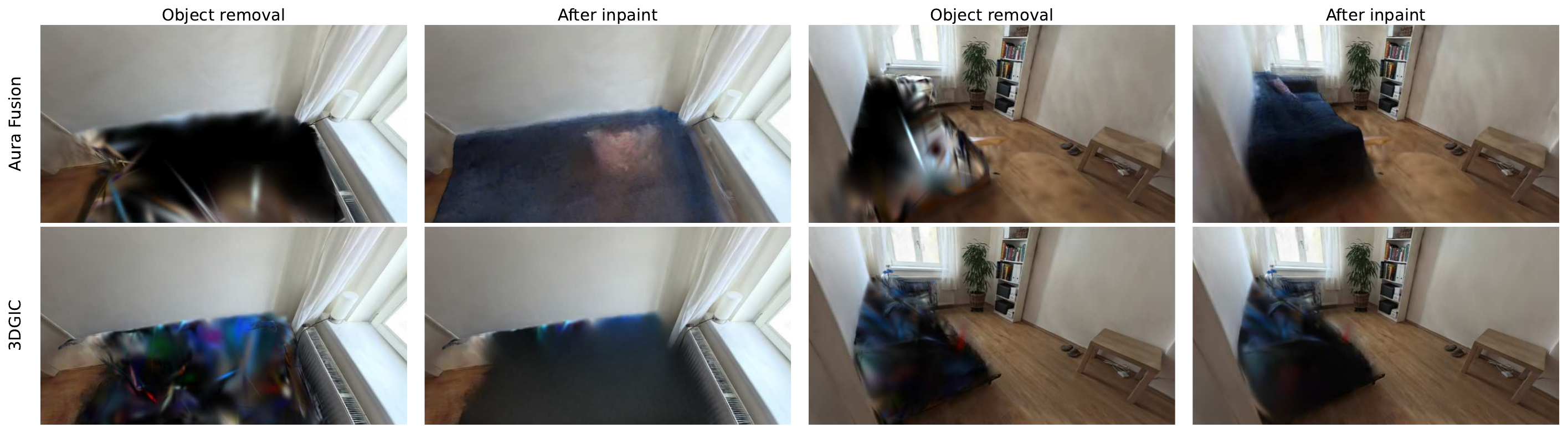}\caption{Sofa.}\end{subfigure}
\end{figure*}

\begin{figure*}[t]
\centering
\caption{\textbf{Removal-only versus inpainting-enabled outputs (3/3).} The final shared representation is evaluated after inpainting; the benchmark therefore captures both useful completion and artifacts introduced by the inpainting prior.}
\label{fig:supp_inpaint_3}
\begin{subfigure}[t]{0.99\linewidth}\centering\includegraphics[width=\linewidth]{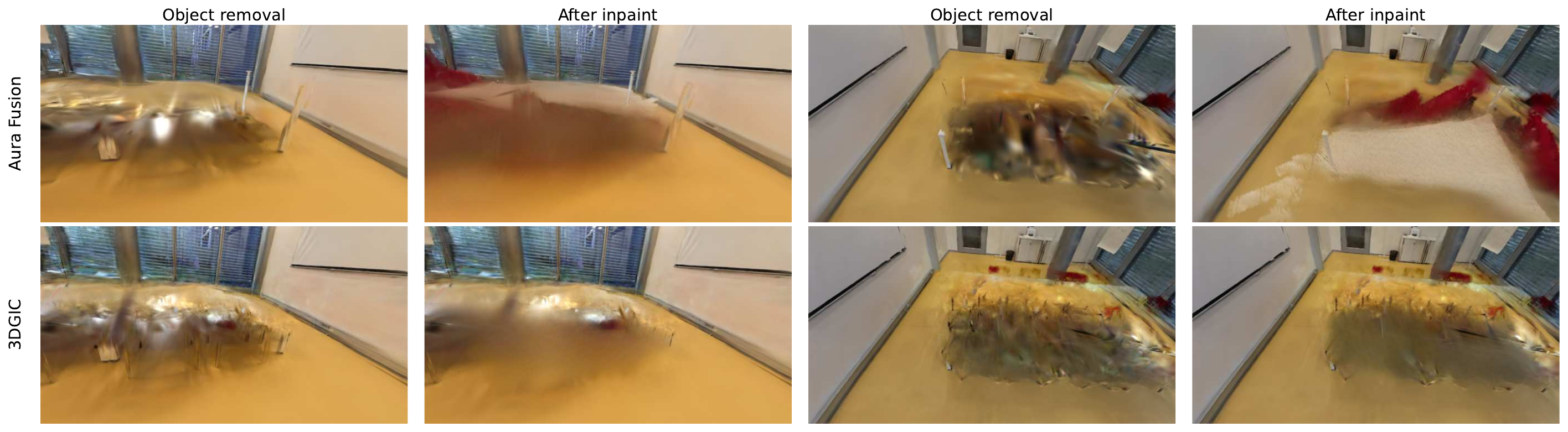}\caption{Office Chairs.}\end{subfigure}
\begin{subfigure}[t]{0.99\linewidth}\centering\includegraphics[width=\linewidth]{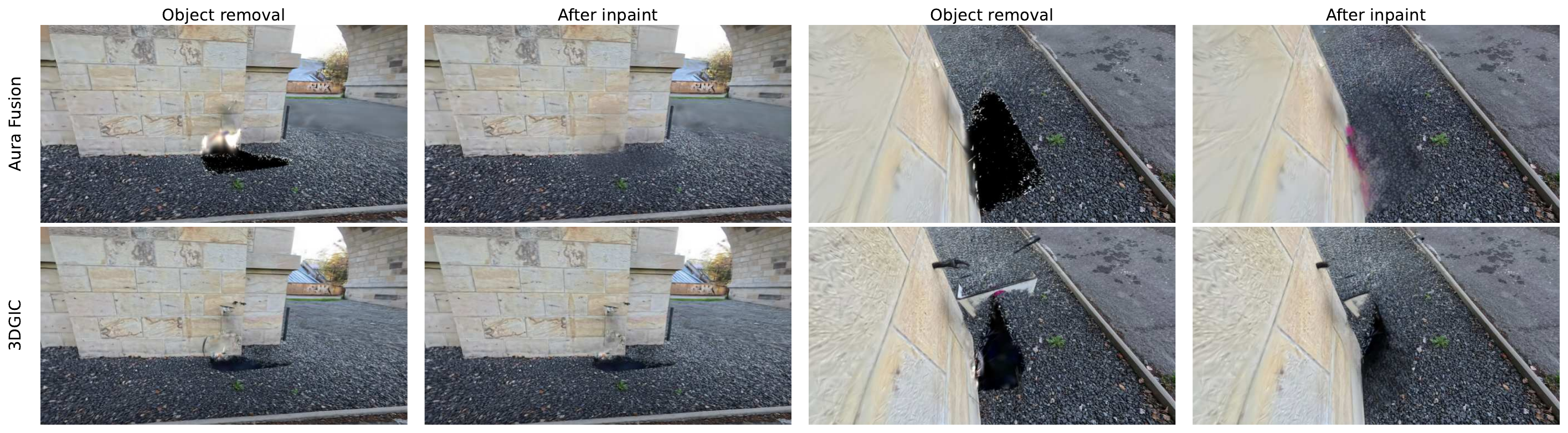}\caption{Bicycle.}\end{subfigure}
\begin{subfigure}[t]{0.99\linewidth}\centering\includegraphics[width=\linewidth]{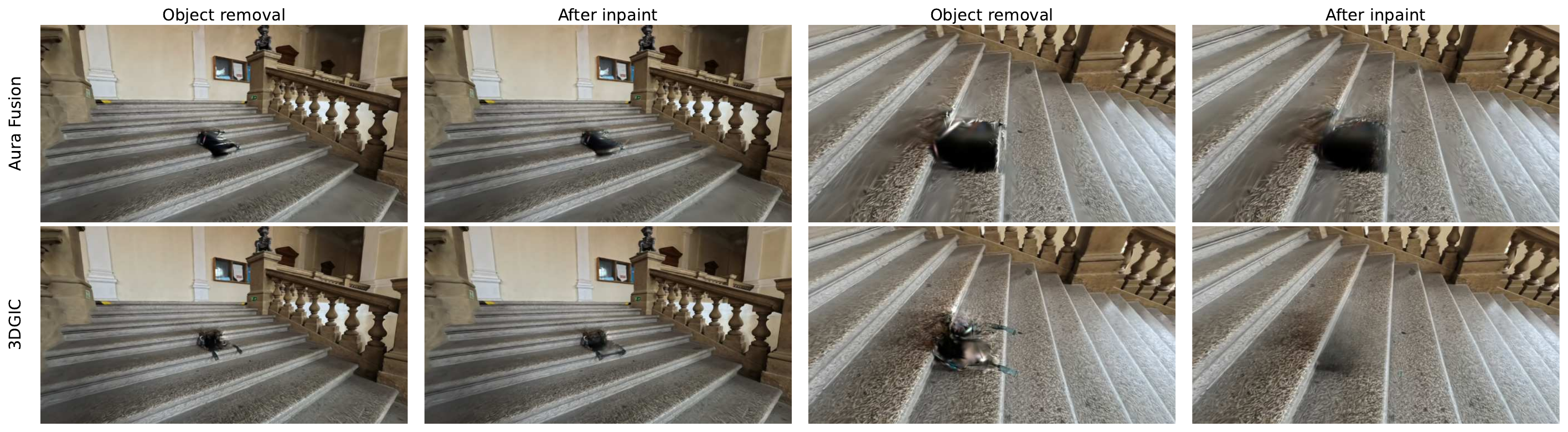}\caption{Backpack.}\end{subfigure}
\end{figure*}

\section{Mip-NeRF360 Results}\label{supp:results_mipnerf360}
\subsection{Quantitative Results}\label{supp:results_mipnerf360_quantitative}
Mip-NeRF360~\cite{barron2022mip} has no physical post-removal references. We therefore report category suppression, depth modification relative to the original reconstruction, and structural change $1-\mathrm{sim}_{\mathrm{SAM}}(I_{\mathrm{pre}},I_{\mathrm{post}})$. These scores quantify change from the original scene and should not be interpreted as agreement with a true physical post-removal state.

\begin{table}[!htbp]
\centering
\caption{\textbf{GroundedSAM2~\cite{kirillov2023segany,liu2023grounding,ren2024grounded,ravi2024sam2} target IoU on Mip-NeRF360~\cite{barron2022mip}.} Each entry reports pre-removal IoU/post-removal IoU/$\mathrm{IoU}_{\mathrm{drop}}$. Bold and underlined components mark the largest and second-largest drop for each target. GG, FGS, SAGS, GC, and AF denote Gaussian Grouping~\cite{gaussian_grouping}, Feature 3DGS~\cite{zhou2024feature}, SAGS~\cite{hu2024semantic}, Gaussian Cut~\cite{jain2024gaussiancut}, and AuraFusion360~\cite{wu2025aurafusion}.}
\label{tab:supp_mip_iou}
\suppTableStyle
\setlength{\tabcolsep}{2.1pt}
\resizebox{\linewidth}{!}{%
\begin{tabular}{llccccc}
\toprule
Scene & Target & GG & FGS & SAGS & GC & AF \\
\midrule
Counter & Baking Tray & .61/.08/.53 & .54/.21/.34 & .62/.52/.10 & .63/.01/\B{.62} & .64/.04/\U{.60} \\
 & Plant & .84/.00/.84 & .75/.00/.75 & .85/.82/.03 & .86/.00/\U{.86} & .87/.00/\B{.87} \\
 & Blue Gloves & .75/.15/\U{.60} & .67/.66/.01 & .74/.64/.10 & .74/.15/\U{.60} & .76/.11/\B{.65} \\
 & Egg Box & .63/.00/\B{.63} & .78/.70/.08 & .60/.04/.56 & .63/.01/\U{.62} & .64/.01/\B{.63} \\
Room & Plant & .50/.23/\U{.26} & .53/.00/\B{.53} & .52/.35/.17 & .53/.00/\B{.53} & .49/.26/.23 \\
 & Slippers & .96/.14/\B{.82} & .96/.96/.00 & .96/.71/.25 & .97/.48/\U{.48} & .43/.37/.06 \\
 & Coffee Table & .88/.02/\B{.86} & .86/.29/\U{.57} & .89/.89/.00 & .89/.03/\B{.86} & .58/.03/.55 \\
Kitchen & Truck & .67/.06/.61 & .67/.05/.62 & .67/.00/\U{.67} & .67/.01/.66 & .96/.01/\B{.95} \\
Garden & Table & .89/.41/.48 & .90/.24/.67 & .90/.09/.81 & .90/.04/\U{.86} & .91/.01/\B{.90} \\
 & Ball & .16/.00/.16 & .06/.06/.00 & .41/.00/\U{.41} & .42/.00/\B{.42} & .42/.00/\B{.42} \\
 & Vase & .85/.22/.64 & .90/.11/.79 & .97/.01/\U{.96} & .97/.01/\B{.97} & .98/.01/\B{.97} \\
\bottomrule
\end{tabular}}
\end{table}

\begin{table}[!htbp]
\centering
\caption{\textbf{Sensitivity of category-level suppression to the IoU threshold on Mip-NeRF360~\cite{barron2022mip}.} Values are target-macro averages of $\mathrm{acc}_{\mathrm{seg},\xi}$ over the 11 evaluated targets. The main paper uses $\xi_{\mathrm{IoU}}=0.3$. Higher thresholds are less discriminative because the score approaches saturation.}
\label{tab:supp_mip_thresholds}
\suppTableStyle
\setlength{\tabcolsep}{5pt}
\begin{tabular}{lcccc}
\toprule
Method & $\xi_{\mathrm{IoU}}=0.3$ & $0.5$ & $0.7$ & $0.9$ \\
\midrule
GG~\cite{gaussian_grouping} & 0.849 & 0.877 & 0.906 & 0.939 \\
FGS~\cite{zhou2024feature} & 0.660 & 0.670 & 0.706 & 0.761 \\
SAGS~\cite{hu2024semantic} & 0.557 & 0.628 & 0.667 & 0.744 \\
GC~\cite{jain2024gaussiancut} & 0.914 & 0.926 & 0.942 & 0.980 \\
AF~\cite{wu2025aurafusion} & 0.938 & 0.938 & 0.942 & 0.976 \\
\bottomrule
\end{tabular}
\end{table}

\subsection{Why the Two Datasets Rank Methods Differently}\label{supp:cross_dataset}
Gaussian Cut~\cite{jain2024gaussiancut} ranks first on Mip-NeRF360 but mid-field on
Remove360. Two properties of the setup account for this, and neither implies a
contradiction.

First, the method sets differ (Sec.~\ref{supp:coverage}), so the two rankings are
taken over different competitors.

Second, and more fundamentally, the structural axis is not the same quantity in
the two settings. Without a post-removal reference,
$s_{\mathrm{str}} = 1-\mathrm{sim}_{\mathrm{SAM}}(I_{\mathrm{pre}},I_{\mathrm{post}})$
credits \emph{any} structural change in the edited region. With the physical
reference available on Remove360, the same axis instead requires \emph{agreement}
with the scene that actually exists after removal. A method can therefore rank
highly on the reference-free form by changing the region substantially, and less
highly on the reference-based form if the resulting structure does not match the
real post-removal scene. This is the benchmark's argument in miniature, and it is
why the two tables should not be read as a cross-dataset comparison of the same
quantity.

\subsection{Qualitative Results}\label{supp:results_mipnerf360_qualitative}
The Mip-NeRF360 grids visualize category-level and category-prompt-free structural changes. No physical post-removal reference is available. Lower pre/post SAM similarity therefore indicates stronger structural change; equivalently, the reported $1-\mathrm{sim}_{\mathrm{SAM}}$ increases. This does not imply that the final structure is correct.

\begin{figure*}[t]
\centering
\caption{\textbf{Category-level probe responses on Mip-NeRF360~\cite{barron2022mip} (1/3).} The examples visualize GroundedSAM2~\cite{kirillov2023segany,liu2023grounding,ren2024grounded,ravi2024sam2} responses before and after removal.}
\label{fig:supp_mip_gsam_1}
\includegraphics[width=0.7\linewidth]{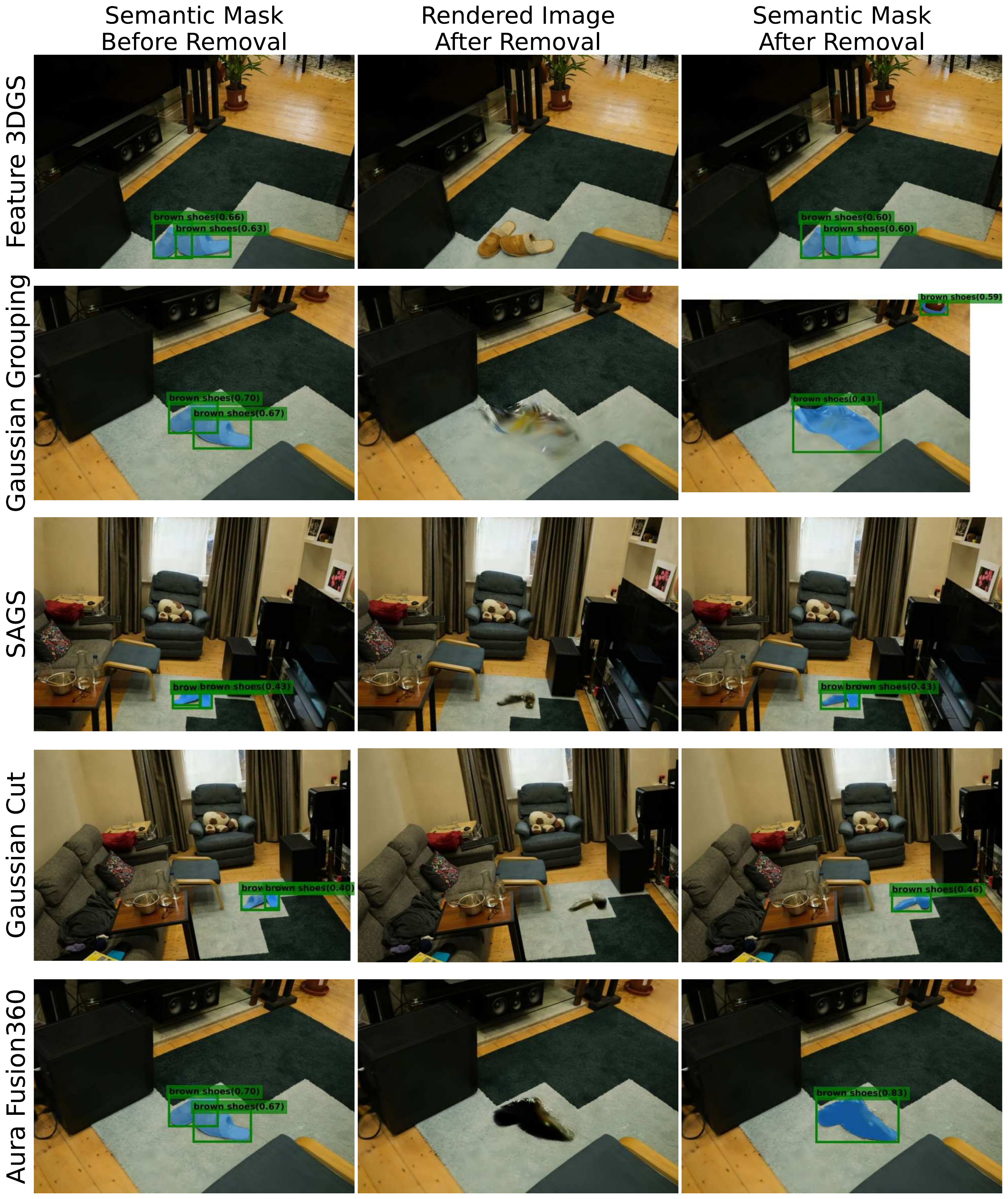}
\end{figure*}

\begin{figure*}[t]
\centering
\caption{\textbf{Category-level probe responses on Mip-NeRF360~\cite{barron2022mip} (2/3).} The examples visualize GroundedSAM2~\cite{kirillov2023segany,liu2023grounding,ren2024grounded,ravi2024sam2} responses before and after removal.}
\label{fig:supp_mip_gsam_2}
\includegraphics[width=0.7\linewidth]{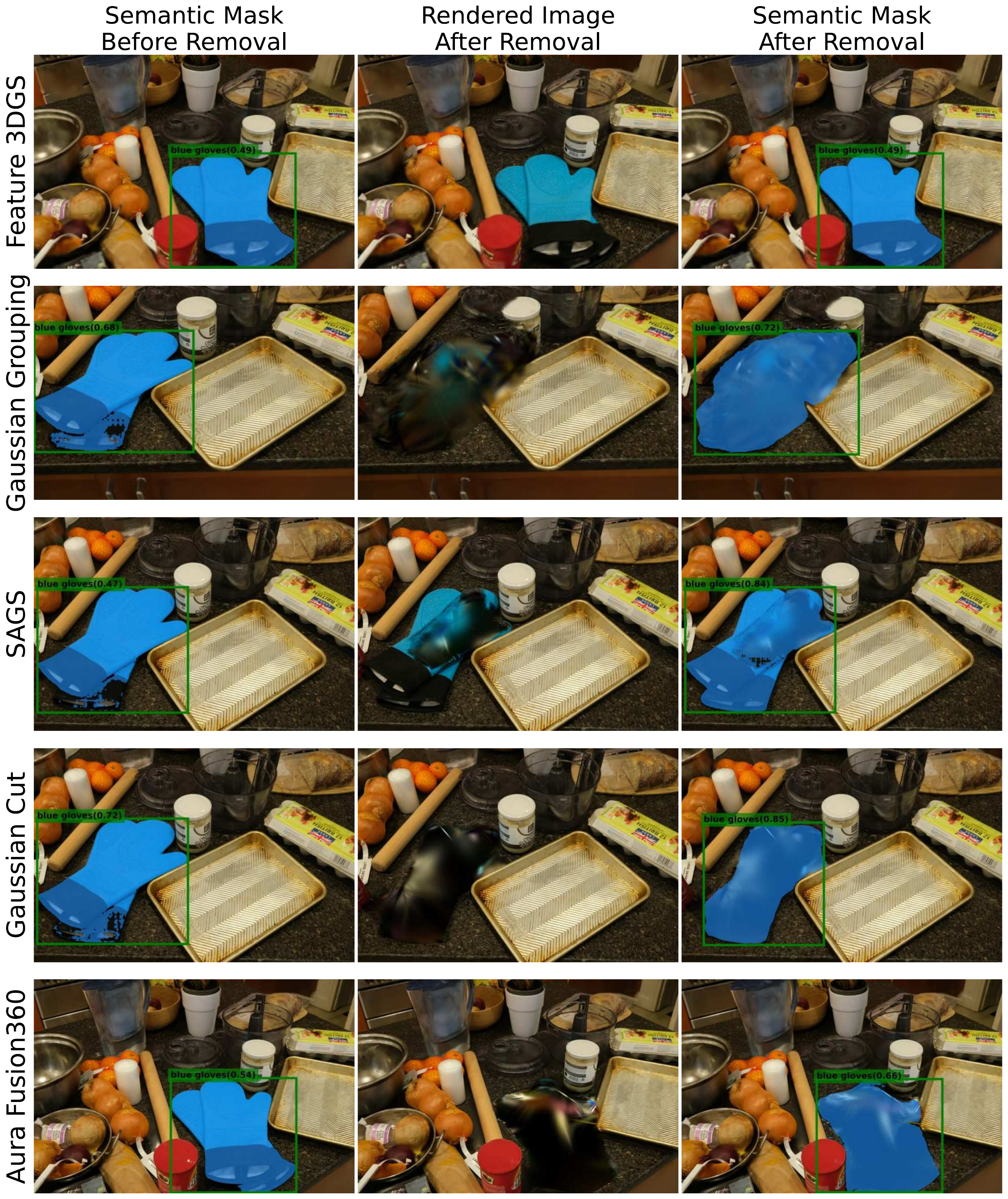}
\end{figure*}

\begin{figure*}[t]
\centering
\caption{\textbf{Category-level probe responses on Mip-NeRF360~\cite{barron2022mip} (3/3).} The examples visualize GroundedSAM2~\cite{kirillov2023segany,liu2023grounding,ren2024grounded,ravi2024sam2} responses before and after removal.}
\label{fig:supp_mip_gsam_gc}
\includegraphics[width=0.7\linewidth]{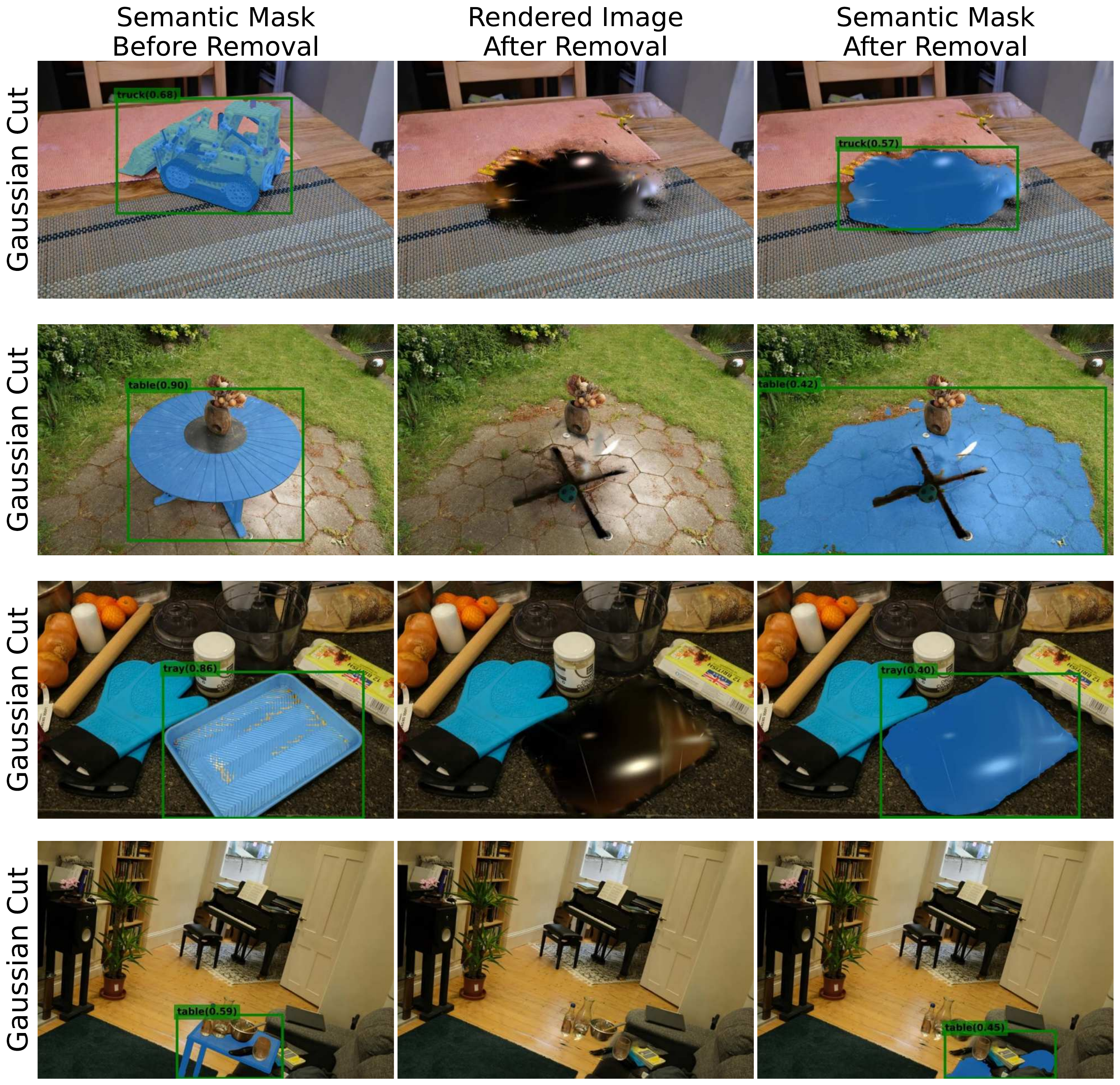}
\end{figure*}

\begin{figure*}[t]
\centering
\caption{\textbf{Category-prompt-free structural changes on Mip-NeRF360 (1/3).} Without physical post-removal references, we compare Segment Anything masks~\cite{kirillov2023segany,ravi2024sam2} before and after removal and report $1-\mathrm{sim}_{\mathrm{SAM}}$.}
\label{fig:supp_mip_struct_1}
\begin{subfigure}[t]{0.85\linewidth}\centering\includegraphics[width=\linewidth]{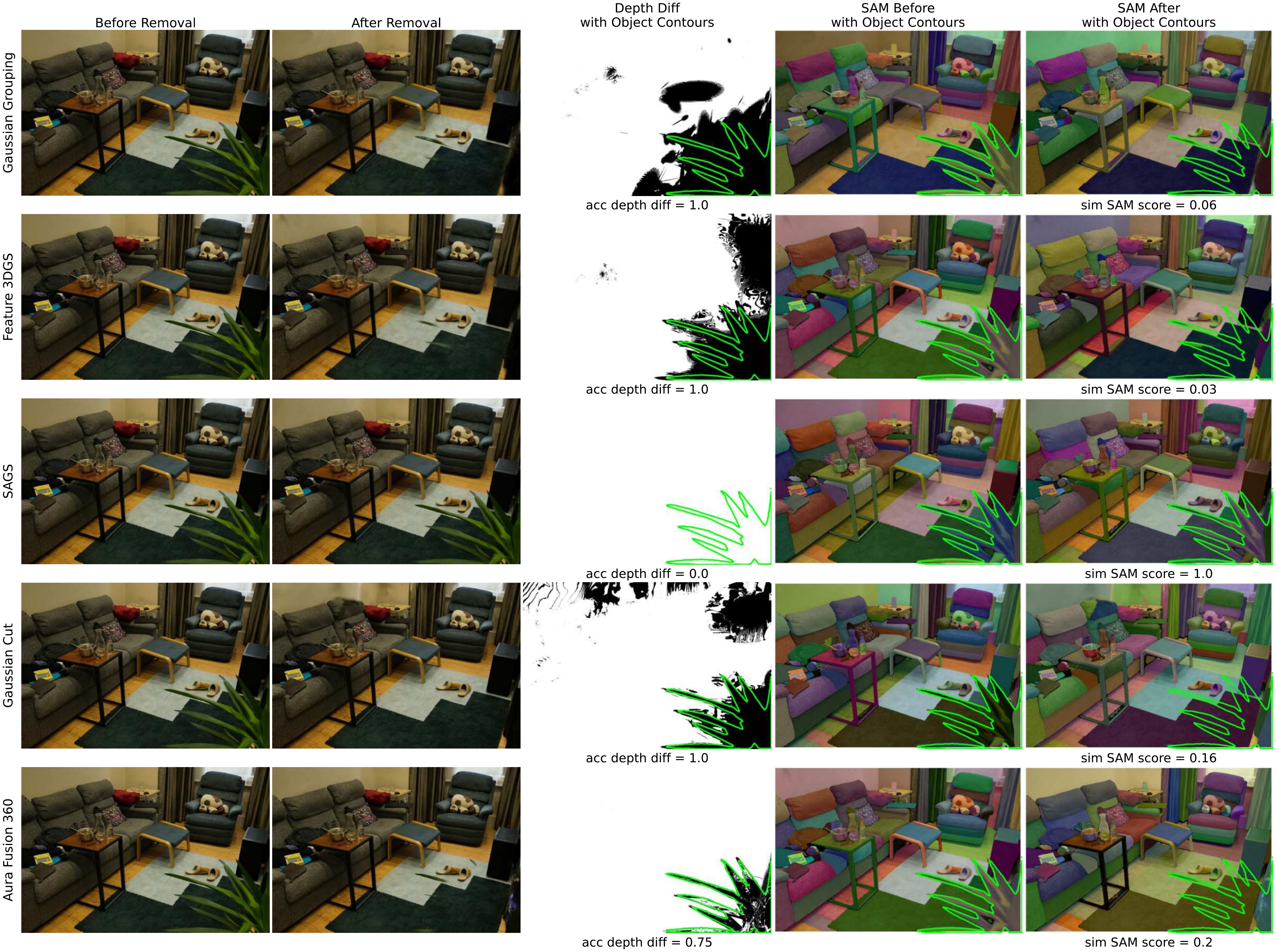}\caption{Plant.}\end{subfigure}\hfill
\begin{subfigure}[t]{0.85\linewidth}\centering\includegraphics[width=\linewidth]{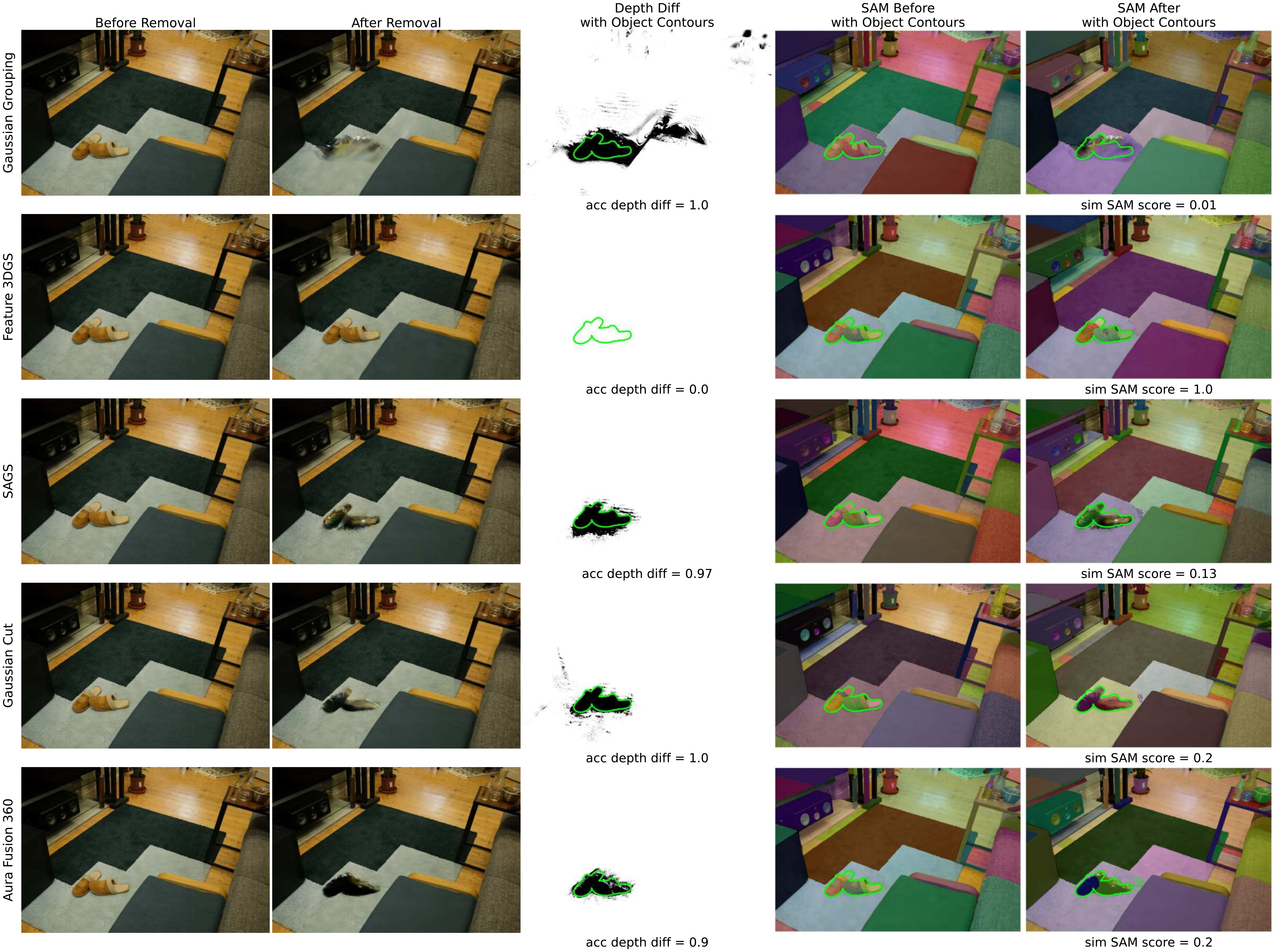}\caption{Slippers.}\end{subfigure}
\end{figure*}

\begin{figure*}[t]
\centering
\caption{\textbf{Category-prompt-free structural changes on Mip-NeRF360 (2/3).} The structural score measures the extent of change from the original rendering and is not a direct measure of correctness.}
\label{fig:supp_mip_struct_2}
\begin{subfigure}[t]{0.95\linewidth}\centering\includegraphics[width=\linewidth]{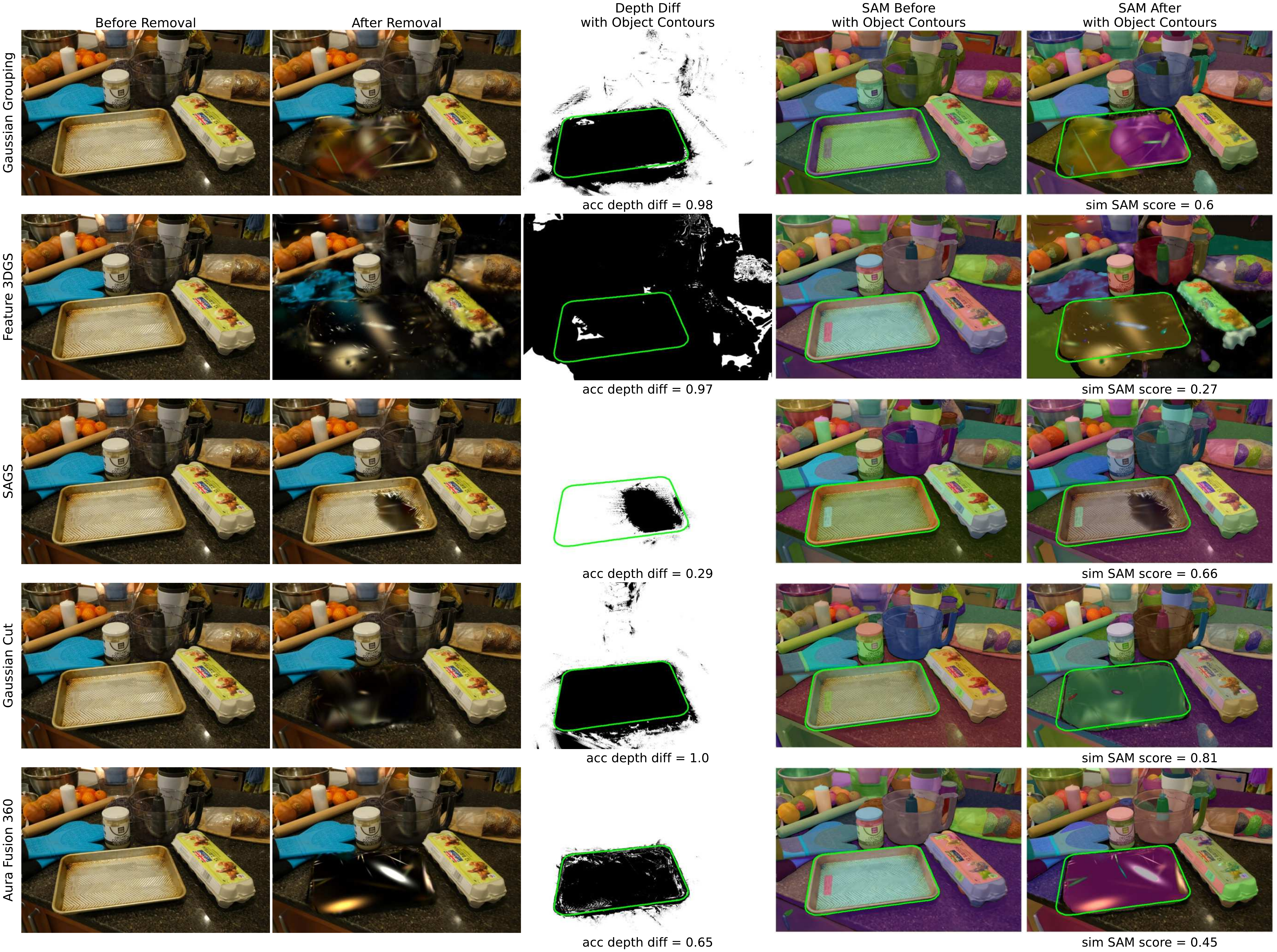}\caption{Baking Tray.}\end{subfigure}\hfill
\begin{subfigure}[t]{0.95\linewidth}\centering\includegraphics[width=\linewidth]{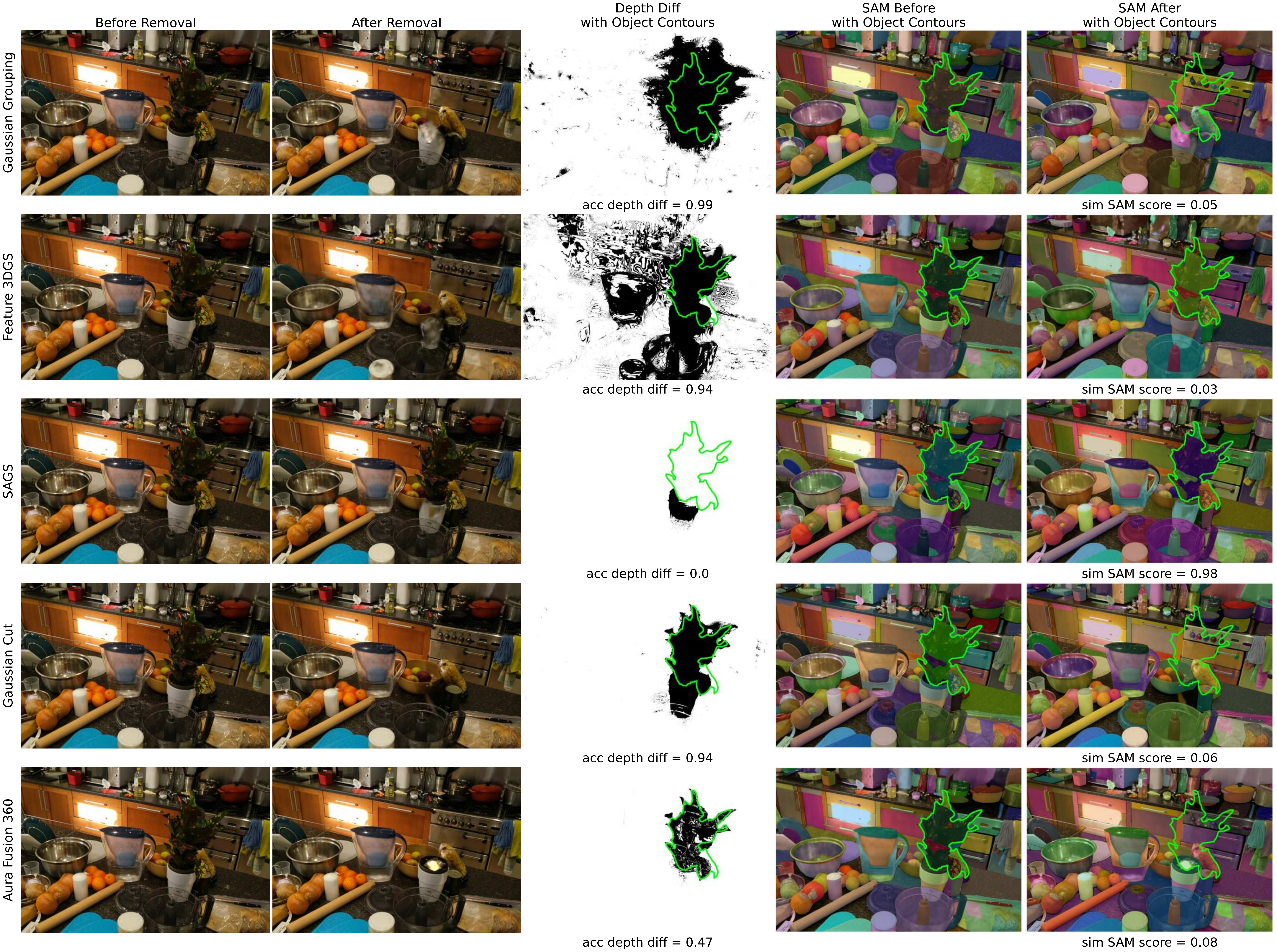}\caption{Plant.}\end{subfigure}
\end{figure*}

\begin{figure*}[t]
\centering
\caption{\textbf{Category-prompt-free structural changes on Mip-NeRF360 (3/3).} These examples illustrate why the reference-free structural axis should be interpreted together with semantic and geometric modification.}
\label{fig:supp_mip_struct_3}
\begin{subfigure}[t]{0.95\linewidth}\centering\includegraphics[width=\linewidth]{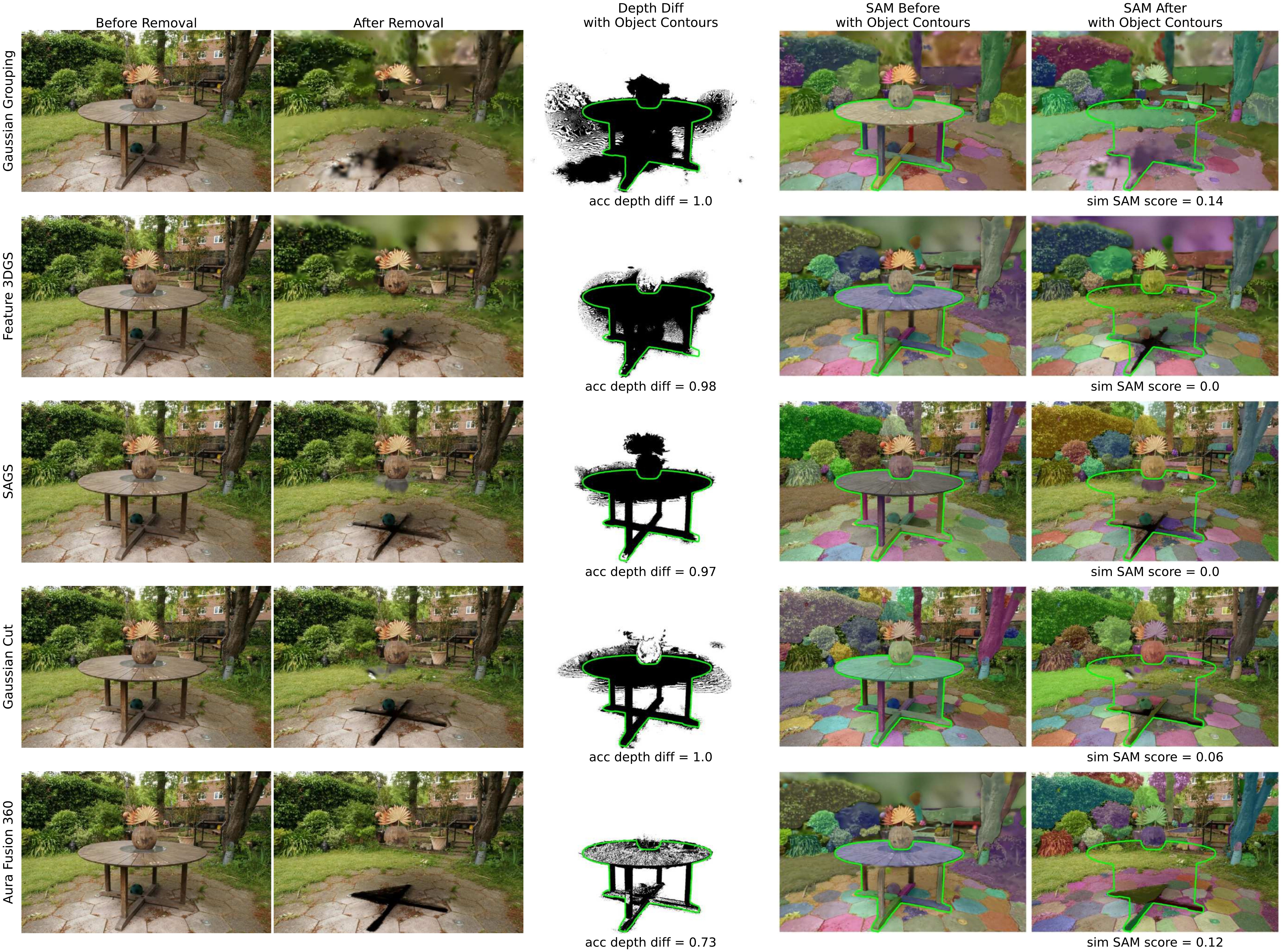}\caption{Table.}\end{subfigure}\hfill
\begin{subfigure}[t]{0.95\linewidth}\centering\includegraphics[width=\linewidth]{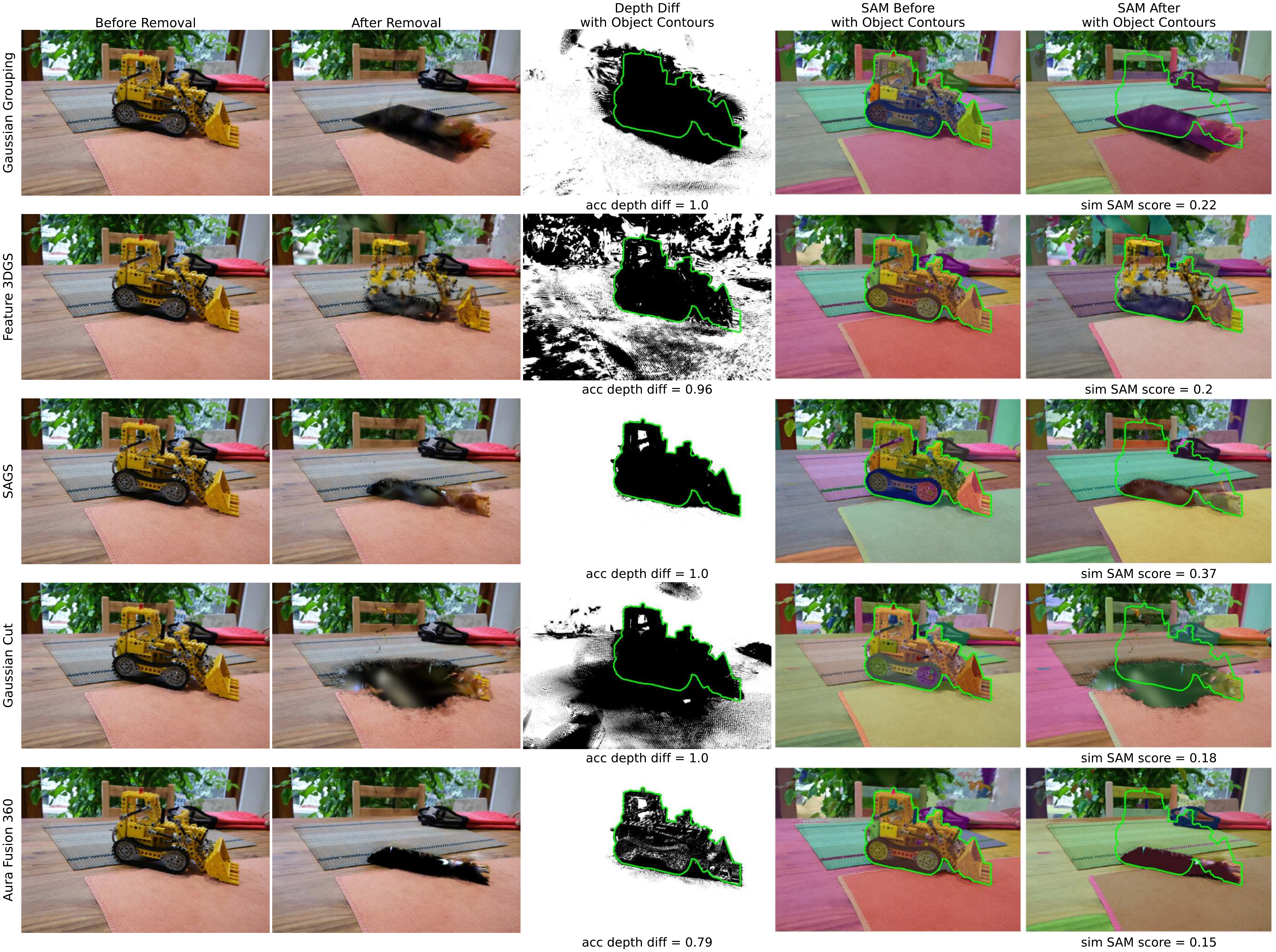}\caption{Truck.}\end{subfigure}
\end{figure*}

\end{document}